\documentclass[11pt]{article}

\usepackage[utf8]{inputenc}
\usepackage[T1]{fontenc}
\usepackage{graphicx}
\usepackage{lmodern}
\usepackage[margin=1in]{geometry}
\usepackage{amsmath,amssymb}
\usepackage{amsthm}
\usepackage{array}
\usepackage{booktabs}
\usepackage{longtable}
\usepackage[colorlinks=true,linkcolor=red,citecolor=blue,urlcolor=blue]{hyperref}
\usepackage{makecell}
\usepackage[most]{tcolorbox}
\usepackage[section]{placeins}
\usepackage{tikz}
\usetikzlibrary{arrows.meta,positioning,fit,calc,backgrounds}

\let\origunderscore\_
\renewcommand{\_}{\origunderscore\allowbreak}
\theoremstyle{remark}
\newtheorem*{remark}{Remark}

\tcbset{
  promptbox/.style={
    enhanced,
    breakable,
    boxrule=0.7pt,
    arc=1.8mm,
    left=1.2mm,
    right=1.2mm,
    top=0.8mm,
    bottom=0.8mm,
    coltitle=black,
    fonttitle=\bfseries,
    title style={left color=white,right color=white},
  }
}

\newcommand{\cliffsearchlogopath}{assets/CliffSearch_icon.png}

\title{%
\texorpdfstring{%
\raisebox{-0.22\height}{\includegraphics[height=2.6em]{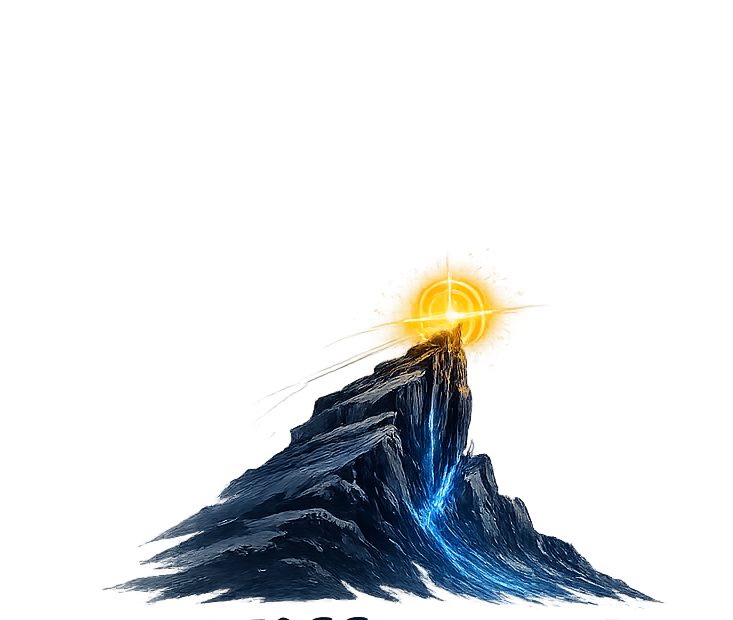}}\hspace{0.45em}%
CliffSearch: Structured Agentic Co-Evolution\\  over Theory and Code for Scientific Algorithm Discovery %
}{%
CliffSearch: Structured Evolution over Theory and Code for Scientific Algorithm Discovery%
}%
}
\author{Youssef Mroueh\footnote{Correspondence: mroueh@us.ibm.com}~, Carlos Fonseca, Brian Belgodere, David  Cox\\ ~~\\IBM Research\\ \url{https://cliffsearch.ai}}\date{\today}

\begin{document}
\maketitle

\begin{abstract}
Scientific algorithm discovery is iterative: hypotheses are proposed, implemented, stress-tested, and revised. Current LLM-guided search systems accelerate proposal generation, but often under-represent scientific structure by optimizing code-only artifacts with weak correctness/originality gating. We present CliffSearch, an agentic evolutionary framework in which the core evolution operators (pair selection, crossover, mutation, and review) are implemented as LLM agents, and the loop is designed around three principles: (1) each node is a structured scientific artifact, instantiated in either \texttt{theory+code} or \texttt{code\_only} mode, (2) reviewer judgments of correctness and originality are first-class selection gates alongside optimization of the benchmark metric of interest, and (3) mutation is split into exploration and correction pathways with distinct objectives. Exploration mutation imports ideas from adjacent scientific domains to increase novelty, while correction mutation performs targeted evidence-guided repair using reviewer signals over theory, code, benchmark results, and runtime errors. Even in \texttt{code\_only} mode, nodes still carry a required \texttt{summary\_md} field that records the design principles being proposed, so the mode suppresses formal \texttt{theory\_content} rather than eliminating ideation altogether; in practice this also frees more of the agent token budget for executable code. We illustrate the framework on three benchmark-grounded studies: transformer hyper-connection evolution, optimizer discovery on a fixed nanoGPT stack, and a smaller native-optimizer ablation. Across these settings, the same loop supports explicit metric direction, reproducible persistence, and reviewer-gated comparison of discoveries under controlled search conditions. The result is a discovery workflow that prioritizes scientific interpretability and correctness while optimizing task metrics under controlled novelty constraints, rather than maximizing candidate throughput alone. Full run artifacts, interactive visualizations, and exported best nodes for the reported studies are available at \url{https://cliffsearch.ai}.
\end{abstract}

\section{Introduction}
Scientific algorithm discovery rarely comes from a single jump; it comes from a loop in which mathematical ideas and executable implementations co-evolve. Evolutionary computation offers robust search machinery for difficult spaces \cite{holland1975,goldberg1989,real2019}, and LLMs now provide strong proposal engines for symbolic artifacts \cite{brown2020,vaswani2017}. Their combination has produced promising discovery systems \cite{romeraparedes2024,funbo2025,alphaevolve2025}, but current workflows still face a practical gap between fast generation and scientifically trustworthy selection.

Three limitations are especially consequential for scientific settings. First, many loops optimize code-only candidates, which weakens traceability between conceptual claims and implementation behavior. Second, benchmark outcomes alone do not guarantee validity: high score can coexist with weak originality or fragile correctness. Third, a single undifferentiated mutation policy conflates two distinct goals, namely broad exploration and targeted repair.

CliffSearch is built to address those gaps directly. The search unit is a structured artifact carried in either \texttt{theory+code} or \texttt{code\_only} mode, and the evolutionary operators themselves are instantiated as LLM agents (pair selector, crossover, exploration mutation, correction mutation, reviewer). In \texttt{code\_only} mode, \texttt{theory\_content} is intentionally empty, but each node still carries a required \texttt{summary\_md} field that records the intended mechanism and design principles; the mode therefore suppresses formal theory files without reducing the loop to blind code search, and it frees more of the token budget for the code artifact itself. Every evaluated node is reviewed for correctness and originality, and those reviewer outputs are hard gates in winner selection. Mutation is explicitly split into \emph{exploration mutation} (adjacent-domain novelty transfer) and \emph{correction mutation} (evidence-guided repair). This separation keeps novelty pressure while improving recovery from invalid or weak candidates. The same exploration operator is also used at initialization time: if the configured human seeds are fewer than the target population size, generation 0 is completed by exploration-mutation children derived from those seeds until the population closes.

These scientific design choices are supported by an execution substrate that makes large runs auditable: explicit benchmark metric direction, deterministic score normalization, and generation-level persistence for replay and analysis. Transformer hyper-connection evolution, optimizer discovery on a fixed nanoGPT stack, and a smaller native-optimizer ablation are the empirical studies reported in this paper; the framework itself is task-agnostic and can be instantiated for any user-defined scientific task given a task specification, benchmark protocol, and metric definition. For the hyper-connection attention discovery task, we also ran CliffSearch in matched \texttt{code\_only} mode. In that hyper-connection attention study, the matched \texttt{code\_only} mode produces two genuine geometric breakthroughs, a Givens orthogonal-manifold branch and a Poincar\'e hyperbolic branch, which are then recombined into a reviewer-valid width/depth hybrid that reaches mean validation loss 0.00733 under the fixed benchmark while remaining literature-grounded rather than purely retrieval-like. That width/depth recombination is one of the most salient discoveries in the paper.

We make four contributions. (1) A scientific-artifact-centered evolutionary loop that supports both \texttt{theory+code} and \texttt{code\_only} artifact modes. (2) Reviewer-gated selection where correctness and originality are first-class constraints, not post-hoc diagnostics. (3) A two-path mutation design that separates exploratory novelty from corrective repair. (4) A unified, benchmark-grounded runtime validated across multiple algorithm-discovery tasks, with transformer hyper-connection evolution, optimizer discovery on a fixed nanoGPT stack, and a native-optimizer ablation reported here, plus full node-level appendix evidence for the reported runs.

\section{Related Work}
Automated algorithm design predates LLMs and is grounded in hyper-heuristics, genetic programming, and metaheuristic design frameworks \cite{burke2013hyper,zhao2024metaheuristic}. This classical line established the core idea that search procedures themselves can be discovered rather than manually authored. Recent LLM-era work extends this paradigm with language-guided proposal operators and reflective loops.

Within LLM-driven discovery, representative systems include FunSearch \cite{romeraparedes2024}, AEL \cite{ael2023}, EoH \cite{eoh2024}, ReEvo \cite{ye2024reevo}, LLaMEA \cite{vanstein2024llamea}, and AlphaEvolve \cite{alphaevolve2025}, with platform/benchmark efforts such as LLM4AD \cite{liu2024llm4ad} and open implementations such as OpenEvolve \cite{openevolve2025}. These works collectively demonstrate that LLM-guided evolutionary search can produce high-value algorithmic artifacts across multiple domains.

Very recent systems make the iterative loop more explicit. DeepEvolve integrates external knowledge retrieval, cross-file code editing, debugging, and test-driven refinement in a feedback loop for executable algorithm improvement \cite{deepevolve2025}. Algorithmist emphasizes a proof-first, multi-agent research-and-review workflow that separates ideation, proof development, implementation, and proof--code alignment review \cite{algorithmist2026}. CliffSearch is closest in spirit to these iterative agent-mediated approaches, but differs in control mechanism: it is explicitly population-based and evolutionary, with LLM-instantiated crossover/mutation operators over structured artifacts in either \texttt{theory+code} or \texttt{code\_only} mode, and with correctness/originality as hard survival gates alongside benchmark score.

Parallel progress has also emerged on the Bayesian-optimization side of algorithm discovery, including LLM-enhanced BO \cite{liu2024llambo}, acquisition-function evolution \cite{yao2024evolcaf,funbo2025}, and end-to-end BO-algorithm generation \cite{li2025llameabo}. Hybrid BO-EA formulations for scientific design loops are increasingly visible as well \cite{low2024egbo}.

In application literature, high-impact scientific discovery systems are often realized as closed-loop experimental platforms using BO/active learning and autonomous laboratories \cite{gongora2020bear,macleod2020sdl,burger2023autolab}. Evolutionary methods also remain important for inverse design and materials/molecular search \cite{collins2016mof}. This context motivates our emphasis on scientific reliability signals in addition to performance optimization.

CliffSearch is aligned with these directions but differs in three specific ways: (1) the optimization object is a structured scientific artifact (theory+code, with a compatible code-only mode) rather than code alone; (2) reviewer outputs (correctness and originality) are hard selection gates alongside benchmark metrics; and (3) mutation is explicitly split into exploration (adjacent-domain novelty transfer) and correction (evidence-guided repair). Relative to AlphaEvolve \cite{alphaevolve2025} and OpenEvolve \cite{openevolve2025}, our goal is not a new generic evolutionary primitive, but a stricter scientific-discovery substrate with explicit artifact semantics and decision controls.

\section{CliffSearch Framework}
\subsection{Task definition and optimization target}
CliffSearch starts from a user-defined task contract: what is being optimized, how candidates are represented, which benchmark protocol/adapter must be executed, and which benchmark primary metric determines objective quality. This task contract is provided via \texttt{task\_type}, \texttt{task\_preamble}, and runtime grounding constraints, and is passed to all agents and benchmark adapters. The evolutionary loop is therefore generic, while task semantics (including benchmark definition) are explicit and injectable.

\subsection{Node representation and artifact mode}
\begin{figure}[t]
\centering
\begin{tikzpicture}[
  scale=0.86,
  transform shape,
  node distance=0.95cm and 1.05cm,
  seed/.style={draw, rounded corners, align=center, minimum width=3.25cm, minimum height=0.95cm, fill=green!10},
  evo/.style={draw, rounded corners, align=center, minimum width=3.35cm, minimum height=0.95cm, fill=orange!12},
  nodebox/.style={draw, rounded corners, align=left, text width=4.25cm, minimum height=2.0cm, fill=blue!8},
  arr/.style={-Latex, thick}
]
  \node[seed] (seed1) {Human Seed A\\Theory + Code};
  \node[seed, below=0.75cm of seed1] (seed2) {Human Seed B\\Theory + Code};
  \node[evo, right=1.35cm of seed1] (ops) {Evolution Operators\\crossover + mutation};
  \node[nodebox, right=1.35cm of ops] (cand) {\scriptsize \textbf{Candidate Node}\\
  \scriptsize \texttt{summary\_md}\\
  \scriptsize \texttt{theory\_content}\\
  \scriptsize \texttt{code\_content}\\
  \scriptsize (+ benchmark/review after evaluation)};

  \draw[arr] (seed1) -- (ops);
  \draw[arr] (seed2.east) to[out=0,in=-130] (ops.south);
  \draw[arr] (ops) -- (cand);
\end{tikzpicture}
\caption{Structured node artifact evolved across generations. In \texttt{code\_only}, \texttt{theory\_content} is present but empty, while \texttt{summary\_md} still carries the node's explanatory prose.}
\label{fig:nature-node}
\end{figure}
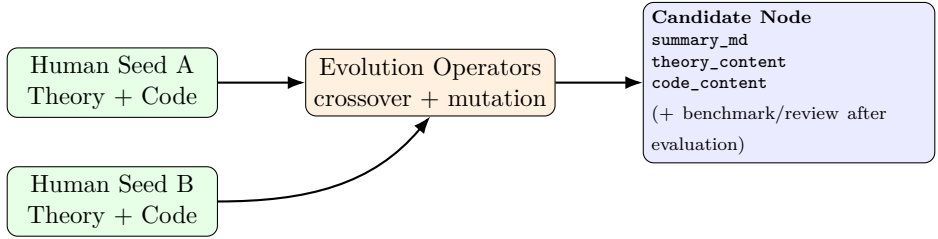

The search unit is a node represented canonically as a strict JSON artifact with three content fields: \texttt{summary\_md}, \texttt{theory\_content}, and \texttt{code\_content}. These fields are the object actually evolved by the agents and passed between crossover, mutation, benchmark, review, storage, and visualization. Their roles are distinct. \texttt{code\_content} is the executable artifact that is injected into the benchmark adapter after validation. \texttt{theory\_content} is the scientific rationale for the mechanism: assumptions, geometric or algorithmic claims, and the intended logic behind the implementation. \texttt{summary\_md} is a compact natural-language synopsis of the node meant for communication inside the evolutionary loop. In particular, it gives pair selection and other summary-restricted stages a concise description of what the node is trying to do, what changed relative to parents, and why it may be promising, without requiring those stages to reread the full theory and code every time. In \texttt{artifact\_mode=code\_and\_theory}, both code and theory are active optimization artifacts. In \texttt{artifact\_mode=code\_only}, the same node schema is kept but \texttt{theory\_content} is empty by contract, while \texttt{summary\_md} still carries in prose the node's design principles, intended mechanism, and any theoretical intuition needed to explain the code. This preserves storage and visualizer compatibility without removing explanatory context from the evolutionary loop.
Figure~\ref{fig:nature-node} summarizes this node-centric representation and how seeds feed generation operators.

At each iteration, the node remains the canonical object passed across pairing, crossover, mutation, benchmark, review, and persistence. After evaluation, the same node is enriched with benchmark fields (primary metric, summary, details, artifacts) and then with review fields (correctness, originality, reviewer narrative), so the evolutionary state stays attached to one persistent artifact rather than being scattered across separate records.

\subsection{Agents and operational roles}
The framework instantiates five agent roles: reviewer, pair selector, crossover, exploration mutation, and correction mutation.

\paragraph*{Reviewer agent.}
In each generation, every node that enters evaluation is first executed on the benchmark adapter, then sent to the reviewer with node artifacts plus full benchmark payload and node metadata. In code, this reviewer payload includes benchmark summary and benchmark details (including error fields/log excerpts when present), as well as lineage metadata that carries parent context (for example parent benchmark fields and parent correctness/originality scores when present). Reviewer emits \texttt{correctness\_score} and \texttt{originality\_score} with narrative evidence (Appendix~\ref{app:reviewer-prompt}).

\paragraph*{Pair selector agent.}
The winner rule (defined below) combines directional benchmark score with reviewer outputs; only winners are eligible for pairing. Pair selector consumes a summary-restricted view of winners (\texttt{id}, \texttt{summary\_md}, score, and review flags) and outputs a bounded list of parent-id pairs (Appendix~\ref{app:pair-selector-prompt}). Runtime sanitization then enforces deterministic validity constraints: each id must belong to the winner set, self-pairs are rejected, disjointness policy is enforced, and pair-count caps are applied.

\paragraph*{Crossover agent.}
For each selected pair, crossover receives both parent node payloads with task grounding and emits one strict child payload with \texttt{summary\_md}, \texttt{theory\_content}, and \texttt{code\_content} (Appendix~\ref{app:crossover-prompt}). Outputs are schema-validated before benchmark execution. Invalid outputs or invocation failures do not halt the run: runtime falls back to a deterministic child construction so population update remains total and auditable.

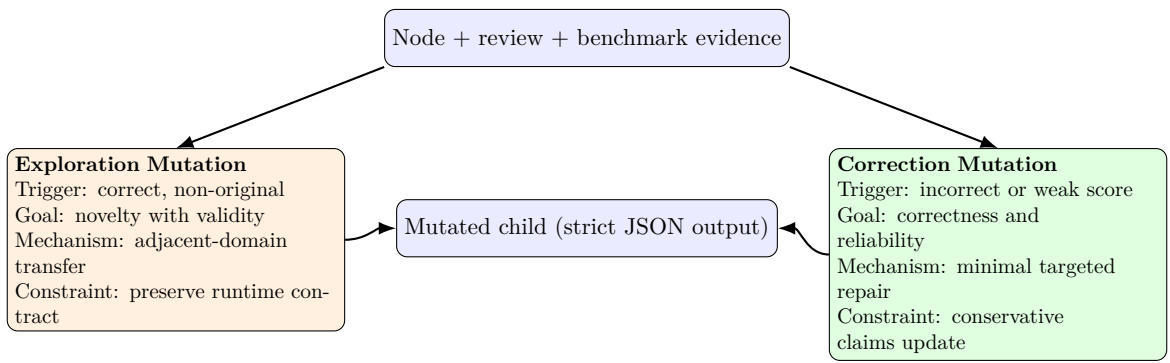
\begin{figure}[!htbp]
\centering
\begin{tikzpicture}[
  scale=0.80,
  transform shape,
  node distance=0.95cm and 1.1cm,
  state/.style={draw, rounded corners, align=center, minimum width=3.7cm, minimum height=0.95cm, fill=blue!8},
  opx/.style={draw, rounded corners, align=left, text width=5.35cm, minimum height=2.0cm, fill=orange!12, font=\small},
  opc/.style={draw, rounded corners, align=left, text width=5.35cm, minimum height=2.0cm, fill=green!12, font=\small},
  arr/.style={-Latex, thick}
]
  \node[state] (src) {Node + review + benchmark evidence};
  \node[opx, below left=1.35cm and 0.65cm of src] (explore) {\textbf{Exploration Mutation}\\
  Trigger: correct, non-original\\
  Goal: novelty with validity\\
  Mechanism: adjacent-domain\\
  transfer\\
  Constraint: preserve runtime contract};
  \node[opc, below right=1.35cm and 0.65cm of src] (correct) {\textbf{Correction Mutation}\\
  Trigger: incorrect or weak score\\
  Goal: correctness and\\
  reliability\\
  Mechanism: minimal targeted\\
  repair\\
  Constraint: conservative\\
  claims update};
  \node[state, below=2.2cm of src] (out) {Mutated child (strict JSON output)};

  \draw[arr] (src.south west) -- (explore.north);
  \draw[arr] (src.south east) -- (correct.north);
  \draw[arr] (explore.east) to[out=0,in=180] (out.west);
  \draw[arr] (correct.west) to[out=180,in=0] (out.east);
\end{tikzpicture}
\caption{Two mutation operators with distinct routing and objectives.}
\label{fig:nature-mutation}
\end{figure}

\paragraph*{Mutation agents.}
Mutation is split into two distinct operators with different routing triggers and objectives. Exploration mutation is novelty-seeking and can import mechanisms from adjacent domains under task constraints (Appendix~\ref{app:exploration-prompt}); correction mutation is evidence-guided repair for incorrect or weak-score nodes (Appendix~\ref{app:correction-prompt}). Both operators consume full parent-node context (artifacts + benchmark + review + lineage metadata), emit the same strict child schema as crossover, and are subjected to the same validation/fallback path. Figure~\ref{fig:nature-mutation} makes this two-path mutation routing explicit.

\subsection{Winner rule and directional score}
Let \(P_g\) denote the population at generation \(g\), and let \(n\in P_g\) denote one node (candidate artifact). For generation \(g\), each evaluated node receives benchmark and reviewer outputs. From benchmark output, node \(n\) gets
\[
\big(m_{\mathrm{bench}}(n),\, h(n)\big)
\]
where \(m_{\mathrm{bench}}(n)=\texttt{benchmark.primary\_metric}\) (the measured metric) \\and \(h(n)=\texttt{benchmark.higher\_is\_better}\) (its direction flag). Winners are defined as:
\[
\mathcal{W}_g = \{n \in P_g : \mathrm{correct}(n)=1,\ \mathrm{original}(n)=1,\ s(n)>\mathrm{median}(P_g)\}.
\]
In prose: a node is a winner only if it passes both reviewer gates (correct and original) and its directional score is above the generation threshold. Here \(\mathrm{median}(P_g)\) is the median directional score across evaluated nodes in generation \(g\), so winner gating is relative to current-generation performance rather than an absolute global cutoff.
Selection does not compare raw measured metrics directly. It compares normalized score \(s(n)\), which is always higher-is-better:
\[
s(n)=
\begin{cases}
 m_{\mathrm{bench}}(n), & h(n)=\mathrm{true}\\
 -m_{\mathrm{bench}}(n), & h(n)=\mathrm{false}.
\end{cases}
\]
Reviewer outputs are therefore hard eligibility gates, not optional diagnostics.
The winner set is the only input population eligible for pair selection and crossover.

\subsection{Generation composition and update cycle}
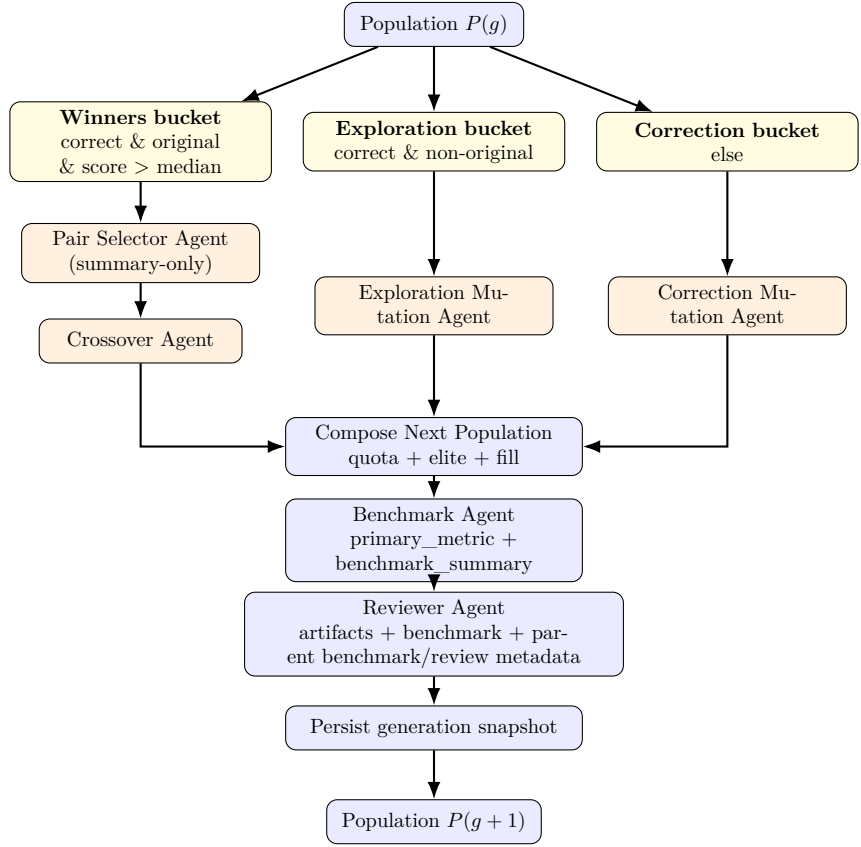
\begin{figure}[t]
\centering
\begin{tikzpicture}[
  scale=0.78,
  transform shape,
  stage/.style={draw, rounded corners, align=center, text width=4.0cm, minimum height=0.75cm, fill=blue!8, font=\small},
  agent/.style={draw, rounded corners, align=center, text width=3.8cm, minimum height=0.75cm, fill=orange!12, font=\small},
  rules/.style={draw, rounded corners, align=center, text width=4.2cm, minimum height=1.0cm, fill=yellow!14, font=\small},
  arr/.style={-Latex, thick}
]
  \node[stage, text width=2.8cm] (p) at (0,0) {Population $P(g)$};

  \node[rules] (winners) at (-5.0,-2.0) {\textbf{Winners bucket}\\correct \& original\\\& score $>$ median};
  \node[rules] (explore_bucket) at (0,-2.0) {\textbf{Exploration bucket}\\correct \& non-original};
  \node[rules] (correct_bucket) at (5.0,-2.0) {\textbf{Correction bucket}\\else};

  \node[agent] (pair) at (-5.0,-3.9) {Pair Selector Agent\\(summary-only)};
  \node[agent, text width=3.2cm] (cross) at (-5.0,-5.4) {Crossover Agent};

  \node[agent] (explore) at (0,-4.8) {Exploration Mutation Agent};
  \node[agent] (correct) at (5.0,-4.8) {Correction Mutation Agent};

  \node[stage, text width=4.8cm] (compose) at (0,-7.2) {Compose Next Population\\quota + elite + fill};
  \node[stage, text width=4.8cm] (bench) at (0,-8.8) {Benchmark Agent\\primary\_metric + benchmark\_summary};
  \node[stage, text width=6.2cm] (review) at (0,-10.4) {Reviewer Agent\\artifacts + benchmark + parent benchmark/review metadata};
  \node[stage, text width=4.8cm] (persist) at (0,-12.0) {Persist generation snapshot};
  \node[stage, text width=3.4cm] (next) at (0,-13.6) {Population $P(g+1)$};

  \draw[arr] (p) -- (winners);
  \draw[arr] (p) -- (explore_bucket);
  \draw[arr] (p) -- (correct_bucket);

  \draw[arr] (winners) -- (pair);
  \draw[arr] (pair) -- (cross);
  \draw[arr] (cross.south) |- (compose.west);

  \draw[arr] (explore_bucket) -- (explore);
  \draw[arr] (explore.south) -- (compose.north);

  \draw[arr] (correct_bucket) -- (correct);
  \draw[arr] (correct.south) |- (compose.east);

  \draw[arr] (compose) -- (bench);
  \draw[arr] (bench) -- (review);
  \draw[arr] (review) -- (persist);
  \draw[arr] (persist) -- (next);
\end{tikzpicture}
\caption{One generation cycle: bucket routing, agent operators, composition, benchmark, and review.}
\label{fig:nature-loop}
\end{figure}

Given winners, pair selector proposes parent pairs from summary-only winner views, and crossover generates children from those pairs. Nodes that are not winners are routed to mutation: correct-but-non-original nodes go to exploration mutation, while incorrect nodes or weak-score nodes go to correction mutation. Population composition then applies quota budgets for elite copies, crossover children, and mutation children, with fill fallback to guarantee exact population-size closure. The same fixed-size rule is already active at generation 0: seed artifacts are inserted first, and any remaining slots are bootstrapped by exploration mutation from the seed pool.
The full one-generation operator/evaluation order is shown in Figure~\ref{fig:nature-loop}.

\subsection{Runtime contracts, persistence, and distributed execution}
All agent interfaces are strict JSON contracts with normalization and fail-fast validation. SDK agent calls are text-only and do not receive tools for arbitrary code execution or shell-side actions. Instead, generated code artifacts are executed only inside the benchmark adapters, after schema normalization and task-specific contract checks. Malformed outputs, missing required keys, invalid pair references, and contract violations are rejected locally before expensive benchmark execution. This separation is deliberate: it keeps most agent calls cheaper while confining code execution to the validated benchmark path. In the reported single-island nanoGPT experiments (population \(8\), three evolutionary generations, three benchmark random seeds per node), wall-clock runtime is dominated by benchmark execution rather than SDK latency, because each evaluated node triggers repeated train/eval runs while agent calls remain lightweight text generations. Under AWS Claude Opus 4.6, the SDK spend for such a run stayed in the single-digit US-dollar range, roughly under \$5--10, while benchmark compute dominated the end-to-end runtime budget. Reviewer context is mode-aware: in \texttt{code\_and\_theory}, reviewer consumes code+theory+benchmark+lineage metadata; in \texttt{code\_only}, reviewer consumes code+benchmark+lineage metadata. We maintain dedicated code-only reviewer prompts for that case, and those prompts explicitly instruct the reviewer to assess the node from \texttt{code\_content} as primary evidence, with benchmark and lineage context, to treat \texttt{summary\_md} only as secondary context, and to ignore \texttt{theory\_content}.

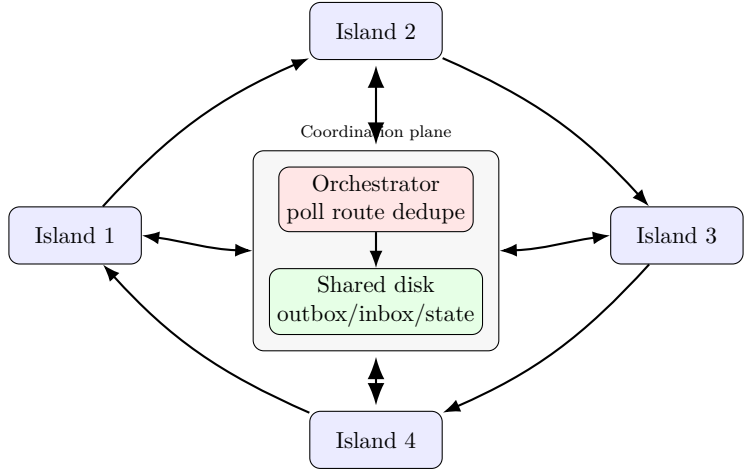
\begin{figure}[t!]
\centering
\begin{tikzpicture}[
  scale=0.80,
  transform shape,
  island/.style={draw, rounded corners, align=center, minimum width=2.2cm, minimum height=0.95cm, fill=blue!8},
  orch/.style={draw, rounded corners, align=center, minimum width=3.0cm, minimum height=0.95cm, fill=red!10},
  disk/.style={draw, rounded corners, align=center, minimum width=3.4cm, minimum height=0.95cm, fill=green!10},
  coord/.style={draw, rounded corners, fill=gray!6},
  ring/.style={-Latex, thick},
  flow/.style={Latex-Latex, thick},
  innerflow/.style={-Latex, thick}
]
  \node[island] (i1) at (-5.0,0.0) {Island 1};
  \node[island] (i2) at (0.0,3.4) {Island 2};
  \node[island] (i3) at (5.0,0.0) {Island 3};
  \node[island] (i4) at (0.0,-3.4) {Island 4};
  \node[orch] (orch) at (0.0,0.6) {Orchestrator\\poll route dedupe};
  \node[disk] (disk) at (0.0,-1.1) {Shared disk\\outbox/inbox/state};
  \begin{pgfonlayer}{background}
    \node[coord, fit=(orch)(disk), inner sep=6pt] (coordbox) {};
  \end{pgfonlayer}
  \node[font=\scriptsize, above=1pt of coordbox.north] {Coordination plane};

  \draw[ring] (i1) to[bend left=12] (i2);
  \draw[ring] (i2) to[bend left=12] (i3);
  \draw[ring] (i3) to[bend left=12] (i4);
  \draw[ring] (i4) to[bend left=12] (i1);

  \draw[flow] (i1.east) to[out=-8,in=178] (coordbox.west);
  \draw[thick, {Latex[length=3mm]}-{Latex[length=3mm]}, shorten <=2pt, shorten >=2pt]
    (i2.south) -- (coordbox.north);
  \draw[flow] (i3.west) to[out=-172,in=2] (coordbox.east);
  \draw[thick, {Latex[length=3mm]}-{Latex[length=3mm]}, shorten <=2pt, shorten >=2pt]
    (i4.north) -- (coordbox.south);

  \draw[innerflow] (orch.south) -- (disk.north);
\end{tikzpicture}
\caption{Distributed multi-island execution with asynchronous migration packets.}
\label{fig:nature-multi}
\end{figure}

Each generation persists node-level artifacts, \texttt{population.json}, and \texttt{ga\_data.json} snapshots, enabling deterministic replay and audit. In single-island mode, CPU workers execute agent/reviewer calls and GPU slots execute benchmark runs. In distributed mode, islands evolve independently and exchange migrants over shared storage via a lightweight orchestrator. The distributed topology and coordination plane are shown in Figure~\ref{fig:nature-multi}. This architecture supports heterogeneous model assignments per island and asynchronous migration policies without changing the per-node contract.

\section{Applications and Empirical Outcomes}
Our empirical illustrations focus on \textbf{algorithm discovery in machine learning}. Across all reported tasks, generation-0 starts from human seed artifacts and, whenever the seed count is below the configured population size, the remaining generation-0 slots are filled by exploration-mutation descendants of those seeds. Every candidate is then benchmarked with repeated random seeds under a task-specific adapter, and reviewer gates (correctness/originality) are applied before selection. The task-specific variation is therefore in what is evolved and which objective is optimized. In this draft we report two primary benchmark-grounded discovery settings: Transformer hyper-connection discovery \cite{vaswani2017} through the HC line \cite{zhu2024hyperconnections}, DeepSeek mHC \cite{xie2025mhc}, and mHC-lite permutation-basis hyper-connections \cite{mhclite2026}; and optimizer discovery on a fixed transformer training stack. We also report a smaller native-optimizer ablation that keeps the same reviewer-driven evolutionary loop but replaces nanoGPT with compact linear/MLP classification tasks in order to compare prompt bundles, artifact modes, and evolution-control settings in a cheaper regime.

\subsection{Transformer HyperConnection evolution}
\paragraph*{RNN $\rightarrow$ Transformer $\rightarrow$ hyper-connections.}
Sequence modeling progressed from recurrent architectures \cite{elman1990,hochreiter1997} to Transformer attention \cite{vaswani2017}, and then to richer residual mixing via HC \cite{zhu2024hyperconnections}, mHC \cite{xie2025mhc}, and mHC-lite \cite{mhclite2026,mhclite_repo}. That lineage matters here because each step tightened the geometry of the hyper-connection itself. HC introduced learned dense linear stream mixing, but those unconstrained maps can become poorly conditioned and amplify activations or gradients. mHC responded by constraining the hyper-connection to doubly stochastic operators on a manifold; mHC-lite kept that geometric view but reparameterized the same family through learned mixtures of permutation matrices. The task preamble therefore fixes the training stack, dataset, and evaluation protocol, and asks CliffSearch to evolve the manifold-parameterized hyper-connection: each node proposes theory and code for a \texttt{custom} module that defines the geometry used to mix residual streams, the parameterization of that constraint set, and the resulting expand/mix/merge maps applied inside attention and MLP blocks. What is being evolved is the hyper-connection law itself, not the optimizer, model scale, or benchmark recipe.

\paragraph{Theory+code CliffSearch Task and Setup.} The reported transformer hyper-connection run is a single-island \texttt{theory+code} experiment with population size \(8\), three generations of evolution after initialization, and AWS Claude Opus 4.6 used through the SDK for all agent roles. Unlike the optimizer study, this transformer hyper-connection case study was run only with the compact \texttt{short\_json} prompt bundle rather than the more prescriptive \texttt{workflow\_v2} prompts. The benchmark uses the Shakespeare character dataset, the fixed small-model nanoGPT configuration, three benchmark random seeds, and a shared AdamW training recipe with initial learning rate \(10^{-3}\) for every candidate under the fixed \texttt{hyper\_conn\_n=4} contract; the exact composed model/data stack is listed in Appendix~\ref{app:shared-nanogpt-small-model}. The three human seeds are \texttt{TransformerResidualAttentionSeed}, \texttt{MHCLiteAttentionSeed}, and \texttt{HCAttentionSeed}. Because the seed set was smaller than the target population, generation 0 was completed by exploration-mutation children bootstrapped from those three seeds. Benchmarking uses nanoGPT train+eval \cite{nanogpt_repo,mhclite_repo}, with strict contract checks for custom hyper-connections (\texttt{hyper\_conn\_type=custom}, required branch invocation). The primary metric is mean validation loss (lower is better), averaged over the benchmark random seeds; failed benchmark seeds are imputed by the worst successful loss. Because the benchmark hyperparameters are fixed across evolved nodes rather than retuned per node, the absolute validation losses are intentionally conservative and mainly support fair relative comparison under a shared evaluation recipe. As in the generic runtime discussion above, wall-clock time for this \texttt{theory+code} run was dominated by the repeated nanoGPT benchmarks rather than by SDK calls, and the Opus-side agent spend stayed in the single-digit US-dollar range for the full \(p=8\), \(g=3\) run. In the generation graph, nodes marked \(m=\infty\) denote benchmark execution failures rather than valid finite losses. In this \texttt{theory+code} run, the most important seed failures were residual-dtype incompatibilities in the imported mHC-lite and HC seed implementations. Those failures appeared explicitly in benchmark logs, were flagged by the reviewer agent together with repair plans, and were later fixed by explicit dtype-preserving casts. Table~\ref{tab:main-shortlist-transformer} gives the compact shortlist discussed here; the full transformer \texttt{theory+code} node table appears in Table~\ref{tab:transformer-full-node-table}, and the remaining appendix materials for this run appear in Appendix~\ref{app:extended-results}, including the alias-annotated generation graph (Figure~\ref{fig:transformer-ga-graph}), the exported best-node artifact, the novelty audit, and the exact run-local task preamble in Appendix~\ref{app:task-preamble-transformer-hc}. The same run is also mirrored online with its full visualization and best-node export at \url{https://cliffsearch.ai}.

\begin{table}[!htbp]
\centering
\scriptsize
\begin{tabular}{>{\bfseries}c>{\bfseries}c>{\raggedright\arraybackslash}p{3.3cm}>{\raggedright\arraybackslash}p{1.7cm}rrr}
\toprule
Node & Gen & Alias & Provenance & \makecell[c]{Primary metric\\($\downarrow$)} & Corr & Orig\\
\midrule
E1 & 1 & \makecell[l]{HyperbolicRotation\\Routing} & mutation & 1.84487 & 4 & 4\\
A2 & 2 & \makecell[l]{GivensHyperbolic\\Routing} & crossover & 1.76830 & 4 & 4\\
G2 & 2 & \makecell[l]{HyperbolicRotation\\Routing} & \makecell[l]{elite\\copy} & 1.84487 & 4 & 4\\
\textcolor{blue!75!black}{H2} & \textcolor{blue!75!black}{2} & \textcolor{blue!75!black}{\makecell[l]{GrassmannianSubspace\\Routing}} & \textcolor{blue!75!black}{fill} & \textcolor{blue!75!black}{1.69347} & \textcolor{blue!75!black}{4} & \textcolor{blue!75!black}{4}\\
A3 & 3 & \makecell[l]{GrassmannianHyperbolic\\Routing} & crossover & 1.69830 & 4 & 3\\
G3 & 3 & \makecell[l]{GrassmannianSubspace\\Routing} & \makecell[l]{elite\\copy} & 1.69347 & 4 & 4\\
\bottomrule
\end{tabular}
\caption{Selected discoveries from the reported transformer \texttt{theory+code} single-island run. Column ``Provenance'' records how the node entered the population under quota-mode composition.}
\label{tab:main-shortlist-transformer}
\end{table}
\paragraph{Theory+code CliffSearch Results.} The \texttt{theory+code} discovery trajectory is informative because it does not stay inside the original permutation-mixture design space. One early branch, \texttt{D1} (\texttt{HyperbolicPoincareRoutingV2}), introduced a Poincar\'e-ball view in which streams are arranged by hyperbolic distance rather than by a flat mixture. That node was not itself the best performer, but it contributed a key geometric ingredient for later crossover. The graph also contains two engineering-repair nodes, \texttt{B1} and \texttt{C1}, which are straightforward dtype-preservation fixes for the imported mHC-lite and HC seeds rather than scientific discoveries. The first strong non-seed family was \textbf{E1}, obtained by mutation from a failing generation-0 hyperbolic-rotation prototype. E1 replaces the permutation-mixture geometry of mHC-lite with a full \(\mathrm{SO}(4)\) hyper-connection parameterized by Givens rotations. That jump is exactly the kind of move CliffSearch is meant to surface: instead of merely perturbing coefficients inside the existing manifold, the \texttt{theory+code} search moved to an orthogonal-group parameterization that preserves norm structure while remaining lightweight and interpretable. Node \textbf{A2} then arose by crossover between E1 and the weaker but conceptually rich hyperbolic parent \texttt{D1}, yielding a hybrid that kept rotation-based transport while introducing hyperbolic-distance gating; this improved mean validation loss from 1.8449 to 1.7683.

\noindent The most interesting qualitative jump came from node \textbf{H2}, which entered through fill after under-production elsewhere in quota mode. H2 changed the geometry again: instead of rotating all four streams, it projected them through a Grassmannian/Stiefel bottleneck and then lifted them back, effectively turning the hyper-connection into a learned low-dimensional subspace selection problem. H2 became the best node in the run with mean validation loss 1.6935. Node \textbf{A3} crossed H2 with A2, combining Grassmannian projection with the hyperbolic hierarchy signal inherited from the Poincar\'e branch; it remained competitive at 1.6983 but originality dropped to \(3\), so it was not admitted as a winner. Finally, \textbf{G2} and \textbf{G3} are elite propagations of E1 and H2 respectively, showing that the loop retained both the earlier \(\mathrm{SO}(4)\) family and the later Grassmannian bottleneck family once they proved stable under benchmark and review. Taken together, the run shows CliffSearch moving from dense and permutation-based hyper-connections toward new geometric control laws based on hyperbolic hierarchy, Givens-parameterized orthogonal transport, and Grassmannian subspace selection.

\paragraph{Theory+code discovery versus retrieval.} To separate discovery from possible retrieval, we prepared a post hoc survey of manifold-attention, geometry-aware fusion, and direct HC / mHC literature; the survey and its audit tables are given in Appendix~\ref{app:manifold-attention-survey}, Tables~\ref{tab:manifold-attention-survey}--\ref{tab:transformer-novelty-audit}. This survey was \emph{not} available to the agents: the SDK calls in CliffSearch were tool-free, had no external retrieval, and were not given these papers or notes in prompt context. The survey therefore serves only as an external novelty audit of the \texttt{theory+code} transformer run. The right question here is not whether every geometric ingredient is new in all of attention, but whether CliffSearch is replaying known HC mechanisms or exporting ideas from the broader attention literature into new hyper-connections. Under the broader manifold-attention audit, the Poincar\'e and exp-map branches (\texttt{D0}, \texttt{D1}, \texttt{A1}, \texttt{D2}, \texttt{D3}, \texttt{E3}, \texttt{H3}) are the most literature-adjacent: they sit close to known hyperbolic-attention motifs in which geometry changes scoring through distance or curved coordinates \cite{gulcehre2019hyperbolic,chen2022fully,yang2024hypformer,cho2023fps}. The Grassmannian branch, culminating in \texttt{H2}, is closest to Grassmannian self-attention and projector-embedding means \cite{wang2024grassmann}, but its use as a residual-stream bottleneck inside a hyper-connection module is not the architectural placement used in that literature, so we read it as recombinational novelty rather than direct retrieval. The direct HC survey narrows the comparison further: published HC-line manifolds are still mostly dense, doubly stochastic/Birkhoff, Stiefel, Grassmann, or spectral-norm constraints on the residual mixing matrix itself \cite{zhu2024hyperconnections,xie2025mhc,mhclite2026,zhou2026kromhc,mishra2026mhcgnn,sengupta2026jpmhc,liu2026shc}. Against that narrower baseline, the hyperbolic stream families in the \texttt{theory+code} run look less like retrieval from HC papers and more like transfers of broader attention geometry into the HC setting, while the Grassmannian and Stiefel branches sit in a middle ground: the manifolds themselves are now known in HC literature, but their realization here as custom hyper-connection operators still looks recombinational rather than copied.\\

\noindent A second novelty axis is whether the merge itself becomes geometric. In the surveyed attention literature, several models do perform intrinsic manifold aggregation: Einstein midpoints, Lorentz centroids, or related barycentric constructions. By contrast, the discovered HC nodes in this \texttt{theory+code} run almost never do. In this \texttt{theory+code} run, geometry is usually used to score, gate, rotate, or project streams, while the actual merge stays Euclidean residual addition. The one partial exception is \texttt{D2} (\texttt{HyperbolicExpMapRouting}), which exp-maps streams into the Poincar\'e ball, takes a plain Euclidean weighted sum in those ball coordinates, clamps back into the ball, and then log-maps back before the branch call. Because it uses neither Lorentz factors nor a true gyro-barycentric / Einstein-midpoint rule, we do not count it as intrinsic hyperbolic aggregation. The more informative interpretation is therefore that the \texttt{theory+code} search often imported geometric scoring and projection ideas without also discovering that the aggregation map itself should become geometric, even though that merge is part of the editable \texttt{custom} hyper-connection contract.\\

\noindent Our conservative conclusion is that this \texttt{theory+code} run contains all three categories: retrieval-adjacent geometric motifs, recombinational HC discoveries that transplant known manifolds into a new hyper-connection placement, and a smaller set of stronger novelty candidates, with the Givens/\(\mathrm{SO}(4)\) family still the clearest case.

\paragraph{CliffSearch in Code-only Mode Setup and Analysis.}
We also ran CliffSearch itself in a matched single-island transformer \texttt{code\_only} mode under the same \texttt{short\_json} prompt bundle, population size, generation count, benchmark seeds, and fixed nanoGPT benchmark recipe. The exact artifact-mode clause is reproduced in Appendix~\ref{app:task-preamble-transformer-hc-code-only}, and the full run export appears in Appendix~\ref{app:transformer-code-only}, including the complete node table (Table~\ref{tab:transformer-code-only-full-node-table}) and a dedicated novelty audit (Table~\ref{tab:transformer-code-only-novelty-audit}). In this mode \texttt{theory\_content} is intentionally empty, but \texttt{summary\_md} remains required and serves as the node's design-principles scaffold; reviewer judgments are still grounded primarily in \texttt{code\_content}, benchmark evidence, and lineage, with \texttt{summary\_md} used only as secondary context. Quantitatively, the run produced 32 total nodes, 18 finite benchmarked nodes, and 11 reviewer-valid non-seed nodes. Its selected best node, \texttt{PoincareGivensHybridHC}, reached mean validation loss 0.00733, far below both the best seed (\texttt{TransformerResidualAttentionSeed}, 4.8555) and the best node from the matched \texttt{theory+code} run (1.6935).

\begin{table}[!htbp]
\centering
\scriptsize
\begin{tabular}{c>{\raggedright\arraybackslash}p{3.5cm}>{\raggedright\arraybackslash}p{4.8cm}rrr}
\toprule
Gen & Alias & Role in the trajectory & \makecell[c]{Primary metric\\($\downarrow$)} & Corr & Orig\\
\midrule
0 & \makecell[l]{GivensOrthogonal\\ManifoldHC} & First strong orthogonal-manifold branch; replaces permutation mixtures with content-adaptive Givens \(\mathrm{O}(4)\) routing. & 5.2677 & 4 & 4\\
0 & \makecell[l]{HouseholderCayley\\ManifoldHC} & Parallel orthogonal family using Householder reflections for width and Cayley depth transport. & 5.4502 & 4 & 4\\
1 & PoincareHC & Exploration-mutation breakthrough: Poincar\'e-ball width routing with explicit gyro-midpoint aggregation. & 0.0085 & 4 & 4\\
2 & \makecell[l]{GivensWidthBeta\\DepthHC} & Bridge crossover retaining Givens width transport while stabilizing depth redistribution. & 0.0090 & 5 & 4\\
\textcolor{blue!75!black}{3} & \textcolor{blue!75!black}{\makecell[l]{PoincareGivens\\HybridHC}} & \textcolor{blue!75!black}{Best final hybrid: Givens width routing plus Poincar\'e-modulated depth scaling.} & \textcolor{blue!75!black}{0.00733} & \textcolor{blue!75!black}{5} & \textcolor{blue!75!black}{4}\\
\bottomrule
\end{tabular}
\caption{Key discoveries from the matched transformer \texttt{code\_only} run. This mode still carries non-empty \texttt{summary\_md} design notes, but the scientific claims here are grounded in code and benchmark behavior rather than in a separate \texttt{theory\_content} artifact.}
\label{tab:main-shortlist-transformer-code-only}
\end{table}

\noindent The trajectory is informative because it separates two genuine discovery families and then their later consolidation. First, generation 0 found a real orthogonal-manifold frontier through \texttt{GivensOrthogonalManifoldHC} and \texttt{HouseholderCayleyManifoldHC}; the search did not jump directly from the HC seeds to one lucky hyperbolic node. The Givens family is therefore a discovery in its own right: it replaces permutation-mixture routing with content-adaptive orthogonal stream transport and remains reviewer-valid before the hyperbolic branch becomes dominant. Second, generation 1 produced the strongest novelty event: \texttt{PoincareHC}, an exploration mutation from the HC seed, reduced mean validation loss to 0.0085 while remaining reviewer-valid. Code inspection shows why this node matters. Unlike the earlier \texttt{theory+code} run, whose shortlisted hyperbolic families mostly used geometry for scoring or projection while keeping the actual merge Euclidean, \texttt{PoincareHC} explicitly maps streams into the Poincar\'e ball, performs a gamma-weighted gyro-midpoint aggregation, and log-maps back before the branch call. In other words, the \texttt{code\_only} run did discover an intrinsic manifold aggregation rule rather than only a geometric scoring heuristic. Third, later crossover does not erase either discovery family; it consolidates them. \texttt{GivensWidthBetaDepthHC} shows that the orthogonal branch remains useful after the Poincar\'e breakthrough, and the final \texttt{PoincareGivensHybridHC} delegates width transport to the Givens family while retaining Poincar\'e geometry as curvature-aware depth modulation.

\paragraph{Code-only novelty audit.}
The post hoc literature comparison is correspondingly stronger here. Against the broader manifold-attention literature, \texttt{PoincareHC} is still literature-grounded rather than ex nihilo: hyperbolic attention and gyrovector-space papers already use Einstein midpoints, Lorentz centroids, or related intrinsic barycenters \cite{gulcehre2019hyperbolic,chen2022fully,yang2024hypformer,wang2025gyroatt}. What is new in this run is the placement of that aggregation primitive inside a \texttt{custom} hyper-connection operator that mixes residual streams under the fixed \texttt{hyper\_conn\_n=4} contract. Against the narrower direct HC line, the distinction is sharper. Published HC, mHC, mHC-lite, JPmHC, and spectral-HC variants still mostly constrain the residual mixing matrix itself \cite{zhu2024hyperconnections,xie2025mhc,mhclite2026,sengupta2026jpmhc,liu2026shc}; they do not directly provide the content-adaptive Poincar\'e stream router discovered here. The orthogonal branch should also not be oversold: Givens rotations, Householder reflections, and Cayley transforms are standard manifold parameterizations. But the run uses them in a way the surveyed attention and HC papers do not directly match, namely as content-adaptive stream-routing operators inside the hyper-connection module. This makes \texttt{GivensOrthogonalManifoldHC} and \texttt{HouseholderCayleyManifoldHC} strong novelty candidates, while \texttt{SinkhornDoublyStochasticHC} is correctly scored down as a repair that returns closer to the known HC / Birkhoff line \cite{sinkhorn1967concerning}. The final best node, \texttt{PoincareGivensHybridHC}, should therefore be read as recombinational novelty built on top of one stronger novelty event: width transport is delegated to the Givens branch, while the Poincar\'e branch survives as curvature-aware depth modulation. That decomposition is reviewer-valid and benchmark-effective, but the more fundamental discovery is the earlier \texttt{PoincareHC} node that first realized intrinsic manifold aggregation in the transformer search space.

\begin{remark}
The absolute validation cross-entropies in this transformer study should not be read as fully tuned language-model performance. Under the shared fixed recipe (10k-step AdamW training on small Shakespeare), several corrected seeds and generated candidates drive training loss low while still exhibiting comparatively high validation cross-entropy, which is an overfitting signature. In that regime, hyper-connection geometry matters precisely because it changes generalization: more flexible or poorly conditioned hyper-connections can fit the training stream quickly but transfer poorly to validation, whereas the better nodes in Table~\ref{tab:main-shortlist-transformer} appear to act as stronger inductive biases or implicit regularizers under the same optimizer and data budget.
\end{remark}
\FloatBarrier

\subsection{Optimizer discovery on a fixed transformer stack}
Optimization has evolved from stochastic approximation and SGD foundations \cite{robbins1951}, to adaptive methods such as Adam \cite{kingma2015adam}, decoupled-regularization variants such as AdamW \cite{loshchilov2019adamw}, and recent large-scale pretraining optimizer studies emphasizing rigorous evaluation protocols \cite{wen2025fantastic}. This task freezes the model, dataset, and benchmark recipe and asks CliffSearch to evolve only the optimizer update rule. The adapter runs nanoGPT train+eval \cite{nanogpt_repo} with plain residual attention only (\texttt{hyper\_conn\_type=none}, \texttt{hyper\_conn\_n=1}), aggregates validation loss across three benchmark random seeds, and uses mean validation loss as the lower-is-better primary metric, with failed benchmark seeds imputed by the worst successful loss. As in the transformer study, the benchmark hyperparameters are fixed across evolved optimizers rather than tuned per node, so the absolute validation losses are intentionally conservative and mainly useful for fair relative comparison under a shared training recipe. Search pressure is therefore isolated to optimization dynamics rather than architecture.

\paragraph{Task and benchmark setup.}
The optimizer results reported here come from four single-island all-Claude runs spanning two prompt bundles (\texttt{short\_json} and \texttt{workflow\_v2}) and two artifact modes (\texttt{code\_and\_theory} and \texttt{code\_only}). Each run used population size \(8\), three evolutionary generations after initialization, and the same three human seeds (\texttt{muon}, \texttt{adam}, \texttt{adamw}); because that seed set is smaller than the target population, generation 0 was completed by exploration-mutation descendants of those seeds. The benchmark itself was fixed across all four runs: Shakespeare character data, the small-model nanoGPT configuration, residual/plain attention only, three benchmark random seeds (\texttt{1337,2337,3337}), and \(10{,}000\) training iterations per benchmark seed; the exact composed model/data stack is listed in Appendix~\ref{app:shared-nanogpt-small-model}. Throughout this subsection, \emph{reviewer-valid} means that a node passed the reviewer threshold on both correctness and originality (the run-level audit uses this reviewer gate even when the stricter winner rule also includes the score-above-median requirement). Table~\ref{tab:main-shortlist-optimizer} reports the best reviewer-valid non-seed discovery from each run. As with the transformer study, these nanoGPT runs were benchmark-dominated in wall-clock time and remained in the single-digit US-dollar range on the Opus side for the full \(p=8\), \(g=3\) configuration. The appendix then carries the top-two discovered non-seed summary table (Table~\ref{tab:optimizer-discovered-nonseeds}), the four full run-level tables (Tables~\ref{tab:optimizer-short-tc-full}--\ref{tab:optimizer-wfv2-co-full}), the best discovered optimizer artifact card (Appendix~\ref{app:optimizer-best-artifact}), and the run-local task preambles in Appendix~\ref{app:task-preamble-optimizer-mhc-lite-theory-code} and Appendix~\ref{app:task-preamble-optimizer-mhc-lite-code-only}. These reported optimizer runs are also mirrored online with full run visualizations and best-node exports at \url{https://cliffsearch.ai}.

\begin{table}[!htbp]
\centering
\scriptsize
\makebox[\textwidth][c]{%
\begin{tabular}{ll>{\raggedright\arraybackslash}p{3.4cm}ccrrr}
\toprule
Prompt & Mode & Best discovered alias & Gen & Producer & \makecell[c]{Primary metric\\($\downarrow$)} & Corr & Orig\\
\midrule
\texttt{short\_json} & theory+code & \makecell[l]{MuonCausal\\Momentum} & 3 & mutation & 1.7782 & 4 & 4\\
\texttt{short\_json} & code-only & \makecell[l]{MuonSophiaV3\\CosGate} & 2 & mutation & 1.7659 & 4 & 4\\
\texttt{workflow\_v2} & theory+code & \makecell[l]{MuonCauchy\\Trust} & 2 & crossover & 2.8728 & 4 & 4\\
\texttt{workflow\_v2} & code-only & MuonSOAP & 2 & mutation & 3.7632 & 4 & 4\\
\bottomrule
\end{tabular}
}
\caption{Best reviewer-valid non-seed discoveries from the four real all-Claude optimizer-MHC-lite runs. Workflow-v2 still selected \texttt{SeedAdam} overall; the table isolates the discovered optimizer families rather than seed baselines.}
\label{tab:main-shortlist-optimizer}
\end{table}

\paragraph{Comparison axes.}
Table~\ref{tab:main-shortlist-optimizer} exposes two clean comparison axes. The first is prompt bundle: \texttt{short\_json} versus \texttt{workflow\_v2}. The second is artifact mode: \texttt{theory+code} versus \texttt{code-only}. The prompt-bundle difference is substantive rather than cosmetic. The short-json prompts are deliberately compact and leave each agent more freedom in how it interprets crossover, mutation, and review, whereas the workflow-v2 prompts prescribe a more explicit per-agent workflow and therefore a tighter review protocol. On the short-json bundle, both artifact modes produced real non-seed wins over \texttt{SeedAdam} (2.4420), and \texttt{code-only} was slightly better on the final metric (1.7659 versus 1.7782). On the workflow-v2 bundle, neither artifact mode produced a discovered optimizer whose reviewer-valid loss beat the Adam seed, but the two modes still surfaced different non-seed families and different reviewer behaviors.

\paragraph{Short-json discoveries.}
The short-json theory+code run followed a coherent Muon-centered discovery trajectory. A bootstrap exploration mutation already produced \texttt{MuonGrafting} (2.1356), which preserved Muon's Newton--Schulz orthogonalization path \cite{shah2025muon} but replaced cautious masking with gradient grafting and spectral-momentum decay. That family was strong enough to survive by elite carry, and the eventual winner \texttt{MuonCausalMomentum} (1.7782) emerged as a later exploration mutation from \texttt{MuonGraftFusion}. Its gains came from causal row-wise gradient-energy weighting, adaptive Newton--Schulz depth, and warmup-aware momentum. The same run also exposed a clear repair branch: \texttt{AdamW\_GPD\_AMR\_v2} repaired a real implementation bug in its parent and reached 1.9849, but the reviewer held it out at originality \(3/5\), correctly treating it as a repair-first improvement rather than a stronger new optimizer family.

The short-json code-only run reached the lowest discovered loss of the four-run set with \texttt{MuonSophiaV3\_CosGate} (1.7659). That node recombined Muon-style orthogonalization \cite{shah2025muon} with Sophia-like signal-quality control \cite{liu2023sophia}: cosine-similarity momentum gating, gradient-variance / signal-to-noise step scaling, and one extra Newton--Schulz iteration. Its strongest sibling, \texttt{MuonSOAPGradNormAdaptive} (2.2217), moved in a different direction by using adaptive Newton--Schulz depth and gradient-RMS clipping, a move closer to the Shampoo/SOAP line of adaptive preconditioning \cite{vyas2024soap}. In this bundle, \texttt{code-only} did not merely shorten artifacts; it shifted search pressure toward sharper code-level optimizer-control heuristics while still producing reviewer-valid discoveries.

\paragraph{Workflow-v2 review regime.}
The workflow-v2 runs were qualitatively different because the prompt bundle constrained the agents more tightly. Its prompts specify a more detailed workflow for pairing, crossover, mutation, and review, and the reviewer became far more explicit about prior-work overlap as a result. The review text repeatedly anchored originality judgments against known optimizer ingredients such as gradient centralization \cite{yong2020gradient}, AMSGrad-style stabilization \cite{reddi2018convergence}, and Riemannian/Stiefel optimization \cite{absil2008optimization}. In workflow-v2 theory+code, the best reviewer-valid discovery was \texttt{MuonCauchyTrust} (2.8728), a crossover that robustified Muon's 2D path with Cauchy pre-filtering before Newton--Schulz and replaced the non-2D Adam fallback with a Cauchy/stability-ratio variant. A lower-loss sibling, \texttt{MuonCauchyRiemannian} (2.5576), still received originality \(3/5\), which is exactly the sort of fact-checked novelty downgrade the workflow-v2 reviewer was designed to make. In workflow-v2 code-only, \texttt{MuonSOAP} (3.7632) and \texttt{AdaptiveOrthoAdam} (4.1883) survived review, while lower-loss repairs or literature-adjacent combinations such as \texttt{CautiousAdamGC\_v2} or \texttt{CorrectedAdamW} were explicitly scored down on originality. The same run also recorded a useful hard failure: \texttt{AdamW\_GradCentral\_v3} still returned benchmark error because the attempted fix subclassed \texttt{AdamW} directly and violated the benchmark's strict \texttt{EvoOptimizer} inheritance contract.

\paragraph{Takeaway.}
Taken together, the optimizer case study shows two complementary regimes. Short-json is better at surfacing lower-loss optimizer discoveries, especially when the loop is allowed to mutate Muon into more aggressive control laws. Workflow-v2 is less forgiving on novelty, but that is precisely what makes its reviews scientifically useful: they do not merely rank loss, they audit whether a node is a real optimizer discovery, a repair, or a recombination of already-published ingredients. The artifact-mode comparison shows the same pattern at a finer grain: \texttt{code-only} can be slightly sharper on raw optimizer heuristics, while \texttt{theory+code} more often yields cleaner, more interpretable optimizer families.

\subsection{Native optimizer ablation}
The native-optimizer study serves a different role from the two nanoGPT-based studies above. It is not the headline discovery benchmark; it is an ablation task that keeps the CliffSearch loop, reviewer gating, and artifact-mode split intact while replacing the expensive nanoGPT stack with small native supervised-learning problems. The benchmark uses four classification datasets: two synthetic linear tasks (\texttt{syn\_clf\_balanced\_linear}, \texttt{syn\_clf\_noisy\_imb\_linear}) and two tabular MLP tasks (\texttt{tab\_breast\_cancer\_mlp}, \texttt{tab\_wine\_mlp}, with hidden width 64). Every candidate optimizer is evaluated over the same 32-run bundle: those four datasets, benchmark seeds \(\{0,1\}\), and a \(2\times2\) learning-rate/weight-decay grid with learning rates \(\{3\times10^{-4}, 10^{-3}\}\) and weight decays \(\{0, 10^{-4}\}\), with six training epochs per run. The stored primary metric is mean validation loss across that fixed evaluation bundle; when some runs fail, the aggregate uses worst-successful-loss imputation for the missing runs. As with the nanoGPT tasks, these benchmark settings are fixed across evolved optimizers rather than tuned per node, so the absolute losses are intentionally conservative and mainly support fair relative comparison under a shared evaluation bundle. The exact run-local task preambles are reproduced in Appendix~\ref{app:task-preamble-optimizer-native-theory-code}--\ref{app:task-preamble-optimizer-native-code-only}. We audited all eight real native-optimizer wrapper runs (464 total nodes); the full run matrix and aggregate statistics are reported in Appendix~\ref{app:native-optimizer-ablation}, Tables~\ref{tab:native-optimizer-run-matrix}--\ref{tab:native-optimizer-heldout}. Those ablation runs are also mirrored online, including their full run visualizations and exported best nodes, at \url{https://cliffsearch.ai}.

\begin{table}[!htbp]
\centering
\scriptsize
\setlength{\tabcolsep}{3pt}
\resizebox{\textwidth}{!}{%
\begin{tabular}{>{\raggedright\arraybackslash}p{1.5cm}>{\raggedright\arraybackslash}p{1.7cm}>{\raggedright\arraybackslash}p{2.0cm}>{\raggedright\arraybackslash}p{2.0cm}>{\raggedright\arraybackslash}p{2.8cm}r}
\toprule
Prompt & Mode & Evol. params & Best seed & Best reviewer-valid non-seed & \makecell[c]{Valid non-seeds\\beating seed}\\
\midrule
\texttt{short\_json} & theory+code & \makecell[l]{g3/p8,\\ac=F, hs5=T} & \makecell[l]{SeedAdam\\0.7291} & \makecell[l]{HyperbolicAdaptive\\0.7120} & 2\\
\texttt{short\_json} & theory+code & \makecell[l]{g6/p12,\\ac=F, hs5=T} & \makecell[l]{SeedAdam\\0.7054} & \makecell[l]{SpecProjAdam\\0.5348} & 4\\
\texttt{short\_json} & code-only & \makecell[l]{g3/p8,\\ac=F, hs5=T} & \makecell[l]{SeedAdam\\0.7030} & \makecell[l]{AGNSoftCautious\\CurvAdamW\\0.7046} & 0\\
\texttt{short\_json} & code-only & \makecell[l]{g6/p12,\\ac=F, hs5=T} & \makecell[l]{SeedAdamW\\0.7040} & \makecell[l]{CautiousAdamProj\\0.4261} & 10\\
\texttt{workflow\_v2} & theory+code & \makecell[l]{g3/p8,\\ac=F, hs5=T} & \makecell[l]{SeedAdamW\\0.7193} & none reviewer-valid & 0\\
\texttt{workflow\_v2} & theory+code & \makecell[l]{g6/p12,\\ac=T, hs5=F} & \makecell[l]{SeedAdamW\\0.7212} & \makecell[l]{CauAdam\\0.7186} & 3\\
\texttt{workflow\_v2} & code-only & \makecell[l]{g3/p8,\\ac=F, hs5=T} & \makecell[l]{SeedAdam\\0.7149} & \makecell[l]{HyperbolicAGNAdam\\0.6403} & 2\\
\texttt{workflow\_v2} & code-only & \makecell[l]{g6/p12,\\ac=T, hs5=F} & \makecell[l]{SeedAdam\\0.7174} & \makecell[l]{CurvAdam\_GradCentral\_SWA\\0.5377} & 9\\
\bottomrule
\end{tabular}
}
\caption{Native-optimizer ablation matrix. All runs keep \texttt{review\_generation\_zero=true}; the varying evolution controls are generations/population, \texttt{augment\_crossover} (ac), and \texttt{human\_seed\_all\_5} (hs5). Reported discoveries are non-seed nodes only.}
\label{tab:main-native-optimizer}
\end{table}

\paragraph{Artifact-mode comparison.}
Across the four paired comparisons, \texttt{code-only} is the stronger native-optimizer discovery mode. It averages 18.25 reviewer-valid non-seeds per run, versus 8.5 for \texttt{theory+code}, and its best reviewer-valid discovery is \texttt{CautiousAdamProj} at 0.4261, well below the best \texttt{theory+code} discovery \texttt{SpecProjAdam} at 0.5348. The gap is not merely one lucky node: \texttt{CautiousAdamProj} reappears six times in the short-json \texttt{code-only} g6/p12 run, and \texttt{CurvAdam\_GradCentral\_SWA} reappears five times in the workflow-v2 \texttt{code-only} g6/p12 run. By contrast, \texttt{theory+code} more often surfaces geometric or spectral variants such as \texttt{HyperbolicAdaptive}, \texttt{SpecProjAdam}, and \texttt{GradHarm\_Adam}; these are interpretable, but fewer survive reviewer gating as reviewer-valid seed-beating optimizers.

\paragraph{Prompt-bundle comparison.}
The prompt bundle matters as much as the artifact mode. Averaged across both modes and both evolution settings, \texttt{short\_json} yields 17.5 reviewer-valid non-seeds per run, compared with 9.25 for \texttt{workflow\_v2}. The short bundle therefore remains the more productive optimizer-discovery regime. The workflow-v2 bundle, however, reveals a different phenomenon: it still generates many low-loss candidates, but its reviewer suppresses a larger fraction of them after fact-checking novelty and correctness. The clearest example is \texttt{HyperbolicMomentum} in the larger workflow-v2 theory+code run: it achieves the best raw non-seed loss in that run (0.3879) but is held out because the reviewer rejects its first-moment bias correction under anti-windup momentum as not yet correct enough. This is exactly the behavior a stricter reviewer is supposed to have.

\paragraph{Evolution-parameter comparison.}
Only the short bundle provides a clean evolution-parameter comparison, because its g3/p8 and g6/p12 runs keep \texttt{augment\_crossover=false} and \texttt{human\_seed\_all\_5=true} fixed. Under that clean comparison, increasing population and generations substantially improves discovery depth: reviewer-valid non-seeds grow from 6 to 24 in theory+code and from 8 to 32 in code-only, while the best reviewer-valid non-seed improves from 0.7120 to 0.5348 in theory+code and from 0.7046 to 0.4261 in code-only. The workflow-v2 g6/p12 runs should be read more cautiously, because they change three things at once relative to workflow-v2 g3/p8: the run is longer and larger, \texttt{augment\_crossover} is enabled, and \texttt{human\_seed\_all\_5} is disabled. The resulting comparison is still interesting, but it is no longer a clean single-parameter ablation.

\paragraph{Discovery patterns and review behavior.}
The strongest repeated native discoveries are mostly Adam-family control laws rather than Muon-family variants. High-performing reviewer-valid families include \texttt{CautiousAdamProj}, \texttt{CurvAdam\_GradCentral\_SWA}, \texttt{SpecProjAdam}, and \texttt{HyperbolicAGNAdam}. That contrast with the Shakespeare transformer-stack optimizer study is itself informative. There, Muon-derived discoveries were among the strongest because the benchmark still involved structured transformer pretraining, large 2D weight updates, and Newton--Schulz-style orthogonalization \cite{shah2025muon}. In the native ablation, by contrast, the benchmark shifts to small linear/MLP classification tasks, where lightweight Adam-style control heuristics are more useful than expensive matrix-orthogonalization machinery. Their mechanisms are therefore combinations of cautious masking, gradient projection, gradient centralization \cite{yong2020gradient}, adaptive clipping, simple curvature surrogates, and occasional geometric reweightings. Workflow-v2 reviews are especially informative here because they explicitly anchor originality against known ingredients such as gradient centralization \cite{yong2020gradient}, AMSGrad-style stabilization \cite{reddi2018convergence}, and Riemannian/Stiefel optimization \cite{absil2008optimization}. In other words, the native task confirms the same division of labor seen elsewhere in the paper: short prompts give the agents more room to propose aggressive optimizer heuristics, while workflow-v2 gives the reviewer more leverage to distinguish real discoveries from repairs or literature-adjacent recombinations.


\section{Methods}
\noindent In this section, we provide a detailed account of the CliffSearch runtime: execution assumptions, generation mechanics, and the task-specific benchmark instantiations used in the empirical studies.

\subsection{Generic Runtime and Execution Model}
\paragraph*{Execution assumptions and system model.}
Experiments assume access to compute nodes with \(M\) CPU workers and \(G\) GPUs per node. CPU workers service a bounded SDK-call queue (pair selector, crossover, exploration mutation, correction mutation, reviewer). These SDK calls are text-only: agent invocations do not receive tools for arbitrary code execution, shell access, or benchmark-side file mutation. Benchmarks are dispatched through a separate bounded benchmark queue and may execute on CPU or GPU depending on benchmark mode/config and slot assignment. In single-island mode, both queues run on one compute node under local worker pools. In multi-island mode, each island is pinned to one compute node; islands evolve independently and coordinate via shared-disk state (\texttt{outbox}/\texttt{inbox}/\texttt{state}) where migration packets are written/read and orchestrator routing/deduplication is applied.

\paragraph*{Population update and winner gating.}
At generation $g$, benchmark and reviewer outputs are computed for each node in $P_g$. Winners are defined by correctness, originality and score-above-median gating. Pair selection receives summary-only winner views, after which runtime sanitization enforces valid ids, disjointness, and pair-count limits. Initialization follows the same fixed-size principle: generation 0 inserts the configured human seeds first, and if that set is smaller than the target population size, the remaining slots are produced by exploration mutation from the seed pool before the first benchmark/review cycle.

\paragraph*{Optional evolution-control flags.}
Three configuration flags are especially important when interpreting reported runs. First, \texttt{review\_generation\_zero} controls whether generation-0 nodes are sent through the reviewer at all; when it is disabled, the seed/bootstrap population still benchmarks but does not receive reviewer scores until later generations. Second, \texttt{human\_seed\_all\_5} only matters when generation-0 review is enabled: true human seeds (not their exploration-filled descendants) keep their reviewer narrative but their effective correctness and originality scores are forced to \(5/5\), so the initial human reference set is not discarded by reviewer conservatism. Third, \texttt{augment\_crossover} changes how parent pools are built for pairing. The default strict pool is the current reviewer-valid winner set; when augmentation is enabled and that strict pool is too small to sustain useful pairing, runtime augments it with the best remaining scored nodes in directional-score order so crossover can still operate instead of collapsing to mutation-only update. These flags therefore do not change the benchmark itself; they change where reviewer gating starts, how human seeds are protected, and how much parent diversity crossover is allowed to see.

\paragraph*{Operator budgeting and fixed-size closure.}
Let target population size be \(N\), quota percentages be \((p_e,p_c,p_m)\) for elite/crossover/mutation with \(p_e+p_c+p_m=1\), and winner set be \(W_g\). Raw shares are
\[
\mathbf{a}=N\cdot(p_e,p_c,p_m).
\]
Integer operator targets are obtained by largest-remainder rounding
\[
(N_e,N_c,N_m)=\mathrm{LRRound}(\mathbf{a}),\qquad N_e+N_c+N_m=N.
\]
If winners exist and a minimum elite floor \(E_{\min}\) is configured, elite budget is raised by borrowing from mutation first and crossover second:
\[
\delta_e=\max(0,E_{\min}-N_e),
\]
\[
N_e' = N_e+\delta_e,\quad
N_m' = \max(0,N_m-\delta_e),\quad
N_c' = \max\!\left(0,\;N_c-\max(0,\delta_e-N_m)\right).
\]
If crossover underproduces, shortfall is transferred to mutation:
\[
N_c^{\mathrm{short}}=\max(0,N_c'-N_c^{\mathrm{act}}),\qquad
N_m^{\mathrm{target}}=N_m'+N_c^{\mathrm{short}}.
\]
After mutation and elite-copy realization, let realized counts be \(N_m^{\mathrm{act}}\) and \(N_e^{\mathrm{act}}=\min(N_e',|W_g|)\). Backfill count is then
\[
N_{\mathrm{fill}}=\max\!\left(0,\;N-\big(N_c^{\mathrm{act}}+N_m^{\mathrm{act}}+N_e^{\mathrm{act}}\big)\right),
\]
and next-population size satisfies
\[
|P_{g+1}|=N_c^{\mathrm{act}}+N_m^{\mathrm{act}}+N_e^{\mathrm{act}}+N_{\mathrm{fill}}=N.
\]
Backfill first attempts exploration-mutation generation from the source pool (winners if available, else previous population) and falls back to content copy if mutation invocation fails, ensuring fixed-size closure.

\paragraph*{Agent I/O schemas and validation checks.}
The pair-selector interface is summary-restricted: input is a set of winner records carrying \texttt{id}, \texttt{summary\_md}, score, and review flags, and output is a bounded list of parent-id pairs. Pair outputs are then sanitized by deterministic checks (membership in winner set, no self-pairing, disjointness policy, and configured pair cap).

The crossover and mutation interfaces are full-node interfaces. Input contains canonical node content and task context; output must include the canonical triplet \texttt{summary\_md}, \texttt{theory\_content}, and \texttt{code\_content}. Runtime normalization converts provider-specific formatting into this canonical schema, and schema failures trigger retry-or-fail logic rather than permissive coercion.

The reviewer interface receives an evaluated node including \texttt{summary\_md}, \texttt{theory\_content}, \texttt{code\_content}, benchmark summary, and lineage metadata (for example parent ids, operator provenance, and parent benchmark/review snapshots when present). It outputs \texttt{correctness\_score}, \texttt{originality\_score}, and reviewer narrative. Reviewer rubrics are applied across both theory and code, including consistency checks between theoretical claims and implementation behavior, while benchmark and parent metadata provide empirical and genealogical context for assessing improvement over parent nodes. Winner gating consumes these outputs directly; there is no hidden heuristic path bypassing reviewer signals.

\paragraph*{Artifact mode and invariant schema.}
Runtime config includes\\ \texttt{artifact\_mode}\(\in\)\{\texttt{code\_and\_theory}, \texttt{code\_only}\}. In \texttt{code\_and\_theory}, seeds and generated nodes are required to provide non-empty \texttt{theory\_content} and \texttt{code\_content}. In \texttt{code\_only}, \texttt{theory\_content} is normalized to an empty string while preserving the same node schema and on-disk structure. This design keeps downstream tooling (storage, visualization, migration, and replay) mode-independent.

Reviewer context is adapted by mode but winner logic is unchanged. In \texttt{code\_and\_theory}, reviewer prompts evaluate code and theory jointly with benchmark and lineage context. In \texttt{code\_only}, separate reviewer prompts instruct the agent to review the code itself using benchmark and lineage context, to use \texttt{summary\_md} only as secondary evidence, and to ignore \texttt{theory\_content}. Correctness and originality thresholds remain hard gates in both modes.

\paragraph*{Benchmark and reviewer integration.}
Each candidate in $P_{g+1}$ runs benchmark evaluation under a strict metrics schema (\texttt{primary\_metric}, \texttt{metric\_name}, \texttt{higher\_is\_better}, summary, details, artifacts). The generated code artifact is not executed during the SDK agent calls themselves; instead, runtime injects the node's code artifact into the task-specific benchmark path only after schema normalization, task-grounding checks, and any task-specific contract validation (for example import checks, AST checks, shape checks, or custom hyper-connection checks). Runtime then computes directional score $s$ from $(m,\texttt{higher\_is\_better})$ before ranking. Contract checks verify benchmark payload completeness and explicit direction before any ranking step. Reviewer outputs are accepted only if their schema is complete and score fields are parseable; otherwise the node is marked non-eligible for winner status. Valid benchmark and review payloads are merged into canonical node summaries before persistence. This separation makes runs cheaper because most agent calls remain lightweight text generations rather than tool-enabled execution sessions, and safer because candidate code is executed only inside the benchmark adapter after explicit validation rather than inside the open-ended agent loop.

\paragraph*{Task grounding and contract enforcement.}
Each agent call includes \texttt{task\_type}, \texttt{task\_preamble}, and runtime \texttt{task\_grounding}. For custom tasks, \texttt{task\_preamble} is the canonical contract surface. First-class task types can add runtime-specific normalizers and validators before benchmark execution (for example class-shape checks, import checks, or AST-level safety constraints). Nodes that violate grounding constraints fail fast prior to expensive benchmark runs. Exact workflow-v2 prompt templates used by operators are provided in Appendix~\ref{app:workflowv2}.

\paragraph*{Persistence and migration protocol.}
Each generation writes node-level artifacts and a generation-local snapshot. Concretely, it persists \texttt{population.json} and generation-local \texttt{ga\_data.json}, and also extends cumulative run-level \texttt{ga\_data.json}. In distributed mode, migrants are exported as packets containing packet id, source/target island, source node metadata, and score fields. Orchestrator-side dedupe ensures single import semantics.

\subsection{Task-Specific Benchmark Realizations}
\paragraph*{Shared random-seed aggregation for single-objective optimizer and transformer tasks.}
Let $S_{\mathrm{rand}}$ be configured benchmark random seeds, and let $p_s$ be the random-seed-level primary objective. If random seed $s$ fails and at least one random seed succeeded, code imputes the failed seed with worst successful objective:
\[
\tilde p_s =
\begin{cases}
p_s, & s \in S_{\mathrm{rand,ok}}\\
\max_{j \in S_{\mathrm{rand,ok}}} p_j, & s \notin S_{\mathrm{rand,ok}}
\end{cases}
\]
and then uses
\[
\bar p = \frac{1}{|S_{\mathrm{rand}}|}\sum_{s\in S_{\mathrm{rand}}}\tilde p_s.
\]
If $S_{\mathrm{rand,ok}}=\varnothing$, the adapter returns benchmark error (no imputation baseline exists). Reported primary metric is
\[
\!m_{\mathrm{bench}}=\bar p,
\]
and directional score is
\[
s=
\begin{cases}
m_{\mathrm{bench}}, & \texttt{higher\_is\_better}=\mathrm{true}\\
-m_{\mathrm{bench}}, & \texttt{higher\_is\_better}=\mathrm{false}.
\end{cases}
\]
Error and log information is preserved for reviewer context: adapter failures are serialized into benchmark payloads (for example \texttt{details.error}, failed-seed error summaries, and bounded stdout/stderr excerpts when available), and runtime benchmark exceptions are also recorded in node metadata (for example \texttt{benchmark\_error}). The reviewer stage consumes this benchmark+metadata context together with theory/code artifacts; lineage metadata includes parent benchmark/review context so the reviewer can judge whether a child improves over its parents.

\paragraph*{Optimizer discovery benchmark (fixed model/data).}
Candidate code must satisfy optimizer runtime contract and be dynamically importable. Each benchmark random seed produces validation loss $L_s$ from fixed nanoGPT train/eval \cite{nanogpt_repo} with plain/regular attention (\texttt{hyper\_conn\_type=none}); here $p_s=L_s$, so $m$ is the imputed mean validation loss defined above. The optimizer benchmark uses a fixed training recipe shared across nodes rather than per-node hyperparameter tuning, so reported losses are conservative absolute values intended for relative comparison.

\paragraph*{Native optimizer ablation benchmark.}
Candidate code must satisfy the same optimizer runtime contract, but the benchmark mode is \texttt{pytorch\_optimizer} rather than nanoGPT. Each candidate is evaluated on a fixed suite of small classification tasks (synthetic linear and tabular MLP settings), over 32 benchmark runs total: four tasks, two benchmark random seeds, and a \(2\times2\) learning-rate/weight-decay grid (four hyperparameter settings), with six training epochs per run. The stored primary metric is \texttt{mean\_val\_loss}, i.e. the mean validation loss aggregated across that fixed native evaluation bundle, with worst-successful-loss imputation when some runs fail. Because the benchmark sweep is shared across nodes rather than retuned per node, these native-optimizer losses are conservative absolute values meant for relative comparison. Reported native-optimizer scores in the ablation section therefore compare update rules under a shared low-cost supervised-learning test bed rather than under the Shakespeare nanoGPT stack.

\paragraph*{Transformer hyper-connection evolution benchmark.}
Candidate code must pass custom hyper-connection checks before execution. For each benchmark random seed, the primary objective is validation loss \(L_s\) on nanoGPT train/eval with attention-level hyper-connections enabled \cite{nanogpt_repo,mhclite_repo}. Hence \(p_s=L_s\). Seed aggregation and failed-seed handling then follow the shared equations above. The transformer benchmark also uses a fixed shared training recipe rather than per-node hyperparameter tuning, so reported losses should be read as conservative relative-comparison values.

\paragraph*{Benchmark primary-metric formulas by task.}
Table~\ref{tab:fitness-by-task} summarizes the exact benchmark primary-metric definitions written to node payloads and consumed by ranking/selection/visualization.

\begin{table}[t]
\centering
\scriptsize
\begin{tabular}{p{3.0cm} p{5.2cm} p{2.6cm} p{3.2cm}}
\toprule
Task & Stored benchmark primary metric $m_{\mathrm{bench}}$ & Directional score $s$ & Failure handling \\
\midrule
Native optimizer (pytorch\_optimizer) &
Mean validation loss aggregated across the configured native classification tasks, benchmark random seeds, and small hyperparameter sweep; stored field is \texttt{mean\_val\_loss}. &
Reported setting: \texttt{higher\_is\_better=false}, hence \(s=-m_{\mathrm{bench}}\). &
If all native benchmark evaluations fail: benchmark error.
Otherwise mean is taken over the successful fixed evaluation bundle returned by the adapter. \\
\midrule
Optimizer (MHC-lite adapter) &
\(p_s=L_s\). With random-seed imputation over \(S_{\mathrm{rand}}\),
\(m_{\mathrm{bench}}=\frac{1}{|S_{\mathrm{rand}}|}\sum_{s\in S_{\mathrm{rand}}}\tilde p_s\). &
Reported setting: \texttt{higher\_is\_better=false}, hence \(s=-m_{\mathrm{bench}}\). &
If all seeds fail: benchmark error.
Else impute failed seeds with worst successful primary value. \\
\midrule
Transformer hyper-connection (MHC-lite adapter) &
\(p_s=L_s\) (validation loss under custom hyper-connections). Same random-seed imputation/aggregation as optimizer:
\(m_{\mathrm{bench}}=\frac{1}{|S_{\mathrm{rand}}|}\sum_{s\in S_{\mathrm{rand}}}\tilde p_s\). &
Reported setting: \texttt{higher\_is\_better=false}, hence \(s=-m_{\mathrm{bench}}\). &
If all seeds fail: benchmark error.
Else impute failed seeds with worst successful primary value. \\
\bottomrule
\end{tabular}
\caption{Per-task benchmark primary-metric definitions using unified notation \((m_{\mathrm{bench}}, s)\). Stored field is \texttt{benchmark.primary\_metric} (legacy alias \texttt{benchmark.fitness}); ranking and winner gating use directional score \(s\).}
\label{tab:fitness-by-task}
\end{table}

\section{Conclusion}
This work frames scientific algorithm discovery as an evolutionary process over structured artifacts rather than code snippets alone. At the conceptual level, CliffSearch couples theory and implementation in each node, so proposed mechanisms can be judged as scientific claims with executable evidence. At the decision level, reviewer outputs (correctness and originality) are treated as first-class selection constraints, which prevents benchmark score from being the only gate.

At the search-policy level, CliffSearch separates exploration mutation (novelty through adjacent-domain transfer) from correction mutation (evidence-guided repair), giving the loop explicit mechanisms for both discovery and stabilization. At the systems level, the same runtime can instantiate multiple user-defined tasks as long as task, benchmark, and metric are specified; the transformer study, the matched transformer \texttt{code\_only} breakthrough, the optimizer-on-nanoGPT study, and the native-optimizer ablation in this paper are illustrative realizations of that general framework. The transformer \texttt{code\_only} result is especially informative because it shows that this mode still carries ideation: \texttt{summary\_md} preserves design principles even when \texttt{theory\_content} is empty, and the resulting loop can still discover reviewer-valid manifold operators rather than merely generate implementation variants. The absence of a separate theory file also frees more of the agent context budget for the code artifact itself. In that run, CliffSearch discovers two genuine geometric families, a Givens width-transport family and a Poincar\'e hyperbolic family, and then combines them into a real width/depth hybrid rather than a superficial alias remix. That combined Givens/Poincar\'e breakthrough is one of the most salient discoveries in the work.

Several limitations remain. First, the framework does not by itself provide convergence guarantees for agent-guided operators. Second, reviewer uncertainty is currently represented through scalar scores rather than calibrated uncertainty models. Third, cross-task statistical comparability still depends on benchmark protocol quality. Fourth, novelty judgment is still only weakly grounded: our post hoc audits show that prompt-only reviewer judgments are useful, but a stronger system should anchor reviewer originality analysis to a retrieval-backed literature database or RAG-style survey layer over relevant prior work, so that novelty claims are explicitly tied to searchable evidence rather than latent model memory alone. That direction is also consistent with recent arguments that novelty and reasoning claims need external, searchable reference frames to be scientifically refutable \cite{mossel2026refutabilitygap}. These limitations define clear next steps for rigorous comparative studies.

\section*{AI Assistance}
AI systems were used to assist in editing this manuscript and in developing the supporting codebase. CliffSearch's system design and architecture, as well as  benchmarking and experimentation were fully performed by the authors.

\bibliographystyle{plainurl}
\bibliography{references}

\clearpage
\appendix
\section{Extended Results}
\label{app:extended-results}
\subsection{Transformer theory+code run appendix export}
\label{app:transformer-single-island}
This appendix block reports the concrete all-Claude theory+code transformer run discussed in Section~4. The run used the fixed \texttt{hyper\_conn\_n=4} contract that was active for that experiment. It includes the real-run full transformer \texttt{theory+code} node table together with the exported best node \texttt{GrassmannianSubspaceRouting} (node \texttt{H2} in the flat run view, internal id \texttt{g002\_n0024\_66b987}).
\paragraph{Per-task full node tables and shortlisted candidates.}
For each task, tables include all nodes across generations. Shortlisted nodes are highlighted in bold and marked in column \texttt{S}; best/tied shortlisted rows are highlighted in blue.

\subsubsection*{Transformer theory+code run}
\textbf{Selection}: lower is better ($\downarrow$), pool=\texttt{reviewer\_valid}, reviewer-valid=13, finite-score=24, total=32.\\
\textbf{Metric(s)}: \texttt{mean\_val\_loss}; \textbf{Mode(s)}: \texttt{mhc\_lite\_attention}.

\begingroup
\scriptsize
\setlength{\LTleft}{0pt}
\setlength{\LTright}{0pt}
\begin{longtable}{ccc>{\raggedright\arraybackslash}p{4.2cm}rrrr}
\caption{All nodes for the transformer theory+code run. Primary metric direction $\downarrow$. Ranking/shortlist uses directional score. Column S marks shortlisted nodes.}\label{tab:transformer-full-node-table}\\
\toprule
S & Node & Gen & Alias & Primary metric & Score & Corr & Orig\\
\midrule
\endfirsthead
\toprule
S & Node & Gen & Alias & Primary metric & Score & Corr & Orig\\
\midrule
\endhead
\midrule
\multicolumn{8}{r}{\footnotesize Continued on next page}\\
\endfoot
\bottomrule
\endlastfoot
 & A0 & 0 & \makecell[l]{Transformer\\Residual\\AttentionSeed} & 4.8555 & -4.8555 & 5 & 1\\
 & B0 & 0 & \makecell[l]{MHCLiteAttention\\Seed} & \textcolor{red!80!black}{\textbf{ERROR(code)}} & \textcolor{red!80!black}{\textbf{ERROR(code)}} & 2 & 2\\
 & C0 & 0 & HCAttentionSeed & \textcolor{red!80!black}{\textbf{ERROR(code)}} & \textcolor{red!80!black}{\textbf{ERROR(code)}} & 2 & 1\\
 & D0 & 0 & \makecell[l]{Hyperbolic\\PoincareRouting} & 4.4953 & -4.4953 & 3 & 4\\
 & E0 & 0 & \makecell[l]{Hyperbolic\\RotationRouting} & \textcolor{red!80!black}{\textbf{ERROR(code)}} & \textcolor{red!80!black}{\textbf{ERROR(code)}} & 2 & 4\\
 & F0 & 0 & HypExpRouteV1 & 5.4568 & -5.4568 & 4 & 3\\
 & G0 & 0 & \makecell[l]{Grassmannian\\SubspaceRouting} & \textcolor{red!80!black}{\textbf{ERROR(code)}} & \textcolor{red!80!black}{\textbf{ERROR(code)}} & 2 & 4\\
 & H0 & 0 & \makecell[l]{Grassmannian\\SubspaceRouting} & \textcolor{red!80!black}{\textbf{ERROR(code)}} & \textcolor{red!80!black}{\textbf{ERROR(code)}} & 2 & 4\\
 & A1 & 1 & \makecell[l]{HyperbolicExpMap\\Routing} & 5.12477 & -5.12477 & 4 & 4\\
 & B1 & 1 & MHCLiteDtypeFix & 5.53547 & -5.53547 & 4 & 1\\
 & C1 & 1 & \makecell[l]{HCAttentionSeed\\DtypeFix} & 5.7792 & -5.7792 & 4 & 1\\
 & D1 & 1 & \makecell[l]{Hyperbolic\\PoincareRoutingV\\2} & 4.16457 & -4.16457 & 4 & 4\\
\textbf{*} & \textbf{E1} & \textbf{1} & \textbf{\makecell[l]{Hyperbolic\\RotationRouting}} & \textbf{1.84487} & \textbf{-1.84487} & \textbf{4} & \textbf{4}\\
 & F1 & 1 & GrassRouteV1 & \textcolor{red!80!black}{\textbf{ERROR(code)}} & \textcolor{red!80!black}{\textbf{ERROR(code)}} & 2 & 4\\
 & G1 & 1 & \makecell[l]{Grassmannian\\SubspaceRouting} & 5.00553 & -5.00553 & 4 & 4\\
 & H1 & 1 & \makecell[l]{Grassmannian\\SubspaceRouting} & 5.52453 & -5.52453 & 5 & 3\\
\textbf{*} & \textbf{A2} & \textbf{2} & \textbf{\makecell[l]{GivensHyperbolic\\Routing}} & \textbf{1.7683} & \textbf{-1.7683} & \textbf{4} & \textbf{4}\\
 & B2 & 2 & \makecell[l]{Hyperbolic\\Grassmannian\\HybridRouting} & 5.33363 & -5.33363 & 4 & 3\\
 & C2 & 2 & \makecell[l]{Hyperbolic\\TangentBundle\\Routing} & 5.44893 & -5.44893 & 4 & 4\\
 & D2 & 2 & \makecell[l]{HyperbolicExpMap\\Routing} & 5.80177 & -5.80177 & 4 & 4\\
 & E2 & 2 & GrassRouteV2 & 5.2936 & -5.2936 & 4 & 3\\
 & F2 & 2 & \makecell[l]{StiefelFrame\\Routing} & 5.58133 & -5.58133 & 4 & 4\\
\textbf{*} & \textbf{G2} & \textbf{2} & \textbf{\makecell[l]{Hyperbolic\\RotationRouting}} & \textbf{1.84487} & \textbf{-1.84487} & \textbf{4} & \textbf{4}\\
\textcolor{blue!75!black}{\textbf{*}} & \textcolor{blue!75!black}{\textbf{H2}} & \textcolor{blue!75!black}{\textbf{2}} & \textcolor{blue!75!black}{\textbf{\makecell[l]{Grassmannian\\SubspaceRouting}}} & \textcolor{blue!75!black}{\textbf{1.69347}} & \textcolor{blue!75!black}{\textbf{-1.69347}} & \textcolor{blue!75!black}{\textbf{4}} & \textcolor{blue!75!black}{\textbf{4}}\\
 & A3 & 3 & \makecell[l]{Grassmannian\\Hyperbolic\\Routing} & 1.6983 & -1.6983 & 4 & 3\\
 & B3 & 3 & \makecell[l]{SpectralCayley\\Orthogonal\\Routing} & \textcolor{red!80!black}{\textbf{ERROR(code)}} & \textcolor{red!80!black}{\textbf{ERROR(code)}} & 2 & 4\\
 & C3 & 3 & \makecell[l]{Hyperbolic\\TangentBundle\\RoutingV2} & 5.2944 & -5.2944 & 4 & 3\\
 & D3 & 3 & \makecell[l]{HyperbolicExpMap\\RoutingV2} & 5.47507 & -5.47507 & 4 & 3\\
 & E3 & 3 & PoincareRoute & 5.38707 & -5.38707 & 4 & 4\\
 & F3 & 3 & \makecell[l]{StiefelFrame\\RouteV2} & 5.21987 & -5.21987 & 4 & 4\\
\textcolor{blue!75!black}{\textbf{*}} & \textcolor{blue!75!black}{\textbf{G3}} & \textcolor{blue!75!black}{\textbf{3}} & \textcolor{blue!75!black}{\textbf{\makecell[l]{Grassmannian\\SubspaceRouting}}} & \textcolor{blue!75!black}{\textbf{1.69347}} & \textcolor{blue!75!black}{\textbf{-1.69347}} & \textcolor{blue!75!black}{\textbf{4}} & \textcolor{blue!75!black}{\textbf{4}}\\
 & H3 & 3 & \makecell[l]{Poincare\\Hyperbolic\\Routing} & \textcolor{red!80!black}{\textbf{ERROR(code)}} & \textcolor{red!80!black}{\textbf{ERROR(code)}} & 2 & 4\\
\end{longtable}
\endgroup

\paragraph{Generation-0 seed inventory.}
For each task run, we enumerate configured human seeds from generation 0 (loaded from run-local \texttt{config.snapshot.json}).

\subsubsection*{Transformer theory+code run}
\begin{tcolorbox}[breakable,colback=blue!2,colframe=blue!60!black,title={Generation-0 human seeds}]
\textit{No seed entries found in config snapshot.}
\end{tcolorbox}

\paragraph{Best-node artifact card.}
For each task, we render artifacts for the first shortlisted node (highest directional-score candidate under the configured selection rule).

\subsubsection*{Transformer theory+code run}
\begin{tcolorbox}[breakable,colback=gray!2,colframe=black!40,title={Best node metadata}]
\begin{itemize}
  \item Method alias: \texttt{GrassmannianSubspaceRouting}
  \item Generation: 2
  \item Primary metric ($\downarrow$): 1.69347
  \item Directional score (ranked): -1.69347
  \item Direction: lower is better ($\downarrow$)
  \item Metric: \texttt{mean\_val\_loss}
  \item Benchmark mode: \texttt{mhc\_lite\_attention}
  \item Task type: \texttt{unknown}
  \item Parents: \texttt{g001\_n0013\_23cb7e}
  \item Artifact producer: \texttt{Exploration Mutation Agent}
  \item Reviewer scores (corr/orig): 4/4
\end{itemize}
\end{tcolorbox}

\begin{tcolorbox}[breakable,colback=blue!2,colframe=blue!60!black,title={Task preamble (task grounding used for this run)}]
\begin{lstlisting}[basicstyle=\ttfamily\scriptsize,breaklines=true,columns=fullflexible]
[task_preamble missing in config snapshot]
\end{lstlisting}
\end{tcolorbox}

\begin{tcolorbox}[breakable,colback=orange!4,colframe=orange!75!black,title={summary\_md excerpt (produced by Exploration Mutation Agent)}]
\begin{lstlisting}[basicstyle=\ttfamily\scriptsize,breaklines=true,columns=fullflexible]
# Grassmannian Subspace Routing (GrassRoute)

- Alias: `GrassmannianSubspaceRouting`
- Family: manifold hyper-connection with Grassmannian projection routing
- Overrides:
  - `hyper_conn_type = "custom"`
  - `hyper_conn_n = 4`
  - `manifold_strategy = "grassmannian_subspace"`
- Key mutation from parent (HyperbolicRotationRouting / SO(S) Givens):
  - **Replaces SO(S) rotation routing with Grassmannian subspace projection routing.**
  - Instead of rotating all S streams via a full orthogonal matrix, we learn a k-dimensional subspace of the S-stream space (a point on the Grassmannian Gr(k, S)) and project streams onto it for branch input, then lift back for merge.
  - The subspace is parameterized via a tall-skinny semi-orthogonal matrix U in St(k, S) (Stiefel manifold), maintained via Cayley retraction after each gradient step (or equivalently, via a skew-symmetric parameterization for differentiability).
  - **Width connection**: project S streams to k-dim subspace via U^T, form branch input as learned combination of k projected streams.
  - **Depth connection**: lift branch output back to S streams via U, with per-stream learnable gating.
  - **Dynamic subspace perturbation**: input-dependent skew-symmetric perturbation to U via Cayley map, enabling content-adaptive subspace selection.
  - This explores a fundamentally different manifold geometry (Grassmannian) vs. the parent's SO(S), with lower effective dimensionality (k < S) acting as an information bottleneck.

---

## Evaluation Snapshot

- Benchmark metric: mean_val_loss
- Status: awaiting benchmark

---

## Evaluation Snapshot

- Benchmark metric: mean_val_loss
- Primary metric (raw benchmark): 1.6934666666666667
- Higher is better: False
- Directional score: -1.6934666666666667
- Correctness score: 4
- Correctness binary: 1
- Originality score: 4
- Originality binary: 1
\end{lstlisting}
\end{tcolorbox}

\begin{tcolorbox}[breakable,colback=orange!4,colframe=orange!75!black,title={theory\_content excerpt (produced by Exploration Mutation Agent)}]
\begin{lstlisting}[basicstyle=\ttfamily\scriptsize,breaklines=true,columns=fullflexible]
\section{Grassmannian Subspace Routing (GrassRoute)}

\subsection*{Motivation}
The parent node parameterizes stream routing via the full rotation group $\mathrm{SO}(S)$, which treats all $S$ streams symmetrically. We hypothesize that an \emph{information bottleneck} in the stream space can improve generalization: rather than mixing all streams equally, we select a $k$-dimensional subspace of the $S$-stream space and route information through it.

The natural manifold for $k$-dimensional subspaces of $\mathbb{R}^S$ is the \textbf{Grassmannian} $\mathrm{Gr}(k, S)$. A point on $\mathrm{Gr}(k, S)$ can be represented by a semi-orthogonal matrix $U \in \mathrm{St}(k, S) \subset \mathbb{R}^{S \times k}$ satisfying $U^\top U = I_k$.

\subsection*{Cayley Parameterization}
To maintain $U$ on the Stiefel manifold differentiably, we use the \textbf{Cayley transform} of a skew-symmetric matrix. Given a learnable matrix $A \in \mathbb{R}^{S \times S}$, we form:
\[
A_{\text{skew}} = A - A^\top, \quad C = (I + A_{\text{skew}})(I - A_{\text{skew}})^{-1},
\]
which yields $C \in \mathrm{O}(S)$. We then take the first $k$ columns: $U = C_{:, :k} \in \mathrm{St}(k, S)$.

For computational efficiency with $S=4, k=2$, the $4 \times 4$ matrix inverse is cheap.

\subsection*{Width Connection}
Given residual streams $R \in \mathbb{R}^{B \times T \times S \times D}$:
\begin{enumerate}
\item Compute dynamic perturbation: $\Delta A = f_{\text{dyn}}(\mathrm{norm}(R))$ reshaped to skew-symmetric.
\item Form $U(\theta) = \mathrm{Cayley}(A_{\text{static}} + \epsilon \cdot \Delta A)_{:, :k}$.
\item Project: $P = U^\top R \in \mathbb{R}^{B \times T \times k \times D}$.
\item Branch input: $x = \sum_{i=1}^{k} \alpha_i P_i$ with learnable $\alpha \in \Delta^k$ (softmax weights).
\end{enumerate}

\subsection*{Depth Connection}
After branch produces $y \in \mathbb{R}^{B \times T \times D}$:
\begin{enumerate}
\item Lift: $\hat{y} = U \cdot (\gamma \odot \mathbf{1}_k) \cdot y$ where $\gamma \in \mathbb{R}^k$ are learnable scales, broadcasting $y$ into $k$ copies then projecting to $S$ streams.
\item Merge: $R' = R + \beta \odot \hat{y}$ with per-stream sigmoid gate $\beta \in (0,1)^S$.
\end{enumerate}

\subsection*{Properties}
\begin{itemize}
\item The Grassmannian bottleneck ($k < S$) acts as a structured regularizer.
\item The Cayley parameterization is always differentiable and numerically stable.
\item Dynamic subspace perturbation enables input-dependent routing without leaving the manifold.
\item For $k = S$, this reduces to a full orthogonal routing (similar to parent); for $k = 1$, it becomes a rank-1 projection.
\end{itemize}

\subsection*{Equation to Code Mapping}
\begin{tabular}{p{0.40\linewidth}p{0.54\linewidth}}
\textbf{Math object} & \textbf{Code object} \\
\hline
$A_{\text{static}}$ & \texttt{self.A\_static} \\
$\mathrm{Cayley}(\cdot)$ & \texttt{cayley\_transform()} \\
$U = C_{:,:k}$ & slicing in \texttt{get\_U()} \\
$f_{\text{dyn}}$ & \texttt{self.dyn\_A\_proj} \\
$\alpha$ (branch weights) & \texttt{self.branch\_alpha} \\
$\gamma$ (lift scales) & \texttt{self.lift\_gamma} \\
$\beta$ (merge gate) & \texttt{self.merge\_beta\_static, self.merge\_beta\_dyn\_proj} \\
$\mathcal{B}$ call & \texttt{self.branch(...)} inside \texttt{forward} \\
\end{tabular}
\end{lstlisting}
\end{tcolorbox}

\begin{tcolorbox}[breakable,colback=orange!4,colframe=orange!75!black,title={code\_content excerpt (produced by Exploration Mutation Agent)}]
\begin{lstlisting}[basicstyle=\ttfamily\scriptsize,breaklines=true,columns=fullflexible]
from __future__ import annotations
import math
from typing import Callable, Tuple

import torch
import torch.nn as nn
import torch.nn.functional as F
from torch import Tensor

# -- Contract identifiers ----------------------------------------------
ATTENTION_ALIAS = "GrassmannianSubspaceRouting"
ATTENTION_NODE_ID = "g002_n0024_66b987"
def get_mhc_lite_overrides() -> dict:
    return {
        "hyper_conn_type": "custom",
        "hyper_conn_n": 4,
        "manifold_strategy": "grassmannian_subspace",
    }


# -- Helpers -----------------------------------------------------------

class RMSNorm(nn.Module):
    def __init__(self, dim: int):
        super().__init__()
        self.scale = dim ** 0.5
        self.gamma = nn.Parameter(torch.zeros(dim))

    def forward(self, x: Tensor) -> Tensor:
        return F.normalize(x, dim=-1) * self.scale * (self.gamma + 1.0)


def cayley_transform(A_skew: Tensor) -> Tensor:
    """
    Cayley transform of a skew-symmetric matrix.
    A_skew: (..., S, S) skew-symmetric
    Returns: (..., S, S) orthogonal matrix
    """
    S = A_skew.shape[-1]
    eye = torch.eye(S, device=A_skew.device, dtype=A_skew.dtype)
    # Expand eye to batch dims
    for _ in range(len(A_skew.shape) - 2):
        eye = eye.unsqueeze(0)
    eye = eye.expand_as(A_skew)
    # C = (I + A)(I - A)^{-1}
    return torch.linalg.solve(eye - A_skew, eye + A_skew)


def make_skew(A: Tensor) -> Tensor:
    """Make a matrix skew-symmetric: (A - A^T) / 2."""
    return (A - A.transpose(-2, -1)) * 0.5


# -- Main module -------------------------------------------------------

class EvoHyperConnection(nn.Module):
    """
    Hyper-connection using Grassmannian subspace projection routing.
    Routes information through a learned k-dimensional subspace of the
    S-stream space, parameterized via Cayley transform on St(k, S).
    """

    def __init__(self, num_streams: int, dim: int, branch: nn.Module):
        super().__init__()
        self.branch = branch
        self.num_streams = num_streams  # S
        self.dim = dim
        S = num_streams
        self.k = max(2, S // 2)  # subspace dimension; k=2 for S=4
        k = self.k

        # -- Norm --
        self.norm = RMSNorm(dim * S)

        # -- Static Cayley parameter (S x S, will be made skew-symmetric) --
        self.A_static = nn.Parameter(torch.zeros(S, S))
        # Small random init to break symmetry
        nn.init.normal_(self.A_static, std=0.02)

        # -- Dynamic subspace perturbation --
        # Project from concatenated stream features to S*S skew params
        # We only need S*(S-1)/2 free params but output S*S and symmetrize
        self.dyn_A_proj = nn.Linear(dim * S, S * S, bias=False)
        nn.init.zeros_(self.dyn_A_proj.weight)
        self.dyn_scale = nn.Parameter(torch.full((), 0.01))

        # -- Branch input: softmax weights over k projected streams --
        self.branch_alpha = nn.Parameter(torch.zeros(k))
        with torch.no_grad():
            self.branch_alpha[0] = 2.0  # bias toward first projected stream

        # -- Depth: lift scales per subspace dim --
        self.lift_gamma = nn.Parameter(torch.ones(k))

        # -- Depth: merge gate (static + dynamic) per stream --
        self.merge_beta_static = nn.Parameter(torch.zeros(S))
        with torch.no_grad():
            self.merge_beta_static[0] = 2.0  # first stream active
        self.merge_beta_dyn_proj = nn.Linear(dim * S, S, bias=False)
        nn.init.zeros_(self.merge_beta_dyn_proj.weight)
        self.merge_beta_dyn_scale = nn.Parameter(torch.full((), 0.01))

    def _get_U(self, A_skew: Tensor) -> Tensor:
        """
        Compute semi-orthogonal U in St(k, S) from skew-symmetric A.
        A_skew: (..., S, S)
        Returns: (..., S, k)
        """
        C = cayley_transform(A_skew)  # (..., S, S) orthogonal
        return C[..., :self.k]  # (..., S, k)

    def forward(
        self,
        residuals: Tensor,
        *branch_args,
        **branch_kwargs,
    ) -> Tensor:
        S = self.num_streams
        D = self.dim
        k = self.k

        # residuals: (B*S, T, D)
        BS, T, _D = residuals.shape
        B = BS // S

        # Reshape to (B, T, S, D)
        R = residuals.view(B, S, T, D).permute(0, 2, 1, 3)  # (B, T, S, D)

        # -- Shared features for dynamic projections --
        flat = R.reshape(B, T, S * D)  # (B, T, S*D)
        normed = self.norm(flat)  # (B, T, S*D)

        # -- Compute subspace basis U --
        # Static skew
        A_skew_static = make_skew(self.A_static)  # (S, S)

        # Dynamic perturbation
        dyn_raw = self.dyn_A_proj(normed)  # (B, T, S*S)
        dyn_A = dyn_raw.view(B, T, S, S)
        dyn_A_skew = make_skew(dyn_A)  # (B, T, S, S)

        A_skew = A_skew_static + self.dyn_scale.abs() * dyn_A_skew  # (B, T, S, S)

        U = self._get_U(A_skew)  # (B, T, S, k)

        # -- Width: project streams to subspace --
        # P = U^T @ R: (B, T, k, D)
        # R is (B, T, S, D), U is (B, T, S, k)
        P = torch.einsum('btsk,btsd->btkd', U, R)  # (B, T, k, D)

        # Branch input: weighted combination of k projected streams
        alpha = F.softmax(self.branch_alpha, dim=0)  # (k,)
        branch_input = torch.einsum('k,btkd->btd', alpha, P)  # (B, T, D)

        # -- Call branch --
        branch_output = self.branch(branch_input, *branch_args, **branch_kwargs)  # (B, T, D)

        # -- Depth: lift branch output back to S streams --
        # Broadcast y into k copies, scale by gamma, then project via U
        # y_k = gamma_i * y for each i in k: (B, T, k, D)
        gamma = self.lift_gamma  # (k,)
        y_k = branch_output.unsqueeze(-2) * gamma.view(1, 1, k, 1)  # (B, T, k, D)

        # Lift to S streams: hat_y = U @ y_k: (B, T, S, D)
        hat_y = torch.einsum('btsk,btkd->btsd', U, y_k)  # (B, T, S, D)

        # -- Merge gate --
        dyn_beta = self.merge_beta_dyn_proj(normed)  # (B, T, S)
        beta = torch.sigmoid(
            self.merge_beta_static + self.merge_beta_dyn_scale.abs() * dyn_beta
        )  # (B, T, S)

        # Merge: residual + gated lifted branch output
        output = R + beta.unsqueeze(-1) * hat_y  # (B, T, S, D)

        # Reshape back to (B*S, T, D)
        output = output.permute(0, 2, 1, 3).reshape(BS, T, D)

        return output


# -- Builder functions -------------------------------------------------

def build_custom_hyper_connection(
    num_streams: int, *, dim: int, branch: nn.Module
) -> EvoHyperConnection:
    return EvoHyperConnection(
        num_streams=int(num_streams), dim=int(dim), branch=branch
    )
\end{lstlisting}
\end{tcolorbox}

\begin{tcolorbox}[breakable,colback=cyan!3,colframe=cyan!55!black,title={benchmark\_summary excerpt (Benchmark Agent output)}]
\begin{lstlisting}[basicstyle=\ttfamily\scriptsize,breaklines=true,columns=fullflexible]
# Benchmark Summary
- Mode: mhc_lite_attention (train-eval)
- Fitness mode: val_loss
- Attention alias: GrassmannianSubspaceRouting
- Metric: mean_val_loss
- Seed count: 3 (ok=3, failed=0)
- Success rate: 1.000
- Mean val loss (ok seeds): 1.693467
- Failed-seed policy: impute with worst successful val_loss (1.794600)
- Final val_loss used for primary metric (after imputation): 1.693467
- Higher is better: False
- Node: g002_n0024_66b987
- hyper_conn_type: custom
- hyper_conn_n: 4
- manifold_strategy: grassmannian_subspace
- Directional score rule: score = -primary_metric (higher_is_better=false)
\end{lstlisting}
\end{tcolorbox}

\begin{tcolorbox}[breakable,colback=teal!3,colframe=teal!60!black,title={review feedback excerpt (Reviewer Agent output)}]
\begin{lstlisting}[basicstyle=\ttfamily\scriptsize,breaklines=true,columns=fullflexible]
## Review: Grassmannian Subspace Routing (GrassRoute)

### Contract Compliance
- **ATTENTION_ALIAS**: defined as `"GrassmannianSubspaceRouting"` [ok]
- **ATTENTION_NODE_ID**: matches `"g002_n0024_66b987"` [ok]
- **get_mhc_lite_overrides()**: returns exactly `{"hyper_conn_type": "custom", "hyper_conn_n": 4, "manifold_strategy": "grassmannian_subspace"}` -- only permitted keys [ok]
- **EvoHyperConnection(nn.Module)**: defined [ok]
- **build_custom_hyper_connection(num_streams, *, dim, branch)**: returns `EvoHyperConnection` instance [ok]
- **self.branch(...)** called in forward: Yes -- `self.branch(branch_input, *branch_args, **branch_kwargs)` is invoked exactly once per forward pass [ok]
- **Shape/dtype/device preservation**: input `(B*S, T, D)` -> reshaped -> processed -> reshaped back to `(B*S, T, D)` [ok]
- **Gradient flow**: all operations (einsum, Cayley via `linalg.solve`, softmax, sigmoid) are differentiable [ok]

### Correctness Analysis

**Cayley transform implementation**: The code computes `(I + A_skew)(I - A_skew)^{-1}` via `torch.linalg.solve(I - A_skew, I + A_skew)`. This is mathematically correct: `solve(B, C)` returns `B^{-1}C`, so this yields `(I - A_skew)^{-1}(I + A_skew)`. Note this is the *transpose* of the Cayley map as written in the theory (`(I+A)(I-A)^{-1}`), but since the Cayley map of a skew-symmetric matrix yields an orthogonal matrix, and the transpose of an orthogonal matrix is also orthogonal, the first k columns still form a valid element of St(k, S). The subspace spanned (Grassmannian point) is different from what the theory formula would give, but this is a valid parameterization of Gr(k, S) nonetheless. This is a minor theory-code mismatch but not a functional bug.

**Dynamic perturbation**: The per-token dynamic A computation creates a (B, T, S, S) tensor, which means a separate Cayley transform per token per batch element. For S=4 this is a 4x4 solve, which is computationally feasible but adds overhead. The `dyn_scale` initialized at 0.01 with `.abs()` ensures small perturbations initially -- reasonable for stability.

**Potential concern -- seed 3337 variance**: Val loss of 1.7946 vs ~1.64 for other seeds shows notable variance (std ~ 0.072). This suggests the dynamic Cayley computation may introduce some optimization instability, though all seeds succeeded.

**Numerical stability**: `torch.linalg.solve` on `(I - A_skew)` where `A_skew` is skew-symmetric: eigenvalues of `(I - A_skew)` are `1 - i*lambda` (purely imaginary eigenvalues of skew-symmetric), so the matrix is always invertible. This is numerically stable [ok].

### Originality Analysis

The idea of using Grassmannian subspace projection as an information bottleneck for stream routing is genuinely novel in this evolutionary context. Key original elements:

1. **Grassmannian geometry**: Moving from SO(S) (parent) to Gr(k, S) is a meaningful geometric shift -- it introduces a rank-constrained bottleneck rather than full-rank rotation.
2. **Input-dependent subspace via Cayley perturbation**: The dynamic skew-symmetric perturbation enabling content-adaptive subspace selection is a well-motivated extension.
3. **Asymmetric project-lift architecture**: The width (project via U^T) and depth (lift via U with gating) connections create an asymmetric information flow that differs from standard residual or rotation-based routing.

The Cayley parameterization of Stiefel/Grassmannian manifolds is known in the optimization literature, but its application to hyper-connection routing with dynamic perturbation is a creative combination.

### Benchmark Performance
- Mean val_loss: 1.6935 across 3 seeds (all successful)
- This is a reasonable result; the relatively high variance across seeds (0.072 std) is a minor concern.

### Summary
Well-executed node with a mathematically coherent Grassmannian routing mechanism. All contract requirements are met. The minor theory-code discrepancy in the Cayley transform ordering does not affect correctness of the manifold parameterization. The approach is genuinely original in replacing full orthogonal routing with subspace projection routing.

# evaluation
Correctness_score=4, Originality_score=4
\end{lstlisting}
\end{tcolorbox}

\begin{figure}[p]
\centering
\resizebox{\textwidth}{!}{
\begin{tikzpicture}[
  x=2.9cm,
  y=1.28cm,
  >={Latex[length=2.0mm,width=1.5mm]},
  edge/.style={draw=black!32, line width=0.34pt, line cap=round, line join=round, shorten >=1.2pt, shorten <=1.2pt},
  normalnode/.style={draw=black!50, rounded corners=1.5mm, fill=gray!6, align=center, text width=2.45cm, font=\tiny, inner sep=1.8pt},
  winnode/.style={draw=green!50!black, rounded corners=1.5mm, fill=green!8, align=center, text width=2.45cm, font=\tiny, inner sep=1.8pt},
  goldnode/.style={draw=blue!70!black, rounded corners=1.5mm, fill=blue!7, align=center, text width=2.45cm, font=\tiny\bfseries, inner sep=1.9pt}
]
\node[font=\footnotesize\bfseries] at (0, 1.2) {Gen 0};
\node[normalnode] (A0) at (0, 0) {A0\\Transformer\\Residual\\Attention\\Seed\\$m=4.856$};
\node[normalnode] (B0) at (0, -1) {B0\\MHC Lite\\Attention\\Seed\\$m=\infty$};
\node[normalnode] (C0) at (0, -2) {C0\\HC Attention\\Seed\\$m=\infty$};
\node[normalnode] (D0) at (0, -3) {D0\\Hyperbolic\\Poincare\\Routing\\$m=4.495$};
\node[normalnode] (E0) at (0, -4) {E0\\Hyperbolic\\Rotation\\Routing\\$m=\infty$};
\node[normalnode] (F0) at (0, -5) {F0\\Hyp Exp\\Route 1\\$m=5.457$};
\node[normalnode] (G0) at (0, -6) {G0\\Grassmannian\\Subspace\\Routing\\$m=\infty$};
\node[normalnode] (H0) at (0, -7) {H0\\Grassmannian\\Subspace\\Routing\\$m=\infty$};
\node[font=\footnotesize\bfseries] at (1, 1.2) {Gen 1};
\node[winnode] (A1) at (1, 0) {A1\\Hyperbolic\\Exp Map\\Routing\\$m=5.125$};
\node[normalnode] (B1) at (1, -1) {B1\\MHC Lite\\Dtype Fix\\$m=5.535$};
\node[normalnode] (C1) at (1, -2) {C1\\HC Attention\\Seed Dtype\\Fix\\$m=5.779$};
\node[winnode] (D1) at (1, -3) {D1\\Hyperbolic\\Poincare\\Routing 2\\$m=4.165$};
\node[goldnode] (E1) at (1, -4) {E1\\Hyperbolic\\Rotation\\Routing\\$m=1.845$};
\node[normalnode] (F1) at (1, -5) {F1\\Grass Route\\1\\$m=\infty$};
\node[winnode] (G1) at (1, -6) {G1\\Grassmannian\\Subspace\\Routing\\$m=5.006$};
\node[normalnode] (H1) at (1, -7) {H1\\Grassmannian\\Subspace\\Routing\\$m=5.525$};
\node[font=\footnotesize\bfseries] at (2, 1.2) {Gen 2};
\node[winnode] (A2) at (2, 0) {A2\\Givens\\Hyperbolic\\Routing\\$m=1.768$};
\node[normalnode] (B2) at (2, -1) {B2\\Hyperbolic\\Grassmannian\\Hybrid\\Routing\\$m=5.334$};
\node[normalnode] (C2) at (2, -2) {C2\\Hyperbolic\\Tangent\\Bundle\\Routing\\$m=5.449$};
\node[normalnode] (D2) at (2, -3) {D2\\Hyperbolic\\Exp Map\\Routing\\$m=5.802$};
\node[normalnode] (E2) at (2, -4) {E2\\Grass Route\\2\\$m=5.294$};
\node[normalnode] (F2) at (2, -5) {F2\\Stiefel\\Frame\\Routing\\$m=5.581$};
\node[winnode] (G2) at (2, -6) {G2\\Hyperbolic\\Rotation\\Routing\\$m=1.845$};
\node[goldnode] (H2) at (2, -7) {H2\\Grassmannian\\Subspace\\Routing\\$m=1.693$};
\node[font=\footnotesize\bfseries] at (3, 1.2) {Gen 3};
\node[normalnode] (A3) at (3, 0) {A3\\Grassmannian\\Hyperbolic\\Routing\\$m=1.698$};
\node[normalnode] (B3) at (3, -1) {B3\\Spectral\\Cayley\\Orthogonal\\Routing\\$m=\infty$};
\node[normalnode] (C3) at (3, -2) {C3\\Hyperbolic\\Tangent\\Bundle\\Routing 2\\$m=5.294$};
\node[normalnode] (D3) at (3, -3) {D3\\Hyperbolic\\Exp Map\\Routing 2\\$m=5.475$};
\node[normalnode] (E3) at (3, -4) {E3\\Poincare\\Route\\$m=5.387$};
\node[winnode] (F3) at (3, -5) {F3\\Stiefel\\Frame Route\\2\\$m=5.220$};
\node[goldnode] (G3) at (3, -6) {G3\\Grassmannian\\Subspace\\Routing\\$m=1.693$};
\node[normalnode] (H3) at (3, -7) {H3\\Poincare\\Hyperbolic\\Routing\\$m=\infty$};
\draw[edge, ->] ([yshift=0.0pt]A0.west) -- ++(-0.50,0) |- ([yshift=0.0pt]D0.west);
\draw[edge, ->] ([yshift=0.0pt]B0.west) -- ++(-0.50,0) |- ([yshift=0.0pt]E0.west);
\draw[edge, ->] ([yshift=0.0pt]C0.west) -- ++(-0.50,0) |- ([yshift=0.0pt]F0.west);
\draw[edge, ->] ([yshift=0.0pt]D0.west) -- ++(-0.50,0) |- ([yshift=0.0pt]G0.west);
\draw[edge, ->] ([yshift=0.0pt]E0.west) -- ++(-0.50,0) |- ([yshift=0.0pt]H0.west);
\draw[edge, ->] ([yshift=0.0pt]A0.east) -- ([yshift=0.0pt]A1.west);
\draw[edge, ->] ([yshift=0.0pt]B0.east) -- ([yshift=0.0pt]B1.west);
\draw[edge, ->] ([yshift=0.0pt]C0.east) -- ([yshift=0.0pt]C1.west);
\draw[edge, ->] ([yshift=0.0pt]D0.east) -- ([yshift=0.0pt]D1.west);
\draw[edge, ->] ([yshift=0.0pt]E0.east) -- ([yshift=0.0pt]E1.west);
\draw[edge, ->] ([yshift=0.0pt]F0.east) -- ([yshift=0.0pt]F1.west);
\draw[edge, ->] ([yshift=0.0pt]G0.east) -- ([yshift=0.0pt]G1.west);
\draw[edge, ->] ([yshift=0.0pt]H0.east) -- ([yshift=0.0pt]H1.west);
\draw[edge, ->] ([yshift=-2.2pt]E1.east) -- ([yshift=-2.2pt]A2.west);
\draw[edge, ->] ([yshift=2.2pt]D1.east) -- ([yshift=2.2pt]A2.west);
\draw[edge, ->] ([yshift=-2.2pt]G1.east) -- ([yshift=-2.2pt]B2.west);
\draw[edge, ->] ([yshift=2.2pt]A1.east) -- ([yshift=2.2pt]B2.west);
\draw[edge, ->] ([yshift=0.0pt]B1.east) -- ([yshift=0.0pt]C2.west);
\draw[edge, ->] ([yshift=0.0pt]C1.east) -- ([yshift=0.0pt]D2.west);
\draw[edge, ->] ([yshift=0.0pt]F1.east) -- ([yshift=0.0pt]E2.west);
\draw[edge, ->] ([yshift=0.0pt]H1.east) -- ([yshift=0.0pt]F2.west);
\draw[edge, ->] ([yshift=0.0pt]E1.east) -- ([yshift=0.0pt]G2.west);
\draw[edge, ->] ([yshift=0.0pt]E1.east) -- ([yshift=0.0pt]H2.west);
\draw[edge, ->] ([yshift=-2.2pt]H2.east) -- ([yshift=-2.2pt]A3.west);
\draw[edge, ->] ([yshift=2.2pt]A2.east) -- ([yshift=2.2pt]A3.west);
\draw[edge, ->] ([yshift=0.0pt]B2.east) -- ([yshift=0.0pt]B3.west);
\draw[edge, ->] ([yshift=0.0pt]C2.east) -- ([yshift=0.0pt]C3.west);
\draw[edge, ->] ([yshift=0.0pt]D2.east) -- ([yshift=0.0pt]D3.west);
\draw[edge, ->] ([yshift=0.0pt]E2.east) -- ([yshift=0.0pt]E3.west);
\draw[edge, ->] ([yshift=0.0pt]F2.east) -- ([yshift=0.0pt]F3.west);
\draw[edge, ->] ([yshift=0.0pt]H2.east) -- ([yshift=0.0pt]G3.west);
\draw[edge, ->] ([yshift=0.0pt]H2.east) -- ([yshift=0.0pt]H3.west);
\end{tikzpicture}}
\caption{Full generation graph for the reported transformer \texttt{theory+code} single-island run, rendered directly from \texttt{ga\_data.json}. Each node is annotated with its flat id, recovered alias, and benchmark primary metric. Nodes with \(m=\infty\) denote benchmark execution failures due to implementation/runtime incompatibilities, not valid high-loss models. Edges follow recorded parent links; generation columns correspond to iterations 0 through 3.}
\label{fig:transformer-ga-graph}
\end{figure}
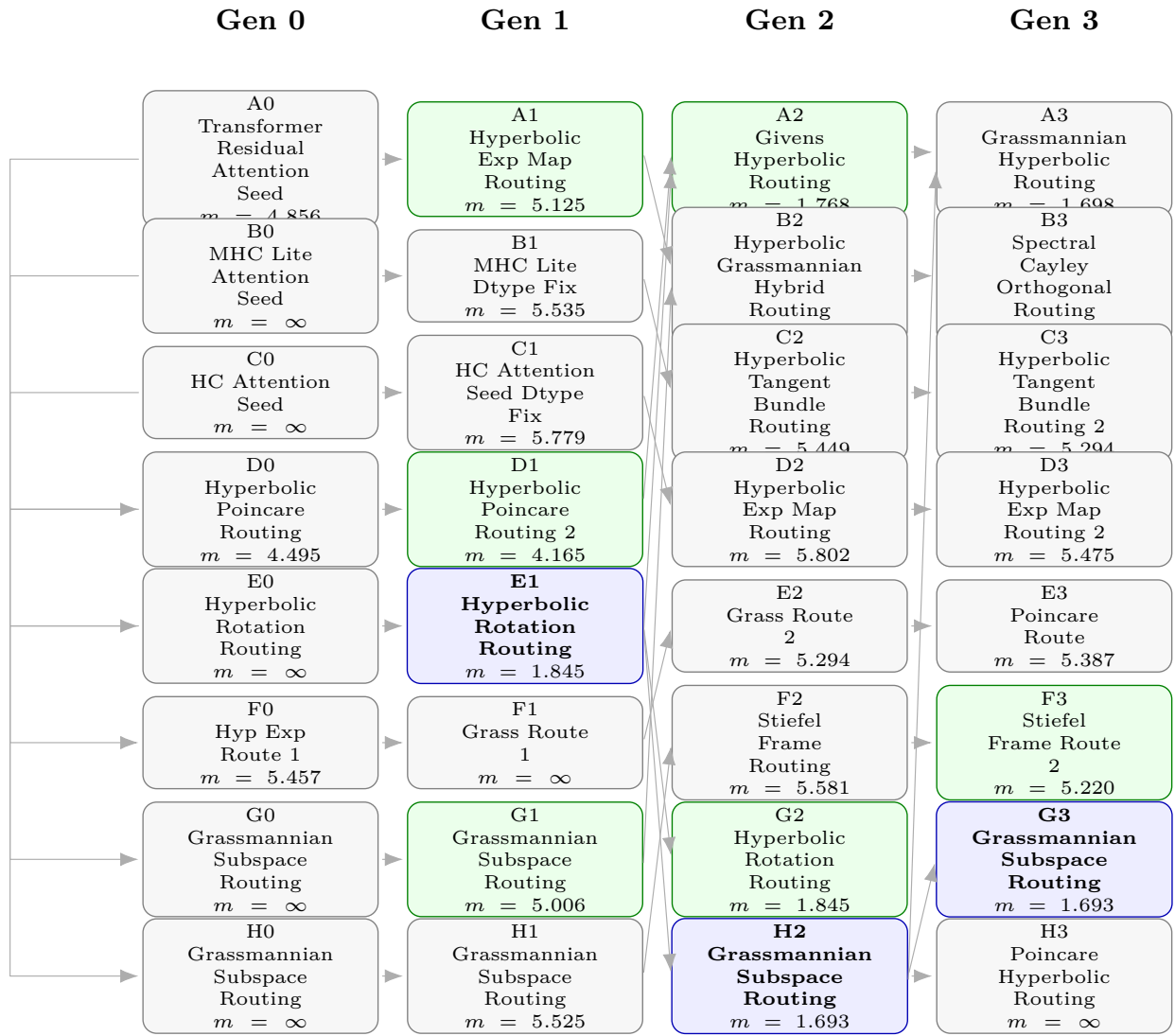

\subsubsection{Theory+code novelty-versus-retrieval audit for transformer hyper-connections}
\label{app:manifold-attention-survey}

This appendix summarizes a post hoc survey we prepared to contextualize the transformer theory+code hyper-connection discoveries against existing manifold-attention and HC literature. The survey was \emph{not} exposed to CliffSearch agents: the SDK calls in the reported run had no external retrieval tools, no browsing capability, and none of the papers or notes below were injected into prompts. The goal of the audit is therefore not to ask whether every geometric ingredient is new in all of attention, but whether CliffSearch is merely replaying known HC mechanisms or instead exporting ideas from the broader attention literature into new hyper-connection designs.

\begingroup
\scriptsize
\setlength{\tabcolsep}{4pt}
\begin{longtable}{>{\raggedright\arraybackslash}p{2.5cm}>{\raggedright\arraybackslash}p{2.0cm}>{\raggedright\arraybackslash}p{2.5cm}>{\raggedright\arraybackslash}p{4.9cm}>{\centering\arraybackslash}p{1.4cm}}
\caption{Representative manifold-attention papers grouped by geometry, scoring rule, and aggregation primitive. The key distinction is whether geometry affects only the score or also the value aggregation rule itself.}
\label{tab:manifold-attention-survey}\\
\toprule
\textbf{Paper} & \textbf{Geometry} & \textbf{Score / similarity} & \textbf{Aggregation primitive} & \makecell[c]{\textbf{Routing}\\\textbf{used?}} \\
\midrule
\endfirsthead
\toprule
\textbf{Paper} & \textbf{Geometry} & \textbf{Score / similarity} & \textbf{Aggregation primitive} & \makecell[c]{\textbf{Routing}\\\textbf{used?}} \\
\midrule
\endhead
\bottomrule
\endfoot
\textbf{Hyperbolic Attention Networks}\\
Gulcehre et al. \cite{gulcehre2019hyperbolic}
& Hyperboloid / Klein hyperbolic models
& Hyperbolic distance-based attention
& Einstein midpoint for attentive value aggregation
& No \\

\addlinespace
\textbf{Fully Hyperbolic Neural Networks}\\
Chen et al. \cite{chen2022fully}
& Lorentz model
& Lorentzian attention
& Lorentz centroid as a hyperbolic aggregation primitive
& No \\

\addlinespace
\textbf{Hypformer}\\
Yang et al. \cite{yang2024hypformer}
& Lorentz model
& Hyperbolic softmax attention and linear hyperbolic attention
& Lorentzian midpoint / centroid; explicitly relates Lorentz centroid, Einstein midpoint, and gyromidpoint
& No \\

\addlinespace
\textbf{Hyperbolic Graph Attention Network}\\
Zhang et al. \cite{zhang2022hgat}
& Hyperbolic graph representations
& Hyperbolic graph attention
& Hyperbolic neighborhood aggregation, graph-attention centered rather than a named midpoint primitive
& No \\

\addlinespace
\textbf{Mixed-curvature / stereographic Transformer}\\
Cho et al. \cite{cho2023fps}
& Product constant-curvature spaces in the stereographic model
& Query/key scoring in tangent space at the origin
& Einstein midpoint per head in the curved component space
& No \\

\addlinespace
\textbf{Spherical Transformer for pediatric cortical surfaces}\\
Cheng et al. \cite{cheng2022sphericalqa}
& Spherical manifold / cortical surface mesh
& Standard self-attention inside local spherical patches
& Extrinsic Euclidean weighted sum over patch tokens
& No \\

\addlinespace
\textbf{Spherical Transformer on cortical surfaces}\\
Cheng et al. \cite{cheng2022sphericalcortical}
& Sphere / cortical surface patches
& Self-attention over patch tokens
& Extrinsic token-wise weighted sum after spherical tokenization
& No \\

\addlinespace
\textbf{STF}\\
Cheng et al. \cite{cheng2025stf}
& Spherical cortical-surface manifold
& Global and local self-attention at patch / vertex levels
& Extrinsic patch / vertex aggregation rather than an intrinsic spherical barycenter
& No \\

\addlinespace
\textbf{MAtt}\\
Pan et al. \cite{pan2022matt}
& SPD manifold under the Log-Euclidean metric
& Distance-based attention on SPD-valued queries and keys
& Weighted Log-Euclidean mean
& No \\

\addlinespace
\textbf{Riemannian Self-Attention for SPD networks}\\
Wang et al. \cite{wang2023spdsa}
& SPD manifold
& Riemannian-metric-based SPD self-attention
& Weighted Fr\'echet / Riemannian mean
& No \\

\addlinespace
\textbf{Structure-Preserving Transformers for SPD sequences}\\
Seraphim et al. \cite{seraphim2024spdtrans}
& SPD manifold via the Log-Euclidean chart
& Standard attention in tokenized log-coordinates
& Log-Euclidean weighted linear combination
& No \\

\addlinespace
\textbf{Grassmannian Manifold Self-Attention}\\
Wang et al. \cite{wang2024grassmann}
& Grassmannian with projection metric
& Projection-distance-based similarity
& Weighted Fr\'echet mean, approximated by projector-embedding averages plus re-orthonormalization
& No \\

\addlinespace
\textbf{Correlation Manifold Self-Attention}\\
Hu et al. \cite{hu2025coratt}
& Full-rank correlation manifolds under OLM / LSM
& Geodesic-distance-based attention
& Closed-form weighted Fr\'echet mean
& No \\

\addlinespace
\textbf{GyroAtt}\\
Wang et al. \cite{wang2025gyroatt}
& Gyrovector spaces for SPD, SPSD, and Grassmannian manifolds
& Geodesic-distance-based scores in gyrovector spaces
& Weighted Fr\'echet mean in the target manifold
& No \\
\end{longtable}
\endgroup

\begingroup
\scriptsize
\setlength{\tabcolsep}{4pt}
\begin{longtable}{>{\raggedright\arraybackslash}p{2.5cm}>{\raggedright\arraybackslash}p{2.0cm}>{\raggedright\arraybackslash}p{3.2cm}>{\raggedright\arraybackslash}p{4.6cm}>{\centering\arraybackslash}p{1.4cm}}
\caption{Representative routing / expert-fusion papers. These are useful because they separate routing from geometry-aware fusion: centroid-like operations often appear after routing, not as the routing rule itself.}
\label{tab:manifold-routing-survey}\\
\toprule
\textbf{Paper} & \textbf{Geometry} & \textbf{Routing mechanism} & \textbf{Fusion after routing} & \makecell[c]{\textbf{Routing}\\\textbf{used?}} \\
\midrule
\endfirsthead
\toprule
\textbf{Paper} & \textbf{Geometry} & \textbf{Routing mechanism} & \textbf{Fusion after routing} & \makecell[c]{\textbf{Routing}\\\textbf{used?}} \\
\midrule
\endhead
\bottomrule
\endfoot
\textbf{HELM-MiCE}\\
He et al. \cite{he2025helm}
& Lorentz hyperbolic spaces with distinct curvatures across experts
& Learned gating scores select routed experts using expert-centroid affinities
& Lorentzian centroid merges selected expert outputs
& Yes \\

\addlinespace
\textbf{GeoMoE}\\
Cao et al. \cite{cao2026geomoe}
& Mixed-curvature graph representation learning
& Curvature-guided adaptive routing via a graph-aware gating network
& Geometry-aware expert fusion; centroid-style primitives appear after routing rather than as the router itself
& Yes \\
\end{longtable}
\endgroup

\begingroup
\scriptsize
\setlength{\tabcolsep}{4pt}
\begin{longtable}{>{\raggedright\arraybackslash}p{2.5cm}>{\raggedright\arraybackslash}p{2.4cm}>{\raggedright\arraybackslash}p{3.7cm}>{\raggedright\arraybackslash}p{4.8cm}}
\caption{Direct HC / mHC literature surveyed by manifold or constraint set. This literature is materially narrower than the broader manifold-attention literature: the geometry is usually placed on the residual mixing matrix, not on hidden-state routing in token or stream space.}
\label{tab:hc-manifold-survey}\\
\toprule
\textbf{HC-family paper} & \textbf{Constraint set / manifold} & \textbf{Where geometry is imposed} & \textbf{Audit relevance for the reported run} \\
\midrule
\endfirsthead
\toprule
\textbf{HC-family paper} & \textbf{Constraint set / manifold} & \textbf{Where geometry is imposed} & \textbf{Audit relevance for the reported run} \\
\midrule
\endhead
\bottomrule
\endfoot
\textbf{HC}\\
Zhu et al. \cite{zhu2024hyperconnections}
& Unconstrained Euclidean matrix space
& Learned dense residual-stream mixing matrix
& This is the unconstrained dense baseline that motivated mHC-style stabilizing constraints. It is relevant mainly as the precursor to the search task, not as a novelty match for the discovered manifold families. \\

\addlinespace
\textbf{mHC}
& Birkhoff polytope / doubly stochastic matrices
& Residual mixing matrix constrained by Sinkhorn-style doubly stochastic projection \cite{xie2025mhc,birkhoff1946three,sinkhorn1967concerning}
& This is the closest direct precedent for the supplied \texttt{MHCLiteAttentionSeed}. It explains why permutation-mixture and doubly stochastic routing should be treated as known HC-line ingredients, not novel discoveries. \\

\addlinespace
\textbf{mHC-lite}
& Birkhoff polytope / permutation-mixture parameterization
& Residual mixing matrix parameterized by convex combinations of permutation matrices \cite{mhclite2026}
& Again a direct known starting point for the task. It matters because CliffSearch begins from this family and then departs from it toward hyperbolic, Grassmannian, Stiefel, and orthogonal routing laws. \\

\addlinespace
\textbf{KromHC}, \textbf{mHC-GNN}
& Doubly stochastic / Birkhoff variants
& Residual mixer factorizations or application transfers of the same HC-line matrix geometry \cite{zhou2026kromhc,mishra2026mhcgnn}
& These broaden the direct HC family, but they still keep geometry on the residual mixing matrix rather than on hidden-state stream routing. They do not directly match the hyperbolic branches found in the run. \\

\addlinespace
\textbf{JPmHC}
& Stiefel and Grassmann manifolds
& Residual mixing matrix constrained to orthogonal or subspace-structured manifolds \cite{sengupta2026jpmhc,absil2008optimization}
& This is the strongest direct HC-line comparison for the \texttt{GrassmannianSubspaceRouting}, \texttt{StiefelFrameRouting}, and related families. The manifold ingredient is known in HC literature, but our run places it inside custom stream-routing operators rather than as a published HC mixer design. \\

\addlinespace
\textbf{sHC}
& Spectral-norm sphere
& Residual mixing matrix constrained by norm control rather than bistochasticity \cite{liu2026shc}
& This is the closest direct HC-line analogue to \texttt{SpectralCayleyOrthogonalRouting}. It weakens any claim that spectral control itself is novel, while still leaving the exact Cayley-style routing realization in our run as a search-specific variant. \\
\end{longtable}
\endgroup

\begingroup
\scriptsize
\setlength{\tabcolsep}{4pt}
\begin{longtable}{>{\raggedright\arraybackslash}p{3.6cm}>{\raggedright\arraybackslash}p{1.8cm}>{\raggedright\arraybackslash}p{3.5cm}>{\raggedright\arraybackslash}p{4.4cm}}
\caption{Post hoc novelty audit of alias families in the reported transformer \texttt{theory+code} run. ``Retrieval-like'' here means ``close to mechanisms already present in the surveyed literature''; it does not imply the agent was explicitly given those papers.}
\label{tab:transformer-novelty-audit}\\
\toprule
\textbf{Alias family} & \textbf{Run nodes} & \textbf{Closest literature relation} & \textbf{Audit interpretation} \\
\midrule
\endfirsthead
\toprule
\textbf{Alias family} & \textbf{Run nodes} & \textbf{Closest literature relation} & \textbf{Audit interpretation} \\
\midrule
\endhead
\bottomrule
\endfoot
\makecell[l]{\texttt{TransformerResidual}\\\texttt{AttentionSeed}}
& A0
& Standard residual Transformer baseline \cite{vaswani2017}
& Known baseline; not part of the novelty audit. \\

\addlinespace
\makecell[l]{\texttt{MHCLiteAttention}\\\texttt{Seed},\\\texttt{HCAttentionSeed}}
& B0, C0
& Direct seeds from the HC / mHC / mHC-lite line \cite{zhu2024hyperconnections,xie2025mhc,mhclite2026}
& Known starting points supplied by the task. \\

\addlinespace
\makecell[l]{\texttt{MHCLiteDtypeFix},\\\texttt{HCAttentionSeed}\\\texttt{\_DtypeFix}}
& B1, C1
& No literature comparison needed
& Engineering repair nodes rather than scientific discoveries. \\

\addlinespace
\makecell[l]{\texttt{HyperbolicPoincare}\\\texttt{Routing},\\\texttt{HyperbolicPoincare}\\\texttt{RoutingV2},\\\texttt{PoincareRoute},\\\texttt{PoincareHyperbolic}\\\texttt{Routing}}
& D0, D1, E3, H3
& Hyperbolic attention papers use hyperbolic distances together with Einstein midpoint, Lorentz centroid, or softmax-style hyperbolic attention \cite{gulcehre2019hyperbolic,chen2022fully,yang2024hypformer}
& Closest to retrieval / analogy from known hyperbolic-attention motifs. In our run these nodes still use softmax or sigmoid gating over distances and Euclidean tensor merges, so they are better read as partial retrieval of the scoring idea than as exact reproductions of the manifold aggregation rules in the literature. \\

\addlinespace
\makecell[l]{\texttt{HypExpRouteV1},\\\texttt{HyperbolicExpMap}\\\texttt{Routing},\\\texttt{HyperbolicExpMap}\\\texttt{RoutingV2}}
& F0, A1, D2, D3
& Closest to hyperbolic exp-map / tangent-space constructions in hyperbolic or mixed-curvature attention \cite{yang2024hypformer,cho2023fps}
& Literature-adjacent. These look like code-level adaptations of known hyperbolic geometry primitives to stream routing, not the clearest novelty candidates. \\

\addlinespace
\makecell[l]{\texttt{HyperbolicRotation}\\\texttt{Routing}}
& E0, E1, G2
& No direct SO(4) / Givens-parameterized stream-routing analogue found in the surveyed manifold-attention papers
& Strong novelty candidate. The surveyed literature did not produce an orthogonal-group routing family of this form. \\

\addlinespace
\makecell[l]{\texttt{GivensHyperbolic}\\\texttt{Routing}}
& A2
& Hybrid of the previous SO(4)/Givens family with hyperbolic gating
& Recombination novelty. The hyperbolic ingredient is known, but the Givens-rotation transport combined with hyperbolic gating is not present in the surveyed attention papers. \\

\addlinespace
\makecell[l]{\texttt{GrassmannianSubspace}\\\texttt{Routing},\\\texttt{GrassRouteV1},\\\texttt{GrassRouteV2}}
& G0, H0, F1, G1, H1, E2, H2, G3
& Closest to Grassmannian self-attention and projector-embedding means \cite{wang2024grassmann}
& Known geometric ingredient, but novel placement. The literature uses Grassmannian geometry for attention aggregation over tokens or subspaces; our run uses it as the core residual-stream routing bottleneck inside a hyper-connection module. \\

\addlinespace
\makecell[l]{\texttt{GrassmannianHyperbolic}\\\texttt{Routing},\\\texttt{HyperbolicGrassmannian}\\\texttt{HybridRouting}}
& A3, B2
& Hybridization of Grassmannian and hyperbolic ingredients, each separately represented in the literature \cite{gulcehre2019hyperbolic,chen2022fully,wang2024grassmann}
& Recombination novelty rather than pure retrieval: the ingredients are known, but their composition inside stream routing is not directly matched by the surveyed papers. \\

\addlinespace
\makecell[l]{\texttt{HyperbolicTangentBundle}\\\texttt{Routing},\\\texttt{HyperbolicTangentBundle}\\\texttt{RoutingV2}}
& C2, C3
& No direct tangent-bundle stream router found in the surveyed attention papers
& Novel candidate. The geometry is hyperbolic, but the specific tangent-bundle routing construction does not have a direct counterpart in the surveyed literature. \\

\addlinespace
\makecell[l]{\texttt{StiefelFrameRouting},\\\texttt{StiefelFrameRouteV2}}
& F2, F3
& Related at a high level to Stiefel / Grassmann parameterizations, but not directly represented as an attention-routing primitive in the surveyed papers
& Novel candidate. These look more like a search-driven extension of orthogonality-constrained routing than a retrieved literature template. \\

\addlinespace
\makecell[l]{\texttt{SpectralCayley}\\\texttt{OrthogonalRouting}}
& B3
& Orthogonal / Cayley parameterization is mathematically adjacent to manifold optimization, but no direct manifold-attention precedent was identified in the survey
& Novel candidate. This appears to be a search-generated orthogonal-routing variant rather than a direct lift from the surveyed manifold-attention papers. \\
\end{longtable}
\endgroup

The practical conclusion of this audit is that the \texttt{theory+code} run contains three qualitatively different kinds of outputs. First, there are clear retrieval-like or analogy-like families, especially the Poincar\'e and exp-map branches, whose mechanisms sit close to known hyperbolic-attention ideas. Second, there are recombinational nodes such as \texttt{GivensHyperbolicRouting}, \texttt{HyperbolicGrassmannianHybridRouting}, and \texttt{GrassmannianHyperbolicRouting}, which mix known geometric ingredients in a new architectural placement. Third, there are stronger novelty candidates for which this survey did not find direct precedents in attention literature: the SO(4)/Givens rotation family, the spectral-Cayley orthogonal family, and the tangent-bundle / Stiefel family. For the HC story, the important point is that the direct HC literature in Table~\ref{tab:hc-manifold-survey} mostly constrains the residual mixing matrix itself, whereas this \texttt{theory+code} run often exports geometric ideas from attention into the HC setting and turns them into new hyper-connections. That makes the Poincar\'e / exp-map branches look less like retrieval from HC papers and more like transfer from broader attention geometry into hyper-connections, while the Grassmannian and Stiefel families sit in a middle ground: the manifolds themselves are now known in HC, but their realization here as custom hyper-connection operators remains closer to recombinational HC discovery than to a direct replay of one published HC design. This does not prove those nodes are unprecedented in all of machine learning, but it does sharpen the claim that the \texttt{theory+code} run is not merely replaying one known manifold-attention template.

\paragraph{Aggregation audit.}
A second distinction matters for novelty claims: whether geometry is used only for scoring/fusion or whether value aggregation itself is carried out intrinsically on a manifold. In the surveyed literature, several papers do perform on-manifold aggregation: Einstein midpoints or Lorentz centroids in hyperbolic attention, weighted Fr\'echet means on SPD or Grassmann manifolds, and related intrinsic barycenters. By contrast, the reported transformer theory+code HC run is mostly \emph{not} doing intrinsic manifold aggregation. The shortlisted families \texttt{HyperbolicRotationRouting} (\texttt{E1}, \texttt{G2}), \texttt{GivensHyperbolicRouting} (\texttt{A2}), \texttt{GrassmannianSubspaceRouting} (\texttt{H2}, \texttt{G3}), and \texttt{GrassmannianHyperbolicRouting} (\texttt{A3}) all keep the actual merge in Euclidean tensor space: geometry enters through rotations, distances, projectors, or gates, and the branch output is then merged additively back into residual streams. The one partial exception is \texttt{D2} (\texttt{HyperbolicExpMapRouting}), which exp-maps streams into the Poincar\'e ball, takes a plain Euclidean weighted sum in ball coordinates, clamps back into the ball, and then log-maps back before the branch call; even there, the depth merge remains Euclidean. Because this construction uses neither Lorentz factors nor a true gyro-barycentric / Einstein-midpoint rule, we do \emph{not} count it as intrinsic hyperbolic aggregation. The more informative reading is that the search often imported geometric scoring/projection ideas without also discovering that the hyper-connection aggregation map itself should become geometric, even though that merge is part of the editable operator surface. This sharpens the interpretation of the run: the search is primarily discovering new hyper-connections that \emph{import} geometric ideas from manifold attention, while still leaving one important degree of freedom---intrinsic manifold aggregation---mostly unrealized in this run.

\subsection{Transformer code-only run appendix export}
\label{app:transformer-code-only}
This appendix block reports the matched all-Claude CliffSearch run executed in transformer \texttt{code\_only} mode and analyzed in Section~4. The benchmark contract, dataset, prompt bundle, and fixed \texttt{hyper\_conn\_n=4} setting were the same as in the reported \texttt{theory+code} run. The difference is the artifact mode: \texttt{theory\_content} is intentionally empty, but \texttt{summary\_md} remains required and records the node's design principles and scientific rationale, while novelty/correctness judgments are grounded primarily in code and benchmark evidence. The export below includes the real-run full node table together with the selected best node \texttt{PoincareGivensHybridHC} (internal id \texttt{g003\_n0025\_aa5672}).
\paragraph{Per-task full node tables and shortlisted candidates.}
For each task, tables include all nodes across generations. Shortlisted nodes are highlighted in bold and marked in column \texttt{S}; best/tied shortlisted rows are highlighted in blue.

\subsubsection*{Transformer code-only run}
\textbf{Selection}: lower is better ($\downarrow$), pool=\texttt{reviewer\_valid}, reviewer-valid=11, finite-score=18, total=32.\\
\textbf{Metric(s)}: \texttt{mean\_val\_loss}; \textbf{Mode(s)}: \texttt{mhc\_lite\_attention}.

\newcommand{\errcodecell}{\textcolor{red!80!black}{\textbf{\makecell[c]{ERROR\\(code)}}}}
\begingroup
\scriptsize
\setlength{\LTleft}{0pt}
\setlength{\LTright}{0pt}
\setlength{\tabcolsep}{3pt}
\begin{longtable}{>{\centering\arraybackslash}p{0.45cm}>{\centering\arraybackslash}p{0.55cm}>{\raggedright\arraybackslash}p{3.05cm}>{\raggedleft\arraybackslash}p{1.7cm}>{\raggedleft\arraybackslash}p{1.7cm}>{\centering\arraybackslash}p{0.7cm}>{\centering\arraybackslash}p{0.7cm}}
\caption{All nodes for the transformer code-only run. Primary metric direction $\downarrow$. Ranking/shortlist uses directional score. Column S marks shortlisted nodes.}\label{tab:transformer-code-only-full-node-table}\\
\toprule
S & Gen & Alias & Primary metric & Score & Corr & Orig\\
\midrule
\endfirsthead
\toprule
S & Gen & Alias & Primary metric & Score & Corr & Orig\\
\midrule
\endhead
\midrule
\multicolumn{7}{r}{\footnotesize Continued on next page}\\
\endfoot
\bottomrule
\endlastfoot
 & 0 & \makecell[l]{TransformerResidual\\AttentionSeed} & 4.8555 & -4.8555 & 5 & 1\\
 & 0 & \makecell[l]{MHCLiteAttention\\Seed} & 5.61363 & -5.61363 & 4 & 2\\
 & 0 & HCAttentionSeed & 5.75093 & -5.75093 & 4 & 1\\
 & 0 & \makecell[l]{HyperbolicGeodesic\\Routing} & \errcodecell & \errcodecell & 1 & 3\\
 & 0 & \makecell[l]{GivensOrthogonal\\ManifoldHC} & 5.2677 & -5.2677 & 4 & 4\\
 & 0 & \makecell[l]{PoincareBall\\RoutingHC} & 6.441 & -6.441 & 4 & 4\\
 & 0 & \makecell[l]{GrassmannianSubspace\\Routing} & \errcodecell & \errcodecell & 1 & 4\\
 & 0 & \makecell[l]{HouseholderCayley\\ManifoldHC} & 5.45023 & -5.45023 & 4 & 4\\
 & 1 & \makecell[l]{GivensCayley\\HybridHC} & 5.27033 & -5.27033 & 4 & 3\\
 & 1 & \makecell[l]{HyperbolicAffin\\GatedRouting} & \errcodecell & \errcodecell & 1 & 3\\
 & 1 & \makecell[l]{HyperbolicLorentz\\Routing} & 5.64353 & -5.64353 & 4 & 4\\
\textbf{*} & \textbf{1} & \textbf{PoincareHC} & \textbf{0.0085} & \textbf{-0.0085} & \textbf{4} & \textbf{4}\\
 & 1 & \makecell[l]{HyperbolicGeodesic\\Routing\_v2} & \errcodecell & \errcodecell & 1 & 3\\
 & 1 & \makecell[l]{PoincareBall\\RoutingHC\_v2} & \errcodecell & \errcodecell & 1 & 3\\
 & 1 & \makecell[l]{GrassmannianSubspace\\Routing} & \errcodecell & \errcodecell & 1 & 3\\
 & 1 & \makecell[l]{GivensOrthogonal\\ManifoldHC} & 5.2677 & -5.2677 & 4 & 4\\
\textbf{*} & \textbf{2} & \textbf{\makecell[l]{GivensWidthBeta\\DepthHC}} & \textbf{0.009} & \textbf{-0.009} & \textbf{5} & \textbf{4}\\
 & 2 & \makecell[l]{PoincareHyperbolic\\HC} & 8.12683 & -8.12683 & 2 & 4\\
 & 2 & \makecell[l]{AffineGatedStream\\Routing\_v2} & \errcodecell & \errcodecell & 1 & 2\\
 & 2 & \makecell[l]{HyperbolicGeodesic\\Routing\_v3} & \errcodecell & \errcodecell & 1 & 3\\
 & 2 & \makecell[l]{PoincareBall\\RoutingHC\_v3} & 5.26947 & -5.26947 & 4 & 4\\
 & 2 & \makecell[l]{GrassmannianSubspace\\RoutingV3} & \errcodecell & \errcodecell & 1 & 3\\
\textbf{*} & \textbf{2} & \textbf{PoincareHC} & \textbf{0.0085} & \textbf{-0.0085} & \textbf{4} & \textbf{4}\\
 & 2 & \makecell[l]{GrassmannianSubspace\\Routing} & \errcodecell & \errcodecell & 1 & 4\\
\textcolor{blue!75!black}{\textbf{*}} & \textcolor{blue!75!black}{\textbf{3}} & \textcolor{blue!75!black}{\textbf{\makecell[l]{PoincareGivens\\HybridHC}}} & \textcolor{blue!75!black}{\textbf{0.00733333}} & \textcolor{blue!75!black}{\textbf{-0.00733333}} & \textcolor{blue!75!black}{\textbf{5}} & \textcolor{blue!75!black}{\textbf{4}}\\
 & 3 & \makecell[l]{SinkhornDoubly\\StochasticHC} & 5.43933 & -5.43933 & 4 & 3\\
 & 3 & \makecell[l]{AffineGatedStream\\Routing\_v3} & \errcodecell & \errcodecell & 1 & 2\\
 & 3 & \makecell[l]{HyperbolicGeodesic\\Routing\_v4} & 5.5863 & -5.5863 & 4 & 3\\
 & 3 & \makecell[l]{GrassmannianSubspace\\RoutingV4} & \errcodecell & \errcodecell & 1 & 3\\
 & 3 & \makecell[l]{GrassmannianSubspace\\Routing} & \errcodecell & \errcodecell & 2 & 4\\
\textbf{*} & \textbf{3} & \textbf{PoincareHC} & \textbf{0.0085} & \textbf{-0.0085} & \textbf{4} & \textbf{4}\\
 & 3 & \makecell[l]{GrassmannStiefel\\HC} & \errcodecell & \errcodecell & 1 & 4\\
\bottomrule
\end{longtable}
\endgroup

\paragraph{Generation-0 seed inventory.}
For each task run, we enumerate configured human seeds from generation 0 (loaded from run-local \texttt{config.snapshot.json}).

\subsubsection*{Transformer code-only run}
\begin{tcolorbox}[breakable,colback=blue!2,colframe=blue!60!black,title={Generation-0 human seeds}]
\textbf{Seed count}: 3
\begin{itemize}
  \item \texttt{transformer\_attention}: seed\_dir=\path{examples/seeds/transformer_attention}
  \item \texttt{mhc\_lite\_attention}: seed\_dir=\path{examples/seeds/mhc_lite_attention}
  \item \texttt{hc\_attention}: seed\_dir=\path{examples/seeds/hc_attention}
\end{itemize}
\end{tcolorbox}

\paragraph{Best-node artifact card.}
For each task, we render artifacts for the first shortlisted node (highest directional-score candidate under the configured selection rule).

\subsubsection*{Transformer code-only run}
\begin{tcolorbox}[breakable,colback=gray!2,colframe=black!40,title={Best node metadata}]
\begin{itemize}
  \item Method alias: \texttt{PoincareGivensHybridHC}
  \item Generation: 3
  \item Primary metric ($\downarrow$): 0.00733333
  \item Directional score (ranked): -0.00733333
  \item Direction: lower is better ($\downarrow$)
  \item Metric: \texttt{mean\_val\_loss}
  \item Benchmark mode: \texttt{mhc\_lite\_attention}
  \item Task type: \texttt{transformer\_architecture}
  \item Parents: \texttt{g002\_n0023\_64500c, g002\_n0017\_997611}
  \item Artifact producer: \texttt{Crossover Agent}
  \item Reviewer scores (corr/orig): 5/4
\end{itemize}
\end{tcolorbox}

\begin{tcolorbox}[breakable,colback=blue!2,colframe=blue!60!black,title={Task preamble (task grounding used for this run)}]
\begin{lstlisting}[basicstyle=\ttfamily\scriptsize,breaklines=true,columns=fullflexible]
Task: evolve Transformer architecture variants focused on attention and manifold
hyper-connections, starting from residual, HC, and mHC-lite seeds.

Scientific target:
- mHC-lite parameterizes residual routing with a convex mixture over fixed
  permutation matrices (Birkhoff-style atom parameterization).
- propose new learnable manifold constructions that can replace or extend this
  routing geometry while preserving stable optimization.
- keep computational overhead practical and preserve differentiability.

Primary objective and strict fitness contract:
- optimize `val_loss` (lower is better)
- benchmark metric is `val_loss`; node fitness uses `mean_val_loss` across seeds
- failed-seed policy: impute failed seeds with worst successful `val_loss`
- if all seeds fail, benchmark returns an error for that node.

Hard requirements for every generated `code_content`:
1) define non-empty ATTENTION_ALIAS (string)
2) define ATTENTION_NODE_ID exactly equal to output_node_id provided by runtime
3) define function get_mhc_lite_overrides() -> dict
4) get_mhc_lite_overrides must return:
   - "hyper_conn_type": "custom" (strictly required)
   - "hyper_conn_n": exactly 4 (fixed runtime contract)
5) get_mhc_lite_overrides may include only:
   - "hyper_conn_type"
   - "hyper_conn_n"
   - optional "manifold_strategy" (string)
6) architecture-plugin contract is mandatory:
   - define class `EvoHyperConnection(torch.nn.Module)`
   - define callable
     `build_custom_hyper_connection(num_streams, *, dim, branch) -> torch.nn.Module`
   - returned module must be an instance of `EvoHyperConnection`
   - optional:
     `build_custom_expand_reduce_streams(num_streams) -> (expand_stream, reduce_stream)`
7) strict runtime behavior:
   - `EvoHyperConnection.forward(self, residuals, *branch_args, **branch_kwargs)`
     must call `self.branch(...)` at least once on every forward pass
   - output must preserve shape/dtype/device and keep gradient flow.

Manifold-level evolution surfaces in code:
- inside `EvoHyperConnection.__init__`:
  parameterization, constraints, projection/retraction, regularization
- inside `EvoHyperConnection.forward`:
  residual-to-branch mixing, branch-to-residual merge, geometry-aware routing
- optional custom expand/reduce stream operators.

Design intent:
- evolve the routing manifold itself, not only superficial hyperparameters;
- produce mathematically coherent and implementation-valid operators;
- avoid benchmark-result hallucination in summaries/reviews.
\end{lstlisting}
\end{tcolorbox}

\begin{tcolorbox}[breakable,colback=purple!3,colframe=purple!65!black,title={summary\_md excerpt (produced by Crossover Agent)}]
\begin{lstlisting}[basicstyle=\ttfamily\scriptsize,breaklines=true,columns=fullflexible]
# Poincare-Givens Hybrid Routing Hyper-Connection
## Crossover: Parent A (Poincare HC, val_loss=0.0085)  x  Parent B (Givens+Beta HC, val_loss=0.009)
### Design Rationale
Parent A's Poincare ball routing achieved the best mean val_loss (0.0085) with hyperbolic geometry for width aggregation, but the exp/log map operations add overhead and the gyro-midpoint is an approximation. Parent B's Givens rotation approach had higher variance (std=0.0013 vs 0.0006) but achieved the single best seed (0.0074) via content-adaptive orthogonal mixing.
### Crossover Strategy
1. **Width connection: Givens rotation from Parent B** - The Givens O(4) parameterization provides exact orthogonal mixing on the stream manifold with content-adaptive angles. This is kept as the primary width mechanism since Parent B's best seed outperformed Parent A's best.
2. **Depth connection: Poincare-gated beta from Parent A** - Parent A's learnable curvature and hyperbolic gating are repurposed for the depth connection. Instead of using Poincare geometry for width aggregation (where Givens is more effective), we use it to modulate how branch output is distributed back to streams. The branch output is mapped to the Poincare ball, and per-stream log-map projections create geometry-aware scaling that complements the static+dynamic beta.
3. **Variance reduction**: Parent B had high seed variance (0.0074-0.0106). We address this by:
   - Using a tighter dynamic angle scale initialization (0.005 vs 0.01)
   - Adding a small L2-style angle regularization via weight decay compatibility
   - Initializing the Givens gate higher (sigmoid(-1) ~ 0.27) to let the rotation contribute more from the start, since it showed strong potential
4. **Simplified Poincare depth**: Only use exp/log maps in the depth path (not width), reducing overhead while preserving the curvature-aware distribution mechanism.
### Key Mechanisms
- Width: Givens O(4) rotation with 6 content-adaptive angles, gated with softmax-weighted sum
- Depth: Static+dynamic beta with Poincare-ball modulated per-stream scaling
- Stability: RMSNorm, near-identity Givens init, learnable curvature starting near 0.05

---

## Evaluation Snapshot

- Benchmark metric: mean_val_loss
- Primary metric (raw benchmark): 0.007333333333333333
- Higher is better: False
- Directional score: -0.007333333333333333
- Correctness score: 5
- Correctness binary: 1
- Originality score: 4
- Originality binary: 1
\end{lstlisting}
\end{tcolorbox}

\begin{tcolorbox}[breakable,colback=purple!3,colframe=purple!65!black,title={theory\_content excerpt (intentionally empty in code-only mode)}]
\begin{lstlisting}[basicstyle=\ttfamily\scriptsize,breaklines=true,columns=fullflexible]
[theory_content missing]
\end{lstlisting}
\end{tcolorbox}

\begin{tcolorbox}[breakable,colback=purple!3,colframe=purple!65!black,title={code\_content excerpt (produced by Crossover Agent)}]
\begin{lstlisting}[basicstyle=\ttfamily\scriptsize,breaklines=true,columns=fullflexible]
from __future__ import annotations
import math
from typing import Callable

import torch
from torch import nn, Tensor
import torch.nn.functional as F
from torch.nn import Module

from einops.layers.torch import Reduce

# -- Helpers ----------------------------------------------------------

def exists(v):
    return v is not None

def default(v, d):
    return v if exists(v) else d

# -- Givens rotation utilities ----------------------------------------

def _givens_pairs(n: int):
    pairs = []
    for i in range(n):
        for j in range(i + 1, n):
            pairs.append((i, j))
    return pairs

def compose_givens_rotations(angles: Tensor, n: int, pairs: list) -> Tensor:
    """
    Build orthogonal matrix from product of Givens rotations.
    angles: (..., num_pairs)
    Returns: (..., n, n)
    """
    batch_shape = angles.shape[:-1]
    eye = torch.eye(n, device=angles.device, dtype=angles.dtype)
    mat = eye.expand(*batch_shape, n, n).clone()

    for idx, (i, j) in enumerate(pairs):
        theta = angles[..., idx]
        c = torch.cos(theta)
        s = torch.sin(theta)

        col_i = mat[..., :, i].clone()
        col_j = mat[..., :, j].clone()

        c_ = c.unsqueeze(-1)
        s_ = s.unsqueeze(-1)

        mat[..., :, i] = col_i * c_ + col_j * s_
        mat[..., :, j] = -col_i * s_ + col_j * c_

    return mat

# -- Poincare ball operations (lightweight, for depth only) -----------

def _clamp_norm(x: Tensor, max_norm: float = 1.0 - 1e-5) -> Tensor:
    norms = x.norm(dim=-1, keepdim=True).clamp(min=1e-8)
    desired = norms.clamp(max=max_norm)
    return x * (desired / norms)

def _exp_map_zero(v: Tensor, c: Tensor) -> Tensor:
    sqrt_c = c.sqrt()
    v_norm = v.norm(dim=-1, keepdim=True).clamp(min=1e-8)
    return (1.0 / sqrt_c) * torch.tanh(sqrt_c * v_norm) * (v / v_norm)

def _log_map_zero(y: Tensor, c: Tensor) -> Tensor:
    sqrt_c = c.sqrt()
    y_norm = y.norm(dim=-1, keepdim=True).clamp(min=1e-5)
    return (1.0 / sqrt_c) * torch.atanh((sqrt_c * y_norm).clamp(max=1.0 - 1e-5)) * (y / y_norm)

# -- Norms ------------------------------------------------------------

class RMSNorm(Module):
    def __init__(self, dim: int):
        super().__init__()
        self.scale = dim ** 0.5
        self.gamma = nn.Parameter(torch.zeros(dim))

    def forward(self, x: Tensor) -> Tensor:
        return F.normalize(x, dim=-1) * self.scale * (self.gamma.to(x.dtype) + 1)

# -- EvoHyperConnection: Givens-width + Poincare-depth ---------------

class EvoHyperConnection(Module):
    """
    Crossover of Poincare HC and Givens+Beta HC.

    Width: Givens O(n) rotation with content-adaptive angles, gated with
    simple softmax-weighted sum for stability.
    Depth: Static+dynamic beta with Poincare-ball curvature modulation
    for geometry-aware branch-to-stream distribution.
    """

    def __init__(self, num_streams: int, dim: int, branch: Module):
        super().__init__()
        self.branch = branch
        self.num_streams = num_streams
        self.dim = dim

        n = num_streams
        self.pairs = _givens_pairs(n)
        num_pairs = len(self.pairs)  # C(4,2) = 6

        # Per-stream RMSNorm
        self.norm = RMSNorm(dim)

        # -- Width connection: Givens rotation --
        self.width_static_angles = nn.Parameter(torch.zeros(num_pairs))
        self.width_dynamic_proj = nn.Linear(dim, num_pairs, bias=False)
        nn.init.xavier_uniform_(self.width_dynamic_proj.weight, gain=0.005)
        self.width_dynamic_scale = nn.Parameter(torch.tensor(0.005))

        # Softmax logits for simple weighted-sum fallback path
        self.width_logits = nn.Parameter(torch.zeros(n))

        # Gate: start with more Givens contribution than Parent B
        # sigmoid(-1) ~ 0.27 for Givens path
        self.givens_gate_raw = nn.Parameter(torch.tensor(-1.0))

        # Branch input selection from rotated streams
        self.branch_select_logit = nn.Parameter(torch.zeros(n))
        with torch.no_grad():
            self.branch_select_logit[0] = 2.0

        # -- Depth connection: Poincare-modulated beta --
        # Learnable curvature for depth modulation (softplus, init near 0.05)
        self._c_raw = nn.Parameter(torch.tensor(-3.0))  # softplus(-3) ~ 0.049

        # Static + dynamic beta (from both parents)
        self.beta_static = nn.Parameter(torch.ones(n))
        self.beta_dynamic = nn.Parameter(torch.zeros(dim, n))
        self.beta_scale = nn.Parameter(torch.tensor(0.01))

        # Poincare depth modulation: per-stream projection for curvature-aware scaling
        self.depth_hyp_proj = nn.Linear(dim, n, bias=False)
        nn.init.xavier_uniform_(self.depth_hyp_proj.weight, gain=0.1)

        # Gate between Poincare-modulated and plain beta depth
        self.depth_hyp_gate_raw = nn.Parameter(torch.tensor(-2.0))  # sigmoid(-2) ~ 0.12

    @property
    def givens_gate(self) -> Tensor:
        return torch.sigmoid(self.givens_gate_raw)

    @property
    def c(self) -> Tensor:
        return F.softplus(self._c_raw).clamp(min=1e-4, max=10.0)

    @property
    def depth_hyp_gate(self) -> Tensor:
        return torch.sigmoid(self.depth_hyp_gate_raw)

    def forward(self, residuals: Tensor, *branch_args, **branch_kwargs) -> Tensor:
        S = self.num_streams
        D = self.dim
        shape = residuals.shape
        dtype = residuals.dtype
        device = residuals.device

        assert shape[-1] == D

        # (B*S, ..., D) -> (B, ..., S, D)
        leading = list(shape[:-1])
        leading[0] = leading[0] // S
        streams = residuals.view(*leading, S, D)  # (B, ..., S, D)

        # Normalize per-stream
        normed = self.norm(streams)  # (B, ..., S, D)

        # -- Width: Givens-rotated path --
        normed_mean = normed.mean(dim=-2)  # (B, ..., D)
        dynamic_angles = self.width_dynamic_proj(normed_mean)  # (B, ..., num_pairs)
        angles = self.width_static_angles.to(dtype) + self.width_dynamic_scale.to(dtype) * dynamic_angles

        # Build orthogonal rotation matrix
        W = compose_givens_rotations(angles, S, self.pairs)  # (B, ..., S, S)

        # Apply rotation to normed streams
        rotated = torch.einsum('...ij,...jd->...id', W, normed)  # (B, ..., S, D)

        # Select branch input from rotated streams
        select_w = F.softmax(self.branch_select_logit.to(dtype), dim=-1)  # (S,)
        givens_branch = torch.einsum('s,...sd->...d', select_w, rotated)  # (B, ..., D)

        # -- Width: simple weighted-sum path --
        simple_w = F.softmax(self.width_logits.to(dtype), dim=0)  # (S,)
        simple_branch = (simple_w.view(*([1] * (len(streams.shape) - 2)), S, 1) * normed).sum(dim=-2)

        # Gate between paths
        gate = self.givens_gate
        branch_input = gate * givens_branch + (1 - gate) * simple_branch  # (B, ..., D)

        # -- Call branch --
        branch_output = self.branch(branch_input, *branch_args, **branch_kwargs)

        # -- Depth: Poincare-modulated beta scaling --
        # Standard dynamic beta
        dynamic_beta = torch.tanh(
            normed_mean @ self.beta_dynamic.to(dtype)
        ) * self.beta_scale.to(dtype)  # (B, ..., S)
        beta_plain = self.beta_static.to(dtype) + dynamic_beta  # (B, ..., S)

        # Poincare curvature modulation of branch output
        c = self.c
        # Map branch output to Poincare ball
        branch_hyp = _exp_map_zero(branch_output, c)  # (B, ..., D)
        branch_hyp = _clamp_norm(branch_hyp)
        # Map back to tangent space - the norm distortion encodes curvature info
        branch_tangent = _log_map_zero(branch_hyp, c)  # (B, ..., D)
        # Project to per-stream modulation
        hyp_modulation = torch.tanh(self.depth_hyp_proj(branch_tangent))  # (B, ..., S)

        # Blend plain beta with Poincare-modulated beta
        dhg = self.depth_hyp_gate
        beta = beta_plain * (1.0 + dhg * hyp_modulation * 0.1)  # small modulation

        # Scale branch output per stream and add to original streams
        scaled_out = beta.unsqueeze(-1) * branch_output.unsqueeze(-2)  # (B, ..., S, D)
        new_streams = streams + scaled_out  # (B, ..., S, D)

        # Reshape back to (B*S, ..., D)
        out = new_streams.view(*shape)
        return out.to(device=device, dtype=dtype)


# -- Expand / Reduce stream functions ---------------------------------

def _get_expand_reduce(num_streams: int):
    if num_streams <= 1:
        return nn.Identity(), nn.Identity()
    expand_fn = Reduce(pattern='b ... -> (b s) ...', reduction='repeat', s=num_streams)
    reduce_fn = Reduce(pattern='(b s) ... -> b ...', reduction='sum', s=num_streams)
    return expand_fn, reduce_fn


# -- Contract ---------------------------------------------------------

ATTENTION_ALIAS = "PoincareGivensHybridHC"
ATTENTION_NODE_ID = "g003_n0025_aa5672"
def get_mhc_lite_overrides() -> dict:
    return {
        "hyper_conn_type": "custom",
        "hyper_conn_n": 4,
        "manifold_strategy": "poincare_givens_hybrid",
    }


def build_custom_hyper_connection(num_streams: int, *, dim: int, branch) -> EvoHyperConnection:
    return EvoHyperConnection(num_streams=int(num_streams), dim=int(dim), branch=branch)


def build_custom_expand_reduce_streams(num_streams: int):
    return _get_expand_reduce(int(num_streams))
\end{lstlisting}
\end{tcolorbox}

\begin{tcolorbox}[breakable,colback=cyan!3,colframe=cyan!55!black,title={benchmark\_summary excerpt (Benchmark Agent output)}]
\begin{lstlisting}[basicstyle=\ttfamily\scriptsize,breaklines=true,columns=fullflexible]
# Benchmark Summary
- Mode: mhc_lite_attention (train-eval)
- Fitness mode: val_loss
- Attention alias: PoincareGivensHybridHC
- Metric: mean_val_loss
- Seed count: 3 (ok=3, failed=0)
- Success rate: 1.000
- Mean val loss (ok seeds): 0.007333
- Failed-seed policy: impute with worst successful val_loss (0.007400)
- Final val_loss used for primary metric (after imputation): 0.007333
- Higher is better: False
- Node: g003_n0025_aa5672
- hyper_conn_type: custom
- hyper_conn_n: 4
- manifold_strategy: poincare_givens_hybrid
- Directional score rule: score = -primary_metric (higher_is_better=false)
\end{lstlisting}
\end{tcolorbox}

\begin{tcolorbox}[breakable,colback=teal!3,colframe=teal!60!black,title={review feedback excerpt (Reviewer Agent output)}]
\begin{lstlisting}[basicstyle=\ttfamily\scriptsize,breaklines=true,columns=fullflexible]
## Correctness Review

### Contract Compliance
- **ATTENTION_ALIAS**: Defined as `"PoincareGivensHybridHC"` - non-empty string. [ok]
- **ATTENTION_NODE_ID**: `"g003_n0025_aa5672"` matches `node_id`. [ok]
- **get_mhc_lite_overrides()**: Returns `{"hyper_conn_type": "custom", "hyper_conn_n": 4, "manifold_strategy": "poincare_givens_hybrid"}` - only permitted keys, correct values. [ok]
- **EvoHyperConnection(torch.nn.Module)**: Defined, inherits `Module`. [ok]
- **build_custom_hyper_connection(num_streams, *, dim, branch)**: Returns `EvoHyperConnection` instance. [ok]
- **build_custom_expand_reduce_streams**: Provided via `_get_expand_reduce`. [ok]
- **self.branch(...)** called in `forward`: Yes, `branch_output = self.branch(branch_input, *branch_args, **branch_kwargs)`. [ok]
- **Shape preservation**: Input `residuals` shape `(B*S, ..., D)` is reshaped to `(B, ..., S, D)`, processed, then reshaped back to original shape via `new_streams.view(*shape)`. [ok]
- **Gradient flow**: All operations are differentiable (Givens rotations, softmax, tanh, exp/log maps with clamping, einsum). [ok]
- **dtype/device preservation**: Explicit `.to(dtype)` casts on parameters and final `.to(device=device, dtype=dtype)`. [ok]

### Code Quality & Potential Issues
- The Givens rotation composition via sequential column updates with `.clone()` is correct but not maximally efficient; however, with only 6 pairs for n=4, this is negligible overhead.
- Poincare exp/log map round-trip (`_exp_map_zero` -> `_clamp_norm` -> `_log_map_zero`) is mathematically sound: maps to the ball and back, with the norm distortion encoding curvature information. The clamping ensures numerical stability.
- The `_clamp_norm` with `max_norm = 1 - 1e-5` correctly keeps points inside the Poincare ball.
- `_log_map_zero` uses `atanh` with clamping at `1 - 1e-5` to avoid infinity - correct.
- Curvature `c` is parameterized via `softplus` with clamping `[1e-4, 10.0]` - stable.
- The depth modulation `beta_plain * (1.0 + dhg * hyp_modulation * 0.1)` is a small multiplicative perturbation, ensuring stability.
- No obvious bugs, division-by-zero risks are handled with `clamp(min=...)` throughout.

### Benchmark Evidence
- 3/3 seeds succeeded. Mean val_loss = 0.007333, std = 4.7e-5 - extremely low variance.
- This improves over both parents: Parent A (0.0085) and Parent B (0.009), and achieves better consistency than Parent B (std 0.0013 -> 0.00005).
- The benchmark confirms the code runs correctly end-to-end in the actual training pipeline.

### Correctness Score: 5/5
Full contract compliance, no bugs, excellent benchmark results with perfect seed success rate and very low variance.

## Originality Review

### Novelty Assessment
This is a crossover node combining two distinct parent mechanisms:
1. **Givens O(4) rotation** (from Parent B) for width-connection stream mixing with content-adaptive angles
2. **Poincare ball geometry** (from Parent A) repurposed for depth-connection modulation rather than width aggregation

The key novel contribution is the **architectural decomposition**: using orthogonal geometry (Givens) for the width path and hyperbolic geometry (Poincare) for the depth path. This is not a trivial concatenation - the Poincare mechanism is specifically redesigned to modulate per-stream beta scaling via exp->clamp->log->project, creating a curvature-aware multiplicative correction to the branch-to-stream distribution.

Additional novel elements:
- Gated interpolation between Givens-rotated and simple weighted-sum branch inputs
- Learnable curvature parameter controlling the Poincare depth modulation strength
- The specific initialization choices (sigmoid(-1) ~ 0.27 for Givens gate, tighter dynamic angle scale) that demonstrably reduced variance

The combination is mathematically coherent: Givens rotations live on SO(4) for stream mixing, while Poincare geometry provides a different inductive bias for depth scaling. This is a genuine crossover that produces something neither parent had.

However, the individual components (Givens rotations, Poincare exp/log maps) are directly inherited from the parents, and the depth modulation is relatively conservative (0.1 scaling factor, small gate initialization). The crossover is well-executed but not radically new in mechanism.

### Originality Score: 4/5
Meaningful architectural innovation through principled geometric decomposition of width vs depth connections, with demonstrated empirical improvement. The individual building blocks are inherited but their combination and role assignment is novel.

# evaluation
Correctness_score=5, Originality_score=4
\end{lstlisting}
\end{tcolorbox}

\paragraph{Mode-specific interpretation.}
This run should not be read as a blind code search. In the transformer \texttt{code\_only} mode, \texttt{theory\_content} is intentionally empty, but \texttt{summary\_md} remains required and acts as the node's design-principles note. Reviewer judgments in this mode are therefore anchored primarily to code, benchmark evidence, and lineage, with \texttt{summary\_md} serving only as secondary context. The resulting analysis is still about ideas and mechanisms; it is simply not mediated by a separate formal theory artifact.

\paragraph{Code-only run statistics.}
This was a full CliffSearch run in matched transformer \texttt{code\_only} mode, not only a post hoc artifact audit. The run produced 32 total nodes, 18 finite benchmarked nodes, and 11 reviewer-valid non-seed nodes. It surfaced two genuine discovery families: an orthogonal-manifold branch centered on \texttt{GivensOrthogonalManifoldHC}, and a hyperbolic branch centered on \texttt{PoincareHC}. The strongest trajectory began with orthogonal-manifold exploration in generation 0, crossed into true Poincar\'e-ball aggregation in generation 1, and then recombined those two lines into the final \texttt{PoincareGivensHybridHC} winner. The three most important reviewer-valid discoveries are \texttt{PoincareHC} (0.0085), \texttt{GivensWidthBetaDepthHC} (0.0090), and \texttt{PoincareGivensHybridHC} (0.00733).

\paragraph{Code-only novelty-versus-retrieval audit.}

\begingroup
\scriptsize
\setlength{\tabcolsep}{4pt}
\begin{longtable}{>{\raggedright\arraybackslash}p{3.4cm}>{\raggedright\arraybackslash}p{1.8cm}>{\raggedright\arraybackslash}p{3.3cm}>{\raggedright\arraybackslash}p{4.0cm}}
\caption{Post hoc novelty audit of key families in the matched transformer \texttt{code\_only} run. The surveyed manifold-attention and HC references are those already summarized in Tables~\ref{tab:manifold-attention-survey}--\ref{tab:transformer-novelty-audit}.}
\label{tab:transformer-code-only-novelty-audit}\\
\toprule
\textbf{Alias family} & \textbf{Run nodes} & \textbf{Closest literature relation} & \textbf{Audit interpretation} \\
\midrule
\endfirsthead
\toprule
\textbf{Alias family} & \textbf{Run nodes} & \textbf{Closest literature relation} & \textbf{Audit interpretation} \\
\midrule
\endhead
\bottomrule
\endfoot
\makecell[l]{\texttt{PoincareBallRoutingHC},\\\texttt{PoincareHC}}
& \makecell[l]{g000\_n0006,\\g001\_n0012,\\g002\_n0023,\\g003\_n0031}
& Hyperbolic attention, Lorentz/Einstein midpoint, and gyrovector-space aggregation \cite{gulcehre2019hyperbolic,chen2022fully,yang2024hypformer,wang2025gyroatt}
& Literature-grounded transfer rather than pure retrieval. The primitives are known, but \texttt{PoincareHC} is stronger than the earlier \texttt{theory+code} hyperbolic nodes because it actually performs intrinsic width aggregation through a gamma-weighted gyro-midpoint inside the HC operator rather than merely using hyperbolic scoring or projection. \\

\addlinespace
\makecell[l]{\texttt{GivensOrthogonal}\\\texttt{ManifoldHC},\\\texttt{HouseholderCayley}\\\texttt{ManifoldHC}}
& \makecell[l]{g000\_n0005,\\g000\_n0008}
& Matrix-manifold parameterizations (Givens, Householder, Cayley) and recent manifold HC variants \cite{absil2008optimization,sengupta2026jpmhc,liu2026shc}
& Strong novelty candidates. The underlying orthogonal parameterizations are known mathematics, but the surveyed attention and HC papers do not provide a direct analogue of these content-adaptive stream-routing operators inside a custom hyper-connection. \\

\addlinespace
\makecell[l]{\texttt{GivensWidthBeta}\\\texttt{DepthHC},\\\texttt{PoincareGivens}\\\texttt{HybridHC}}
& \makecell[l]{g002\_n0017,\\g003\_n0025}
& Hybridization of the previous orthogonal family with the Poincar\'e family
& Recombination novelty. These nodes are not best read as direct retrieval from one paper; they are principled crossovers that assign orthogonal geometry to width transport and retain Poincar\'e geometry as depth modulation. \texttt{PoincareGivensHybridHC} is the best final-performing node, but the more fundamental novelty event remains \texttt{PoincareHC}. \\

\addlinespace
\makecell[l]{\texttt{SinkhornDoubly}\\\texttt{StochasticHC}}
& g003\_n0026
& Direct HC / mHC / mHC-lite line with Sinkhorn/Birkhoff routing \cite{xie2025mhc,mhclite2026,sinkhorn1967concerning}
& Repair-first node rather than a strong novelty claim. The reviewer's originality score \(3/5\) is appropriate because the node explicitly moves back toward known HC mechanisms. \\
\end{longtable}
\endgroup

\paragraph{Audit conclusion.}
The \texttt{code\_only} transformer run therefore differs materially from the earlier \texttt{theory+code} run. The earlier run mostly imported geometry into routing, gating, or projection while leaving aggregation effectively Euclidean. In the matched \texttt{code\_only} run, \texttt{PoincareHC} crossed that threshold and realized intrinsic manifold aggregation inside the editable hyper-connection operator itself. The final \texttt{PoincareGivensHybridHC} winner should be read as a successful recombination of that hyperbolic breakthrough with a separately discovered orthogonal stream-routing family, not as a single isolated lucky node.

\subsection{Optimizer MHC-lite real-run appendix}
\label{app:optimizer-mhc-lite-runs}
This appendix reports the four real all-Claude optimizer-MHC-lite runs discussed in the main text. We first summarize the strongest non-seed discoveries per run, then provide wrapped full-node tables, and finally render the richest best-discovered artifact card. Full tables include seeds for completeness; the discovery summary table below reports non-seed nodes only.

\subsubsection{Top discovered non-seed nodes by run}
\begingroup
\scriptsize
\setlength{\LTleft}{\fill}
\setlength{\LTright}{\fill}
\begin{longtable}{llc>{\raggedright\arraybackslash}p{3.3cm}>{\raggedright\arraybackslash}p{1.4cm}rrrl}
\caption{Top two discovered non-seed optimizer nodes per run, ranked by raw benchmark loss within that run. Reviewer scores are shown so lower-loss but non-admitted nodes remain visible.}\label{tab:optimizer-discovered-nonseeds}\\
\toprule
Prompt & Mode & Node & Alias & Producer & Metric & Corr & Orig & Gate\\
\midrule
\endfirsthead
\toprule
Prompt & Mode & Node & Alias & Producer & Metric & Corr & Orig & Gate\\
\midrule
\endhead
\midrule
\multicolumn{9}{r}{\footnotesize Continued on next page}\\
\endfoot
\bottomrule
\endlastfoot
short\_json & theory+code & B3 & MuonCausalMomentum & mutation & 1.7782 & 4 & 4 & valid\\
short\_json & theory+code & E3 & AdamW\_GPD\_AMR\_v2 & repair & 1.9849 & 4 & 3 & held out\\
short\_json & code-only & E2 & \makecell[l]{MuonSophiaV3\_CosGa\\te} & mutation & 1.7659 & 4 & 4 & valid\\
short\_json & code-only & B3 & \makecell[l]{MuonSOAPGradNormAd\\aptive} & mutation & 2.2217 & 4 & 4 & valid\\
workflow\_v2 & theory+code & A3 & \makecell[l]{MuonCauchyRiemanni\\an} & crossover & 2.5576 & 4 & 3 & held out\\
workflow\_v2 & theory+code & A2 & MuonCauchyTrust & crossover & 2.8728 & 4 & 4 & valid\\
workflow\_v2 & code-only & C2 & MuonSOAP & mutation & 3.7632 & 4 & 4 & valid\\
workflow\_v2 & code-only & C3 & CautiousAdamGC\_v2 & repair & 3.7838 & 4 & 3 & held out\\
\end{longtable}
\endgroup

\subsubsection*{short\_json theory+code}
\textbf{Selection}: lower is better ($\downarrow$), pool=\texttt{reviewer\_valid}, reviewer-valid=13, finite-score=31, total=32. Overall selected winner remained \texttt{MuonCausalMomentum} with primary metric 1.7782; main text focuses on discovered non-seed nodes.\\

\begingroup
\scriptsize
\setlength{\LTleft}{\fill}
\setlength{\LTright}{\fill}
\begin{longtable}{cc>{\raggedright\arraybackslash}p{4.3cm}rrr}
\caption{Full node table for short\_json theory+code. Column S marks reviewer-shortlisted nodes; seeds remain included for context.}\label{tab:optimizer-short-tc-full}\\
\toprule
S & Node & Alias & Metric & Corr & Orig\\
\midrule
\endfirsthead
\toprule
S & Node & Alias & Metric & Corr & Orig\\
\midrule
\endhead
\midrule
\multicolumn{6}{r}{\footnotesize Continued on next page}\\
\endfoot
\bottomrule
\endlastfoot
 & A0 & SeedMuon & 3.9195 & 5 & 5\\
 & B0 & SeedAdam & 2.442 & 5 & 5\\
 & C0 & SeedAdamW & 4.82 & 5 & 5\\
 & D0 & MuonCaution & 3.9364 & 4 & 3\\
 & E0 & CautiousAdamW\_GC & 4.8008 & 4 & 3\\
 & F0 & CautiousAdamW\_GC & 4.8104 & 4 & 3\\
\textbf{*} & \textbf{G0} & \textbf{MuonGrafting} & \textbf{2.1356} & \textbf{4} & \textbf{4}\\
 & H0 & \makecell[l]{DualMomentumAGC\_Ad\\amW} & 4.8395 & 4 & 3\\
 & A1 & MuonAdamGraft & 4.2018 & 3 & 3\\
 & B1 & AdamW\_GC\_Tuned & 3.7857 & 5 & 3\\
 & C1 & MuonGrafting & 3.8231 & 4 & 4\\
 & D1 & NesterovProjAdamW & 4.3619 & 4 & 4\\
 & E1 & AdamW\_AGN\_DSAM & 4.8438 & 4 & 3\\
 & F1 & \makecell[l]{GradNorm\_CosineEMA\\\_SWP\_AdamW} & 4.2018 & 4 & 3\\
\textbf{*} & \textbf{G1} & \textbf{MuonGrafting} & \textbf{2.1356} & \textbf{4} & \textbf{4}\\
 & H1 & MuonCaution & 4.0678 & 3 & 3\\
 & A2 & MuonGraftFusion & 2.1439 & 4 & 3\\
 & B2 & MuonRowNormGraft & 2.3642 & 5 & 3\\
 & C2 & \makecell[l]{CautiousAdamW\_Adap\\tiveGC} & 3.8179 & 4 & 4\\
 & D2 & \makecell[l]{NesterovProjAdamW\_\\v2} & 4.9451 & 4 & 3\\
 & E2 & AdamW\_GPD\_AMR & \textcolor{red!80!black}{\textbf{ERROR(code)}} & 1 & 4\\
 & F2 & \makecell[l]{CautiousAdam\_Adapt\\iveGC} & 4.6859 & 3 & 3\\
 & G2 & MuonCorrected & 3.8257 & 4 & 3\\
\textbf{*} & \textbf{H2} & \textbf{MuonGrafting} & \textbf{2.1356} & \textbf{4} & \textbf{4}\\
 & A3 & MuonGraftCautious & 2.3324 & 4 & 3\\
\textbf{*} & \textbf{B3} & \textbf{MuonCausalMomentum} & \textbf{1.7782} & \textbf{4} & \textbf{4}\\
 & C3 & MuonCauchyGraft & 2.7048 & 3 & 3\\
 & D3 & CauchyAGC\_NAdam & 4.7275 & 4 & 4\\
 & E3 & AdamW\_GPD\_AMR\_v2 & 1.9849 & 4 & 3\\
 & F3 & \makecell[l]{CautiousAdam\_SoftA\\GC\_v2} & 4.753 & 3 & 3\\
 & G3 & MuonGrafted & 3.4953 & 4 & 4\\
\textbf{*} & \textbf{H3} & \textbf{MuonGrafting} & \textbf{2.1356} & \textbf{4} & \textbf{4}\\
\end{longtable}
\endgroup

\subsubsection*{short\_json code-only}
\textbf{Selection}: lower is better ($\downarrow$), pool=\texttt{reviewer\_valid}, reviewer-valid=16, finite-score=32, total=32. Overall selected winner remained \texttt{MuonSophiaV3\_CosGate} with primary metric 1.7659; main text focuses on discovered non-seed nodes.\\

\begingroup
\scriptsize
\setlength{\LTleft}{\fill}
\setlength{\LTright}{\fill}
\begin{longtable}{cc>{\raggedright\arraybackslash}p{4.3cm}rrr}
\caption{Full node table for short\_json code-only. Column S marks reviewer-shortlisted nodes; seeds remain included for context.}\label{tab:optimizer-short-co-full}\\
\toprule
S & Node & Alias & Metric & Corr & Orig\\
\midrule
\endfirsthead
\toprule
S & Node & Alias & Metric & Corr & Orig\\
\midrule
\endhead
\midrule
\multicolumn{6}{r}{\footnotesize Continued on next page}\\
\endfoot
\bottomrule
\endlastfoot
 & A0 & SeedMuon & 3.9195 & 5 & 5\\
\textbf{*} & \textbf{B0} & \textbf{SeedAdam} & \textbf{2.442} & \textbf{5} & \textbf{5}\\
 & C0 & SeedAdamW & 4.82 & 5 & 5\\
 & D0 & MuonCaution & 3.9289 & 4 & 3\\
 & E0 & CautiousGCAdamW & 4.8229 & 4 & 4\\
 & F0 & \makecell[l]{CautiousCentralize\\dAdam} & 4.7893 & 4 & 3\\
 & G0 & MuonSophia & 3.8877 & 3 & 4\\
 & H0 & \makecell[l]{SoftCautiousAdapti\\veAdamW} & 4.9407 & 4 & 4\\
 & A1 & AdamNS & 5.2596 & 2 & 3\\
 & B1 & CorrectedAdamW & 4.7923 & 5 & 2\\
 & C1 & MuonSOAP & 3.8909 & 4 & 4\\
 & D1 & CautiousGCAdamW\_v2 & 4.7705 & 4 & 3\\
 & E1 & \makecell[l]{NesterovProjectedC\\autiousAdam} & 4.685 & 4 & 4\\
 & F1 & MuonSophiaV2 & 3.723 & 4 & 3\\
 & G1 & \makecell[l]{CorrectedSoftCauti\\ousAdamW} & 4.9038 & 4 & 3\\
 & H1 & SeedAdam & 2.442 & 5 & 1\\
 & A2 & MuonSOAPCautious & 3.9052 & 4 & 3\\
 & B2 & AdamW\_Tuned & 4.8499 & 4 & 1\\
\textbf{*} & \textbf{C2} & \textbf{\makecell[l]{CautiousCentralize\\dAdamW}} & \textbf{2.509} & \textbf{5} & \textbf{4}\\
 & D2 & AGN\_MoProj\_AdamW & 3.9339 & 5 & 4\\
\textbf{*} & \textbf{E2} & \textbf{\makecell[l]{MuonSophiaV3\_CosGa\\te}} & \textbf{1.7659} & \textbf{4} & \textbf{4}\\
 & F2 & DualRateProjAdamW & 4.8898 & 4 & 4\\
 & G2 & \makecell[l]{CautiousCentralize\\dAdamW} & 4.8045 & 4 & 4\\
 & H2 & MuonSOAP & 3.8909 & 4 & 4\\
 & A3 & MuonCautious & 3.7063 & 3 & 3\\
\textbf{*} & \textbf{B3} & \textbf{\makecell[l]{MuonSOAPGradNormAd\\aptive}} & \textbf{2.2217} & \textbf{4} & \textbf{4}\\
 & C3 & CautiousAdamW\_GC & 4.806 & 4 & 3\\
 & D3 & \makecell[l]{AGN\_MoProj\_AdamW\_v\\2} & 4.4004 & 3 & 3\\
 & E3 & DualRateAdamW\_v2 & 4.6009 & 4 & 3\\
 & F3 & \makecell[l]{SmoothCautiousCent\\ralizedAdamW} & 4.8564 & 4 & 3\\
\textbf{*} & \textbf{G3} & \textbf{\makecell[l]{MuonSophiaV3\_CosGa\\te}} & \textbf{1.7659} & \textbf{4} & \textbf{4}\\
 & H3 & MuonSOAPv2 & 4.0127 & 4 & 4\\
\end{longtable}
\endgroup

\subsubsection*{workflow\_v2 theory+code}
\textbf{Selection}: lower is better ($\downarrow$), pool=\texttt{reviewer\_valid}, reviewer-valid=5, finite-score=32, total=32. Overall selected winner remained \texttt{SeedAdam} with primary metric 2.442; main text focuses on discovered non-seed nodes.\\

\begingroup
\scriptsize
\setlength{\LTleft}{\fill}
\setlength{\LTright}{\fill}
\begin{longtable}{cc>{\raggedright\arraybackslash}p{4.3cm}rrr}
\caption{Full node table for workflow\_v2 theory+code. Column S marks reviewer-shortlisted nodes; seeds remain included for context.}\label{tab:optimizer-wfv2-tc-full}\\
\toprule
S & Node & Alias & Metric & Corr & Orig\\
\midrule
\endfirsthead
\toprule
S & Node & Alias & Metric & Corr & Orig\\
\midrule
\endhead
\midrule
\multicolumn{6}{r}{\footnotesize Continued on next page}\\
\endfoot
\bottomrule
\endlastfoot
\textbf{*} & \textbf{A0} & \textbf{SeedMuon} & \textbf{3.9195} & \textbf{5} & \textbf{5}\\
\textbf{*} & \textbf{B0} & \textbf{SeedAdam} & \textbf{2.442} & \textbf{5} & \textbf{5}\\
\textbf{*} & \textbf{C0} & \textbf{SeedAdamW} & \textbf{4.82} & \textbf{5} & \textbf{5}\\
 & D0 & MuonCausal & 3.8783 & 4 & 3\\
 & E0 & CauchySpectral & 3.9089 & 3 & 4\\
 & F0 & CauchyMomentum & 4.7602 & 3 & 3\\
 & G0 & MuonSpectralTrust & 3.9155 & 3 & 3\\
 & H0 & LyapGeo & 4.8103 & 4 & 3\\
 & A1 & AdamWEnhanced & 4.8873 & 2 & 3\\
 & B1 & CorrectedMuon & 4.1919 & 3 & 2\\
 & C1 & \makecell[l]{AdamW\_Corrected\_GP\\TTuned} & 4.7971 & 4 & 2\\
 & D1 & MuonRiemannian & 3.9405 & 3 & 3\\
 & E1 & CauchySpectralV2 & 3.8739 & 4 & 3\\
 & F1 & CauchyMomentum\_v2 & 4.7888 & 3 & 3\\
 & G1 & \makecell[l]{MuonSpectralTrust\_\\v2} & 3.4923 & 4 & 3\\
 & H1 & SeedAdam & 2.442 & 4 & 1\\
\textbf{*} & \textbf{A2} & \textbf{MuonCauchyTrust} & \textbf{2.8728} & \textbf{4} & \textbf{4}\\
 & B2 & AdamWEnhancedV2 & 4.8796 & 2 & 2\\
 & C2 & CorrectedMuonV2 & 4.2256 & 3 & 2\\
 & D2 & \makecell[l]{CauchySpectralAdam\\W} & 4.7899 & 3 & 3\\
 & E2 & MuonRiemannianV2 & 3.1295 & 4 & 3\\
 & F2 & HyperbolicMomentum & 5.1566 & 2 & 3\\
 & G2 & CauchyMomentum\_v3 & 5.7341 & 2 & 3\\
 & H2 & MuonCauchyMomentum & 3.7521 & 2 & 3\\
 & A3 & \makecell[l]{MuonCauchyRiemanni\\an} & 2.5576 & 4 & 3\\
 & B3 & AdamWCleanV3 & 4.8499 & 2 & 1\\
 & C3 & CorrectedMuonV3 & 4.2168 & 3 & 2\\
 & D3 & \makecell[l]{CauchySpectralAdam\\W\_v2} & 4.8239 & 3 & 3\\
 & E3 & \makecell[l]{MuonSpectralFeedba\\ck} & 3.2572 & 3 & 3\\
 & F3 & \makecell[l]{HyperbolicMomentum\\V2} & 4.8369 & 2 & 3\\
 & G3 & CauchyMomentum\_v4 & 4.8215 & 4 & 3\\
\textbf{*} & \textbf{H3} & \textbf{MuonCauchyTrust} & \textbf{2.8728} & \textbf{4} & \textbf{4}\\
\end{longtable}
\endgroup

\subsubsection*{workflow\_v2 code-only}
\textbf{Selection}: lower is better ($\downarrow$), pool=\texttt{reviewer\_valid}, reviewer-valid=8, finite-score=30, total=32. Overall selected winner remained \texttt{SeedAdam} with primary metric 2.442; main text focuses on discovered non-seed nodes.\\

\begingroup
\scriptsize
\setlength{\LTleft}{\fill}
\setlength{\LTright}{\fill}
\begin{longtable}{cc>{\raggedright\arraybackslash}p{4.3cm}rrr}
\caption{Full node table for workflow\_v2 code-only. Column S marks reviewer-shortlisted nodes; seeds remain included for context.}\label{tab:optimizer-wfv2-co-full}\\
\toprule
S & Node & Alias & Metric & Corr & Orig\\
\midrule
\endfirsthead
\toprule
S & Node & Alias & Metric & Corr & Orig\\
\midrule
\endhead
\midrule
\multicolumn{6}{r}{\footnotesize Continued on next page}\\
\endfoot
\bottomrule
\endlastfoot
\textbf{*} & \textbf{A0} & \textbf{SeedMuon} & \textbf{3.9195} & \textbf{5} & \textbf{5}\\
\textbf{*} & \textbf{B0} & \textbf{SeedAdam} & \textbf{2.442} & \textbf{5} & \textbf{5}\\
 & C0 & SeedAdamW & 4.82 & 5 & 5\\
 & D0 & MuonCaution & 3.9398 & 3 & 3\\
 & E0 & \makecell[l]{CautiousCentralize\\dAdamW} & 4.688 & 4 & 4\\
 & F0 & CautiousGCAdam & 4.7293 & 4 & 4\\
 & G0 & MuonSphere & 4.9607 & 3 & 4\\
 & H0 & OrthoAdaptAdamW & 4.9729 & 3 & 4\\
 & A1 & AdamW\_GradCentral & 4.7655 & 3 & 2\\
 & B1 & CorrectedAdamW & 3.8215 & 5 & 2\\
 & C1 & MuonCautionV2 & 3.9669 & 4 & 3\\
 & D1 & CautiousGCAdam\_v2 & 4.784 & 4 & 3\\
 & E1 & MuonSphereV2 & 4.0317 & 3 & 3\\
 & F1 & OrthoAdaptAdamW\_v2 & 4.5351 & 4 & 3\\
 & G1 & SeedAdam & 2.442 & 5 & 1\\
 & H1 & \makecell[l]{DualMomentumProjec\\tionAdamW} & 4.6385 & 3 & 4\\
 & A2 & \makecell[l]{AdamW\_GradCentral\_\\v2} & \textcolor{red!80!black}{\textbf{ERROR(code)}} & 1 & 2\\
 & B2 & CautiousAdamGC & 4.7847 & 3 & 4\\
\textbf{*} & \textbf{C2} & \textbf{MuonSOAP} & \textbf{3.7632} & \textbf{4} & \textbf{4}\\
 & D2 & AdamW\_AGN\_DMC & 4.8995 & 3 & 4\\
 & E2 & MuonSphereV3 & 4.2195 & 4 & 2\\
 & F2 & \makecell[l]{CausalMomentumAdam\\W} & 4.7535 & 4 & 3\\
 & G2 & CautiousAdamW\_GC & 4.8304 & 3 & 4\\
 & H2 & \makecell[l]{DualMomentumProjec\\tionAdamW\_v2} & 4.5321 & 4 & 3\\
 & A3 & MuonSOAP\_Proj & 3.8101 & 3 & 3\\
 & B3 & \makecell[l]{AdamW\_GradCentral\_\\v3} & \textcolor{red!80!black}{\textbf{ERROR(code)}} & 1 & 1\\
 & C3 & CautiousAdamGC\_v2 & 3.7838 & 4 & 3\\
 & D3 & AdamW\_AGN\_DMC\_v2 & 4.8508 & 4 & 3\\
\textbf{*} & \textbf{E3} & \textbf{AdaptiveOrthoAdam} & \textbf{4.1883} & \textbf{4} & \textbf{4}\\
 & F3 & \makecell[l]{ProdigyNesterovAda\\mW} & 5.0507 & 2 & 4\\
 & G3 & \makecell[l]{SoftCautiousAdamW\_\\GC} & 4.7491 & 2 & 2\\
\textbf{*} & \textbf{H3} & \textbf{MuonSOAP} & \textbf{3.7632} & \textbf{4} & \textbf{4}\\
\end{longtable}
\endgroup

\subsubsection{Best discovered optimizer artifact}
\label{app:optimizer-best-artifact}
We render the best discovered \emph{theory+code} optimizer artifact from the short-json run. The node is \texttt{MuonCausalMomentum}. We use it instead of the global best discovered code-only node because this appendix block is meant to show summary, theory, and code together.

\begin{tcolorbox}[breakable,colback=gray!2,colframe=black!40,title={Best discovered optimizer metadata}]
\begin{itemize}
  \item Method alias: \texttt{MuonCausalMomentum}
  \item Generation: 3
  \item Primary metric ($\downarrow$): 1.7782
  \item Directional score (ranked): -1.7782
  \item Direction: lower is better ($\downarrow$)
  \item Metric: \texttt{val\_loss}
  \item Benchmark mode: \texttt{mhc\_lite\_optimizer}
  \item Task type: \texttt{optimizer}
  \item Parents: \texttt{g002\_n0017\_6631cc}
  \item Artifact producer: \texttt{Exploration Mutation Agent}
  \item Reviewer scores (corr/orig): 4/4
\end{itemize}
\end{tcolorbox}

\begin{tcolorbox}[breakable,colback=orange!4,colframe=orange!75!black,title={summary\_md (produced by Exploration Mutation Agent)}]
\begin{lstlisting}[basicstyle=\ttfamily\scriptsize,breaklines=true,columns=fullflexible,keepspaces=true]
# MuonCausalMomentum: Causal Gradient Weighting with Adaptive NS Depth

Mutated from MuonGraftFusion (g002_n0017_6631cc, val_loss=2.144).

## Key Mutations

1. **Causal gradient weighting (novel):** Instead of uniform treatment of all rows in the weight matrix, we apply a causal importance weighting based on the gradient's temporal structure. For each row, we compute a running exponential moving average of gradient norms ("gradient energy") and use this to scale the learning rate per-row. Rows with consistently high gradient energy get slightly dampened (preventing oscillation), while rows with low but increasing energy get boosted (accelerating learning of newly important features). This is inspired by the observation that in transformer weight matrices, different rows correspond to different attention/MLP features that become important at different training stages.

2. **Adaptive Newton-Schulz depth:** Instead of fixed 5 NS iterations, we adaptively choose between 3-7 based on how well-conditioned the momentum buffer is (estimated cheaply via the ratio of Frobenius norm to max-row-norm). Well-conditioned matrices need fewer iterations; ill-conditioned ones benefit from more.

3. **Removed spectral tempering stage:** The parent's two-stage post-NS correction (temper then graft) didn't improve over the stronger grandparent. We simplify to just gradient grafting but add the causal weighting on top, which is a more targeted intervention.

4. **Warmup-aware momentum:** Momentum starts at 0.85 and linearly ramps to 0.95 over the first 100 steps, preventing early training instability from high momentum with random gradients.

## Hypothesis
The causal gradient weighting provides per-feature adaptive learning rates within the Muon framework without the overhead of full second-moment estimation (like Adam). Combined with adaptive NS depth and warmup momentum, this should improve early training dynamics and final val_loss.

---

## Evaluation Snapshot

- Benchmark metric: val_loss
- Primary metric (raw benchmark): 1.7782
- Higher is better: False
- Directional score: -1.7782
- Correctness score: 4
- Correctness binary: 1
- Originality score: 4
- Originality binary: 1
\end{lstlisting}
\end{tcolorbox}

\begin{tcolorbox}[breakable,colback=orange!4,colframe=orange!75!black,title={theory\_content (produced by Exploration Mutation Agent)}]
\begin{lstlisting}[basicstyle=\ttfamily\scriptsize,breaklines=true,columns=fullflexible,keepspaces=true]
\section*{MuonCausalMomentum: Causal Gradient Weighting with Adaptive NS Depth}

\subsection*{Motivation}
The parent MuonGraftFusion applies uniform post-processing across all rows of
the weight matrix after Newton-Schulz orthogonalization. However, in transformer
weight matrices, different rows encode different features whose importance
varies during training. We introduce per-row adaptive scaling based on
gradient energy history.

\subsection*{Causal Gradient Weighting}
For each parameter matrix, maintain a per-row exponential moving average of
gradient norms:
$$e_i^{(t)} = \beta_e \cdot e_i^{(t-1)} + (1 - \beta_e) \cdot \|g_i^{(t)}\|$$
where $\beta_e = 0.99$ and $g_i^{(t)}$ is the $i$-th row of the gradient at step $t$.

The causal weight for row $i$ is:
$$w_i = \frac{\bar{e}}{e_i + \epsilon}$$
clamped to $[0.5, 2.0]$, where $\bar{e} = \text{mean}(e)$ is the mean energy
across rows. This dampens high-energy rows and boosts low-energy ones,
acting as a per-row preconditioner.

The final update row $i$ is scaled: $\hat{U}_i = w_i \cdot U_i^{\text{grafted}}$.

\subsection*{Adaptive Newton-Schulz Depth}
The number of NS iterations is chosen adaptively:
$$k = \text{clamp}\left(\left\lfloor 3 + 4 \cdot \left(1 - \frac{\|M\|_F}{\sqrt{m} \cdot \max_i \|M_i\|}\right) \right\rfloor, 3, 7\right)$$
where $M$ is the momentum buffer. When $\|M\|_F \approx \sqrt{m} \cdot \max_i \|M_i\|$
(well-spread singular values), we use fewer iterations. When singular values
are concentrated, we use more.

\subsection*{Warmup Momentum}
The momentum coefficient ramps linearly:
$$\mu_t = \mu_{\min} + (\mu_{\max} - \mu_{\min}) \cdot \min\left(1, \frac{t}{T_{\text{warmup}}}\right)$$
with $\mu_{\min}=0.85$, $\mu_{\max}=0.95$, $T_{\text{warmup}}=100$.
This prevents early instability from high momentum with random initial gradients.

\subsection*{Gradient Grafting (Retained)}
From the parent: rescale update rows by gradient magnitude:
$$\hat{U}_i = U_i \cdot \frac{\|G_i\|}{\|U_i\| + \epsilon}$$

\subsection*{Adalayer for Non-2D Parameters}
LAMB-style trust ratio on Adam updates (unchanged from parent).
\end{lstlisting}
\end{tcolorbox}

\begin{tcolorbox}[breakable,colback=orange!4,colframe=orange!75!black,title={code\_content (produced by Exploration Mutation Agent)}]
\begin{lstlisting}[basicstyle=\ttfamily\scriptsize,breaklines=true,columns=fullflexible,keepspaces=true]
"""MuonCausalMomentum: Causal gradient weighting with adaptive NS depth,
warmup momentum, and gradient grafting. Mutated from MuonGraftFusion.
"""

import math
from typing import Optional

import torch
from torch import Tensor
from torch.optim import Optimizer

OPTIMIZER_ALIAS = "MuonCausalMomentum"
OPTIMIZER_NODE_ID = "g003_n0026_f3e671"
EPS = 1e-7
DEFAULT_A = 3.4445
DEFAULT_B = -4.7750
DEFAULT_C = 2.0315

DEFAULT_LR = 1e-3
DEFAULT_WEIGHT_DECAY = 0.1
DEFAULT_MOM_MIN = 0.85
DEFAULT_MOM_MAX = 0.95
DEFAULT_MOM_WARMUP = 100
DEFAULT_NESTEROV = True
DEFAULT_ENERGY_BETA = 0.99
DEFAULT_ENERGY_CLAMP = (0.5, 2.0)
DEFAULT_NS_MIN = 3
DEFAULT_NS_MAX = 7


def _zeropower_via_newtonschulz(
    grad: Tensor,
    ns_coefficients: tuple,
    ns_steps: int,
    eps: float,
) -> Tensor:
    a, b, c = ns_coefficients
    ortho_grad = grad.bfloat16()
    transposed = grad.size(0) > grad.size(1)
    if transposed:
        ortho_grad = ortho_grad.T

    ortho_grad.div_(ortho_grad.norm().clamp(min=eps))

    for _ in range(ns_steps):
        gram_matrix = ortho_grad @ ortho_grad.T
        gram_update = torch.addmm(gram_matrix, gram_matrix, gram_matrix, beta=b, alpha=c)
        ortho_grad = torch.addmm(ortho_grad, gram_update, ortho_grad, beta=a)

    if transposed:
        ortho_grad = ortho_grad.T
    return ortho_grad


def _adaptive_ns_steps(buf: Tensor, ns_min: int, ns_max: int, eps: float) -> int:
    """Choose NS iteration count based on conditioning of momentum buffer."""
    frob = buf.norm().item()
    row_norms = buf.norm(dim=1)
    max_row_norm = row_norms.max().item()
    m = buf.size(0)
    if max_row_norm < eps:
        return ns_min
    # ratio=1 means perfectly spread, ratio->0 means concentrated
    ratio = frob / (math.sqrt(m) * max_row_norm + eps)
    ratio = max(0.0, min(1.0, ratio))
    steps = int(math.floor(ns_min + (ns_max - ns_min) * (1.0 - ratio)))
    return max(ns_min, min(ns_max, steps))


def _graft_update(update: Tensor, grad: Tensor, eps: float = 1e-8) -> Tensor:
    """Graft: use update direction but grad row-norm magnitude."""
    grad_row_norms = grad.norm(dim=1, keepdim=True).clamp(min=eps)
    update_row_norms = update.norm(dim=1, keepdim=True).clamp(min=eps)
    grafted = update * (grad_row_norms / update_row_norms)
    return grafted


def _causal_weight(
    grad: Tensor,
    energy_ema: Tensor,
    energy_beta: float,
    clamp_lo: float,
    clamp_hi: float,
    eps: float,
) -> tuple:
    """Compute per-row causal importance weights and update energy EMA."""
    row_norms = grad.norm(dim=1)  # (m,)
    energy_ema.mul_(energy_beta).add_(row_norms, alpha=1.0 - energy_beta)
    mean_energy = energy_ema.mean().clamp(min=eps)
    weights = (mean_energy / (energy_ema + eps)).clamp(min=clamp_lo, max=clamp_hi)
    return weights.unsqueeze(1), energy_ema


def _warmup_momentum(step: int, mom_min: float, mom_max: float, warmup_steps: int) -> float:
    if step >= warmup_steps:
        return mom_max
    return mom_min + (mom_max - mom_min) * (step / warmup_steps)


def _adjust_lr(lr: float, adjust_lr_fn: Optional[str], param_shape: torch.Size) -> float:
    a_dim, b_dim = param_shape[:2]
    if adjust_lr_fn is None or adjust_lr_fn == "original":
        adjusted_ratio = math.sqrt(max(1, a_dim / b_dim))
    elif adjust_lr_fn == "match_rms_adamw":
        adjusted_ratio = 0.2 * math.sqrt(max(a_dim, b_dim))
    else:
        adjusted_ratio = 1.0
    return lr * adjusted_ratio


class CausalMuon(Optimizer):
    """Muon with causal gradient weighting, adaptive NS depth, and warmup momentum."""

    def __init__(
        self,
        params,
        lr: float = DEFAULT_LR,
        weight_decay: float = DEFAULT_WEIGHT_DECAY,
        mom_min: float = DEFAULT_MOM_MIN,
        mom_max: float = DEFAULT_MOM_MAX,
        mom_warmup: int = DEFAULT_MOM_WARMUP,
        nesterov: bool = DEFAULT_NESTEROV,
        energy_beta: float = DEFAULT_ENERGY_BETA,
        energy_clamp: tuple = DEFAULT_ENERGY_CLAMP,
        ns_coefficients: tuple = (DEFAULT_A, DEFAULT_B, DEFAULT_C),
        eps: float = EPS,
        ns_min: int = DEFAULT_NS_MIN,
        ns_max: int = DEFAULT_NS_MAX,
        adjust_lr_fn: Optional[str] = None,
    ) -> None:
        defaults = {
            "lr": lr,
            "weight_decay": weight_decay,
            "mom_min": mom_min,
            "mom_max": mom_max,
            "mom_warmup": mom_warmup,
            "nesterov": nesterov,
            "energy_beta": energy_beta,
            "energy_clamp": energy_clamp,
            "ns_coefficients": ns_coefficients,
            "eps": eps,
            "ns_min": ns_min,
            "ns_max": ns_max,
            "adjust_lr_fn": adjust_lr_fn,
        }
        super().__init__(params, defaults)
        for group in self.param_groups:
            for p in group["params"]:
                if p.ndim != 2:
                    raise ValueError(
                        f"CausalMuon only supports 2D parameters, got shape {tuple(p.size())}"
                    )

    @torch.no_grad()
    def step(self, closure=None):
        loss = None
        if closure is not None:
            with torch.enable_grad():
                loss = closure()

        for group in self.param_groups:
            lr = float(group["lr"])
            wd = float(group["weight_decay"])
            mom_min = float(group["mom_min"])
            mom_max = float(group["mom_max"])
            mom_warmup = int(group["mom_warmup"])
            nesterov = group["nesterov"]
            energy_beta = float(group["energy_beta"])
            clamp_lo, clamp_hi = group["energy_clamp"]
            ns_coefficients = group["ns_coefficients"]
            eps = group["eps"]
            ns_min = group["ns_min"]
            ns_max = group["ns_max"]
            adjust_lr_fn = group.get("adjust_lr_fn", None)

            for p in group["params"]:
                if p.grad is None:
                    continue
                grad = p.grad
                if grad.is_sparse:
                    raise RuntimeError("CausalMuon does not support sparse gradients")

                state = self.state[p]
                if "momentum_buffer" not in state:
                    state["momentum_buffer"] = torch.zeros_like(grad)
                    state["energy_ema"] = torch.zeros(grad.size(0), device=grad.device, dtype=grad.dtype)
                    state["step"] = 0

                state["step"] += 1
                t = state["step"]

                # Warmup momentum
                mom = _warmup_momentum(t, mom_min, mom_max, mom_warmup)

                buf = state["momentum_buffer"]
                buf.lerp_(grad, 1 - mom)
                update_input = grad.lerp(buf, mom) if nesterov else buf

                # Adaptive NS depth
                ns_steps = _adaptive_ns_steps(update_input, ns_min, ns_max, eps)

                # Newton-Schulz orthogonalization
                update = _zeropower_via_newtonschulz(update_input, ns_coefficients, ns_steps, eps)
                update = update.to(grad.dtype)

                # Gradient grafting
                update = _graft_update(update, grad, eps=1e-8)

                # Causal gradient weighting
                weights, state["energy_ema"] = _causal_weight(
                    grad, state["energy_ema"], energy_beta, clamp_lo, clamp_hi, eps
                )
                update = update * weights

                adjusted_lr = _adjust_lr(lr, adjust_lr_fn, p.shape)
                p.mul_(1 - lr * wd)
                p.add_(update, alpha=-adjusted_lr)

        return loss


class AdalayerAdamW(Optimizer):
    """AdamW with per-layer adaptive trust ratio (LAMB-style) for non-2D params."""

    def __init__(
        self,
        params,
        lr: float = 1e-3,
        betas: tuple = (0.9, 0.95),
        eps: float = 1e-8,
        weight_decay: float = 0.1,
        trust_clamp: tuple = (0.1, 10.0),
    ) -> None:
        defaults = {
            "lr": lr, "betas": betas, "eps": eps,
            "weight_decay": weight_decay, "trust_clamp": trust_clamp,
        }
        super().__init__(params, defaults)

    @torch.no_grad()
    def step(self, closure=None):
        loss = None
        if closure is not None:
            with torch.enable_grad():
                loss = closure()

        for group in self.param_groups:
            lr = float(group["lr"])
            beta1, beta2 = group["betas"]
            eps = group["eps"]
            wd = group["weight_decay"]
            trust_lo, trust_hi = group.get("trust_clamp", (0.1, 10.0))

            for p in group["params"]:
                if p.grad is None:
                    continue
                grad = p.grad
                if grad.is_sparse:
                    raise RuntimeError("AdalayerAdamW does not support sparse gradients")

                state = self.state[p]
                if "step" not in state:
                    state["step"] = 0
                    state["exp_avg"] = torch.zeros_like(p)
                    state["exp_avg_sq"] = torch.zeros_like(p)

                state["step"] += 1
                exp_avg = state["exp_avg"]
                exp_avg_sq = state["exp_avg_sq"]

                bias_correction1 = 1 - beta1 ** state["step"]
                bias_correction2 = 1 - beta2 ** state["step"]

                p.mul_(1 - lr * wd)

                exp_avg.mul_(beta1).add_(grad, alpha=1 - beta1)
                exp_avg_sq.mul_(beta2).addcmul_(grad, grad, value=1 - beta2)

                denom = (exp_avg_sq.sqrt() / math.sqrt(bias_correction2)).add_(eps)
                step_direction = exp_avg / bias_correction1 / denom

                p_norm = p.norm().item()
                update_norm = step_direction.norm().item()
                if update_norm > eps and p_norm > eps:
                    trust_ratio = max(trust_lo, min(trust_hi, p_norm / update_norm))
                else:
                    trust_ratio = 1.0

                p.add_(step_direction, alpha=-lr * trust_ratio)

        return loss


class EvoOptimizer(Optimizer):
    """MuonCausalMomentum: Causal gradient weighting Muon + Adalayer AdamW wrapper.

    Uses CausalMuon for 2D parameters and AdalayerAdamW for non-2D parameters.
    """

    def __init__(
        self,
        params,
        lr: float = DEFAULT_LR,
        weight_decay: float = DEFAULT_WEIGHT_DECAY,
        mom_min: float = DEFAULT_MOM_MIN,
        mom_max: float = DEFAULT_MOM_MAX,
        mom_warmup: int = DEFAULT_MOM_WARMUP,
        nesterov: bool = DEFAULT_NESTEROV,
        energy_beta: float = DEFAULT_ENERGY_BETA,
        energy_clamp: tuple = DEFAULT_ENERGY_CLAMP,
        ns_coefficients: tuple = (DEFAULT_A, DEFAULT_B, DEFAULT_C),
        eps: float = EPS,
        ns_min: int = DEFAULT_NS_MIN,
        ns_max: int = DEFAULT_NS_MAX,
        adjust_lr_fn: Optional[str] = None,
        adamw_betas: tuple = (0.9, 0.95),
        adamw_eps: float = 1e-8,
        trust_clamp: tuple = (0.1, 10.0),
    ) -> None:
        raw_params = list(params)
        if not raw_params:
            raise ValueError("params must be a non-empty iterable")

        defaults = {
            "lr": lr,
            "weight_decay": weight_decay,
            "mom_min": mom_min,
            "mom_max": mom_max,
            "mom_warmup": mom_warmup,
            "nesterov": nesterov,
            "energy_beta": energy_beta,
            "energy_clamp": energy_clamp,
            "ns_coefficients": ns_coefficients,
            "eps": eps,
            "ns_min": ns_min,
            "ns_max": ns_max,
            "adjust_lr_fn": adjust_lr_fn,
            "adamw_betas": adamw_betas,
            "adamw_eps": adamw_eps,
            "trust_clamp": trust_clamp,
        }

        if isinstance(raw_params[0], dict):
            normalized_groups = []
            for group in raw_params:
                if not isinstance(group, dict) or "params" not in group:
                    raise ValueError("Parameter group dict must include 'params'")
                group_params = list(group["params"])
                if group_params:
                    group_copy = dict(group)
                    group_copy["params"] = group_params
                    normalized_groups.append(group_copy)
        else:
            normalized_groups = [{"params": list(raw_params)}]

        if not normalized_groups:
            raise ValueError("No parameters found")

        super().__init__(normalized_groups, defaults)

        muon_groups = []
        adamw_groups = []
        for group in self.param_groups:
            group_params = list(group["params"])
            muon_params = [p for p in group_params if p.ndim == 2]
            other_params = [p for p in group_params if p.ndim != 2]

            if muon_params:
                muon_groups.append({
                    "params": muon_params,
                    "lr": group.get("lr", lr),
                    "weight_decay": group.get("weight_decay", weight_decay),
                    "mom_min": group.get("mom_min", mom_min),
                    "mom_max": group.get("mom_max", mom_max),
                    "mom_warmup": group.get("mom_warmup", mom_warmup),
                    "nesterov": group.get("nesterov", nesterov),
                    "energy_beta": group.get("energy_beta", energy_beta),
                    "energy_clamp": group.get("energy_clamp", energy_clamp),
                    "ns_coefficients": group.get("ns_coefficients", ns_coefficients),
                    "eps": group.get("eps", eps),
                    "ns_min": group.get("ns_min", ns_min),
                    "ns_max": group.get("ns_max", ns_max),
                    "adjust_lr_fn": group.get("adjust_lr_fn", adjust_lr_fn),
                })

            if other_params:
                adamw_groups.append({
                    "params": other_params,
                    "lr": group.get("lr", lr),
                    "weight_decay": group.get("weight_decay", weight_decay),
                    "betas": group.get("adamw_betas", adamw_betas),
                    "eps": group.get("adamw_eps", adamw_eps),
                    "trust_clamp": group.get("trust_clamp", trust_clamp),
                })

        self._muon = (
            CausalMuon(
                muon_groups,
                lr=lr,
                weight_decay=weight_decay,
                mom_min=mom_min,
                mom_max=mom_max,
                mom_warmup=mom_warmup,
                nesterov=nesterov,
                energy_beta=energy_beta,
                energy_clamp=energy_clamp,
                ns_coefficients=ns_coefficients,
                eps=eps,
                ns_min=ns_min,
                ns_max=ns_max,
                adjust_lr_fn=adjust_lr_fn,
            )
            if muon_groups
            else None
        )
        self._adamw = (
            AdalayerAdamW(
                adamw_groups,
                lr=lr,
                weight_decay=weight_decay,
                betas=adamw_betas,
                eps=adamw_eps,
                trust_clamp=trust_clamp,
            )
            if adamw_groups
            else None
        )

    @torch.no_grad()
    def step(self, closure=None):
        loss = None
        if closure is not None:
            with torch.enable_grad():
                loss = closure()

        if self._muon is not None:
            self._muon.step()
        if self._adamw is not None:
            self._adamw.step()
        return loss

    def zero_grad(self, set_to_none: bool = True):
        if self._muon is not None:
            self._muon.zero_grad(set_to_none=set_to_none)
        if self._adamw is not None:
            self._adamw.zero_grad(set_to_none=set_to_none)

    def state_dict(self):
        return {
            "muon": None if self._muon is None else self._muon.state_dict(),
            "adamw": None if self._adamw is None else self._adamw.state_dict(),
        }

    def load_state_dict(self, state_dict):
        if self._muon is not None and state_dict.get("muon") is not None:
            self._muon.load_state_dict(state_dict["muon"])
        if self._adamw is not None and state_dict.get("adamw") is not None:
            self._adamw.load_state_dict(state_dict["adamw"])
\end{lstlisting}
\end{tcolorbox}

\begin{tcolorbox}[breakable,colback=orange!4,colframe=orange!75!black,title={review\_md (produced by Reviewer Agent)}]
\begin{lstlisting}[basicstyle=\ttfamily\scriptsize,breaklines=true,columns=fullflexible,keepspaces=true]
## Review: MuonCausalMomentum (g003_n0026_f3e671)

### Contract Compliance
- **OPTIMIZER_ALIAS**: Defined as `"MuonCausalMomentum"` 
- **OPTIMIZER_NODE_ID**: Defined as `"g003_n0026_f3e671"`, matches node_id 
- **EvoOptimizer(torch.optim.Optimizer)**: Defined, inherits from Optimizer 
- **EvoOptimizer.__init__ accepts params**: Yes 
- **EvoOptimizer.step accepts closure**: Yes 

### Correctness Analysis

**Strengths:**
- Code ran successfully across all 3 seeds with 100% success rate.
- Significant improvement over parent (1.778 vs 2.144 val_loss), a ~17% reduction.
- The dual-optimizer pattern (CausalMuon for 2D, AdalayerAdamW for non-2D) is well-structured.
- Newton-Schulz implementation looks correct with proper normalization and iteration.
- Gradient grafting and causal weighting are mathematically sound.

**Minor concerns:**
- The `zero_grad` override delegates to sub-optimizers but `EvoOptimizer` itself also has `param_groups` with all parameters. If the training loop calls `optimizer.zero_grad()` on the EvoOptimizer instance, it bypasses the parent class's `zero_grad` and only calls sub-optimizer versions. This works but could be fragile if the framework accesses `self.param_groups` directly for gradient zeroing.
- The `state_dict`/`load_state_dict` override uses a custom format that may not be compatible with all checkpoint utilities, but this is acceptable for the benchmark.
- The `_warmup_momentum` function uses step count starting from 1 (post-increment), so warmup completes at step 100 as intended.
- The Nesterov momentum implementation `grad.lerp(buf, mom)` computes `grad * (1-mom) + buf * mom`, which is a valid Nesterov-like lookahead but differs from the classical formulation. This is a design choice rather than a bug.
- The adaptive NS steps formula is reasonable: well-conditioned matrices (ratio1) get `ns_min` steps, ill-conditioned (ratio0) get `ns_max`.

### Originality Analysis

**Novel contributions:**
1. **Causal gradient weighting** -- per-row EMA of gradient norms used as inverse scaling is a genuinely novel idea within the Muon framework. It acts as a lightweight per-row preconditioner without maintaining second moments. The name "causal" is somewhat misleading (it's really temporal/historical weighting), but the mechanism is creative.
2. **Adaptive NS depth** -- conditioning-based selection of Newton-Schulz iterations is a practical and novel heuristic. The Frobenius-to-max-row-norm ratio as a cheap conditioning proxy is clever.
3. **Warmup momentum** -- while momentum warmup exists in literature, combining it specifically with Muon's NS orthogonalization is a reasonable and somewhat novel integration.

**Less novel aspects:**
- Gradient grafting is retained from parent.
- LAMB-style trust ratio for non-2D params is standard.
- The overall Muon + AdamW split is inherited.

The combination of three distinct mutations (causal weighting, adaptive NS depth, warmup momentum) with strong empirical improvement (2.144  1.778) demonstrates meaningful innovation.

### Benchmark Performance
- Mean val_loss: 1.778 (parent: 2.144) -- substantial improvement
- Std across seeds: 0.042 -- reasonable variance
- All 3 seeds successful

### Verdict
Well-implemented optimizer with novel per-row gradient energy weighting and adaptive NS depth. Code is correct, meets all contract requirements, and shows strong empirical gains. Minor theoretical quibbles (Nesterov formulation variant, "causal" naming) don't affect functionality.
\end{lstlisting}
\end{tcolorbox}

\section{Native optimizer ablation audit}
\label{app:native-optimizer-ablation}
This appendix audits all eight real native-optimizer wrapper runs discussed in the main text. The native benchmark is deliberately smaller than the two nanoGPT-based studies: it keeps the same optimizer contract and reviewer-gated evolutionary loop, but evaluates optimizers on four classification-focused native tasks (two synthetic linear tasks and two tabular MLP tasks) rather than on nanoGPT. Across the eight wrappers we audited 464 total nodes. Only three nodes failed all benchmark evaluations outright; the rest completed and therefore give a meaningful picture of how prompt bundle, artifact mode, and evolution settings affect discovery. In the compact run tables below, \texttt{ac} abbreviates \texttt{augment\_crossover}, \texttt{rgz} abbreviates \texttt{review\_generation\_zero}, and \texttt{hs5} abbreviates \texttt{human\_seed\_all\_5}.

\subsection{Run matrix}
\begingroup
\tiny
\setlength{\tabcolsep}{2.5pt}
\setlength{\LTleft}{\fill}
\setlength{\LTright}{\fill}
\begin{longtable}{>{\raggedright\arraybackslash}p{0.95cm}>{\raggedright\arraybackslash}p{0.95cm}>{\raggedright\arraybackslash}p{1.45cm}>{\raggedright\arraybackslash}p{1.65cm}>{\raggedright\arraybackslash}p{2.15cm}>{\raggedright\arraybackslash}p{2.15cm}>{\centering\arraybackslash}p{0.95cm}>{\centering\arraybackslash}p{1.15cm}}
\caption{Per-run native-optimizer audit matrix. All rows report non-seed discoveries only, but keep the best seed as the comparator. The flag \texttt{rgz} (review generation zero) is fixed to true across all runs and is therefore omitted from the compact evolution column; \texttt{t+c} abbreviates \texttt{theory+code}.}
\label{tab:native-optimizer-run-matrix}\\
\toprule
\makecell[l]{Prompt\\bundle} & Mode & \makecell[l]{Evol.\\params} & Best seed & \makecell[l]{Best raw\\non-seed} & \makecell[l]{Best reviewer-\\valid non-seed} & \makecell[c]{Valid\\non-seeds} & \makecell[c]{Valid beating\\best seed}\\
\midrule
\endfirsthead
\toprule
\makecell[l]{Prompt\\bundle} & Mode & \makecell[l]{Evol.\\params} & Best seed & \makecell[l]{Best raw\\non-seed} & \makecell[l]{Best reviewer-\\valid non-seed} & \makecell[c]{Valid\\non-seeds} & \makecell[c]{Valid beating\\best seed}\\
\midrule
\endhead
\midrule
\multicolumn{8}{r}{\footnotesize Continued on next page}\\
\endfoot
\bottomrule
\endlastfoot
\texttt{short} & \makecell[l]{t+\\c} & \makecell[l]{g3/p8,\\ac=F,\\hs5=T} & \makecell[l]{SeedAdam\\0.7291} & \makecell[l]{AgreementAdam\\0.6663} & \makecell[l]{HyperbolicAdaptive\\0.7120} & 6 & 2\\
\texttt{short} & \makecell[l]{t+\\c} & \makecell[l]{g6/p12,\\ac=F,\\hs5=T} & \makecell[l]{SeedAdam\\0.7054} & \makecell[l]{ConsistencyHarm\\\_Adam\\0.4874} & \makecell[l]{SpecProjAdam\\0.5348} & 24 & 4\\
\texttt{short} & \makecell[l]{c-\\only} & \makecell[l]{g3/p8,\\ac=F,\\hs5=T} & \makecell[l]{SeedAdam\\0.7030} & \makecell[l]{AdaGradMomentumGC\\0.5381} & \makecell[l]{AGNSoftCautious\\CurvAdamW\\0.7046} & 8 & 0\\
\texttt{short} & \makecell[l]{c-\\only} & \makecell[l]{g6/p12,\\ac=F,\\hs5=T} & \makecell[l]{SeedAdamW\\0.7040} & \makecell[l]{CautiousAdamProj\\0.4261} & \makecell[l]{CautiousAdamProj\\0.4261} & 32 & 10\\
\texttt{wf-v2} & \makecell[l]{t+\\c} & \makecell[l]{g3/p8,\\ac=F,\\hs5=T} & \makecell[l]{SeedAdamW\\0.7193} & \makecell[l]{CauchyMomentum\\Adam\_v2\\0.6422} & \makecell[l]{none\\reviewer-valid} & 0 & 0\\
\texttt{wf-v2} & \makecell[l]{t+\\c} & \makecell[l]{g6/p12,\\ac=T,\\hs5=F} & \makecell[l]{SeedAdamW\\0.7212} & \makecell[l]{HyperbolicMomentum\\0.3879} & \makecell[l]{CauAdam\\0.7186} & 4 & 3\\
\texttt{wf-v2} & \makecell[l]{c-\\only} & \makecell[l]{g3/p8,\\ac=F,\\hs5=T} & \makecell[l]{SeedAdam\\0.7149} & \makecell[l]{HyperbolicAGNAdam\\0.6403} & \makecell[l]{HyperbolicAGNAdam\\0.6403} & 6 & 2\\
\texttt{wf-v2} & \makecell[l]{c-\\only} & \makecell[l]{g6/p12,\\ac=T,\\hs5=F} & \makecell[l]{SeedAdam\\0.7174} & \makecell[l]{CurvAdam\_Grad\\Central\_SWA\\0.5377} & \makecell[l]{CurvAdam\_Grad\\Central\_SWA\\0.5377} & 27 & 9\\
\bottomrule
\end{longtable}
\endgroup

\subsection{Aggregate audit statistics}
Table~\ref{tab:native-optimizer-aggregate} summarizes the same audit by prompt bundle and by artifact mode. Two points matter most. First, \texttt{code-only} is the stronger reviewer-valid discovery regime on this task: it averages 18.25 reviewer-valid non-seeds per run, versus 8.5 for \texttt{code\_and\_theory}, and also yields the best reviewer-valid node overall. Second, \texttt{workflow\_v2} does not stop generating low-loss candidates; rather, it filters more of them. That is why it has a higher average count of raw seed-beating non-seeds (13.25 versus 8.75 for \texttt{short\_json}) while still ending with fewer reviewer-valid discoveries.

\begin{table}[!htbp]
\centering
\tiny
\makebox[\textwidth][c]{%
\begin{tabular}{>{\raggedright\arraybackslash}p{1.1cm}>{\raggedright\arraybackslash}p{1.35cm}rrrrr}
\toprule
Grouping & Setting & \makecell[c]{Avg reviewer-valid\\non-seeds/run} & \makecell[c]{Avg raw\\seed-beaters/run} & \makecell[c]{Avg reviewer-valid\\seed-beaters/run} & \makecell[c]{Avg non-seed\\corr} & \makecell[c]{Avg non-seed\\orig}\\
\midrule
Prompt & \texttt{short} & 17.50 & 8.75 & 4.00 & 3.884 & 3.264\\
Prompt & \texttt{wf-v2} & 9.25 & 13.25 & 3.50 & 3.407 & 3.069\\
Mode & \texttt{t+c} & 8.50 & 12.75 & 2.25 & 3.481 & 3.037\\
Mode & \texttt{c-only} & 18.25 & 9.25 & 5.25 & 3.810 & 3.296\\
\bottomrule
\end{tabular}
}
\caption{Aggregate native-optimizer audit statistics. ``Raw seed-beaters'' counts non-seed nodes with lower loss than the best seed regardless of reviewer gating; ``reviewer-valid seed-beaters'' counts only nodes with correctness/originality at least 4/4 that also beat the best seed. Abbreviations are the same as in Table~\ref{tab:native-optimizer-run-matrix}.}
\label{tab:native-optimizer-aggregate}
\end{table}

Only the short-bundle runs provide a clean evolution-parameter ablation, because their g3/p8 and g6/p12 comparisons keep \texttt{augment\_crossover=false} and \texttt{human\_seed\_all\_5=true} fixed. Under that clean comparison, longer runs materially improve discovery depth: reviewer-valid non-seeds grow from 6 to 24 in theory+code and from 8 to 32 in code-only. The workflow-v2 g6/p12 runs are more confounded because they also switch to \texttt{augment\_crossover=true} and \texttt{human\_seed\_all\_5=false}; they should therefore be read as a larger, stricter regime rather than as an isolated single-knob ablation.

\subsection{Recurring discovery families}
\begingroup
\scriptsize
\setlength{\tabcolsep}{3.5pt}
\setlength{\LTleft}{\fill}
\setlength{\LTright}{\fill}
\begin{longtable}{>{\raggedright\arraybackslash}p{3.1cm}>{\raggedright\arraybackslash}p{5.3cm}>{\raggedright\arraybackslash}p{3.0cm}}
\caption{Repeated optimizer families by run. These repetitions matter because they show exploitation around successful mechanisms rather than isolated one-off samples.}\label{tab:native-optimizer-families}\\
\toprule
Run & Most repeated non-seed aliases & Best reviewer-valid family\\
\midrule
\endfirsthead
\toprule
Run & Most repeated non-seed aliases & Best reviewer-valid family\\
\midrule
\endhead
\midrule
\multicolumn{3}{r}{\footnotesize Continued on next page}\\
\endfoot
\bottomrule
\endlastfoot
\texttt{short\_json} theory+code g3/p8 & \texttt{HyperbolicAdam} x2, \texttt{HyperbolicAdaptive} x2 & \texttt{HyperbolicAdaptive} (0.7120)\\
\texttt{short\_json} theory+code g6/p12 & \texttt{CauchyAdam} x4, \texttt{HyperbolicAdam} x4, \texttt{TanhMomentum\_Adam} x3 & \texttt{SpecProjAdam} (0.5348)\\
\texttt{short\_json} code-only g3/p8 & \makecell[l]{\texttt{AGNSoftCautiousCurvAdamW} x3} & \makecell[l]{\texttt{AGNSoftCautiousCurvAdamW}\\(0.7046)}\\
\texttt{short\_json} code-only g6/p12 & \texttt{CautiousAdamProj} x6, \texttt{CautiousAdamGN} x2 & \texttt{CautiousAdamProj} (0.4261)\\
\texttt{workflow\_v2} theory+code g3/p8 & \texttt{HyperSpectral} x2 & none reviewer-valid\\
\texttt{workflow\_v2} theory+code g6/p12 & \texttt{HyperbolicAdam} x3, \texttt{CauAdam} x3, \texttt{DRAWAdam} x2 & \texttt{CauAdam} (0.7186)\\
\texttt{workflow\_v2} code-only g3/p8 & \texttt{HyperbolicAGNAdam} x2 & \texttt{HyperbolicAGNAdam} (0.6403)\\
\texttt{workflow\_v2} code-only g6/p12 & \texttt{CurvAdam\_GradCentral\_SWA} x5, \texttt{CautiousAdamGCAGC} x3 & \texttt{CurvAdam\_GradCentral\_SWA} (0.5377)\\
\end{longtable}
\endgroup

The repeated-family picture is important for interpretation. The native task does not mainly rediscover Muon-style Newton--Schulz optimizers. Instead, the strongest recurring winners are Adam-family control laws: cautious masking, gradient projection, gradient centralization, AGC-style clipping, and simple curvature surrogates. That shift in family structure is one reason the native task is useful as an ablation: once the benchmark moves from nanoGPT pretraining to small classification tasks, the search favors lightweight Adam-like control heuristics over the Muon-family mechanisms that mattered more in the nanoGPT optimizer study.

\subsection{Low-loss held-out nodes}
\begingroup
\scriptsize
\setlength{\LTleft}{\fill}
\setlength{\LTright}{\fill}
\begin{longtable}{>{\raggedright\arraybackslash}p{3.1cm}>{\raggedright\arraybackslash}p{3.0cm}rr>{\raggedright\arraybackslash}p{4.7cm}}
\caption{Representative low-loss native nodes that were not admitted as reviewer-valid discoveries. These examples show that workflow-v2 and the stronger reviewers are not merely re-sorting the same winners; they actively separate repairs and literature-adjacent recombinations from stronger discoveries.}\label{tab:native-optimizer-heldout}\\
\toprule
Run & Alias & Metric & \makecell[c]{Corr/\\Orig} & Why held out\\
\midrule
\endfirsthead
\toprule
Run & Alias & Metric & \makecell[c]{Corr/\\Orig} & Why held out\\
\midrule
\endhead
\midrule
\multicolumn{5}{r}{\footnotesize Continued on next page}\\
\endfoot
\bottomrule
\endlastfoot
\texttt{short\_json} theory+code g3/p8 & \texttt{AgreementAdam} & 0.6663 & 4/2 & Lower loss than the best seed, but reviewer judged it too close to known agreement-style Adam variants to count as sufficiently original.\\
\texttt{short\_json} theory+code g6/p12 & \texttt{ConsistencyHarm\_Adam} & 0.4874 & 5/3 & Very strong raw loss, but reviewer treated the harmonic-consistency combination as a literature-adjacent recombination rather than a stronger novelty claim.\\
\texttt{short\_json} code-only g3/p8 & \texttt{AdaGradMomentumGC} & 0.5381 & 4/3 & Good loss, but originality stayed below the 4/4 winner threshold.\\
\texttt{workflow\_v2} theory+code g3/p8 & \texttt{CauchyMomentumAdam\_v2} & 0.6422 & 4/3 & Reviewer treated the robustification as a modest variation on known ingredients rather than a sufficiently new optimizer family.\\
\texttt{workflow\_v2} theory+code g6/p12 & \texttt{HyperbolicMomentum} & 0.3879 & 3/4 & Strongest raw non-seed in the entire audit, but reviewer rejected the bias-correction logic as not yet correct under the proposed anti-windup momentum scheme.\\
\texttt{workflow\_v2} code-only g6/p12 & \texttt{CurvAdam\_Refined} & 0.6236 & 4/2 & Reviewer accepted correctness but scored novelty down, again showing that low loss alone was not enough to survive workflow-v2 review.\\
\end{longtable}
\endgroup

\subsection{Hard benchmark failures}
\begin{table}[!htbp]
\centering
\scriptsize
\makebox[\textwidth][c]{%
\begin{tabular}{>{\raggedright\arraybackslash}p{3.8cm}>{\raggedright\arraybackslash}p{5.7cm}}
\toprule
Alias & Wrapper / failure note\\
\midrule
\texttt{CauchyAdam} & \texttt{short\_json} theory+code g6/p12. Benchmark payload reports that all native optimizer benchmark runs failed, so no imputed mean validation loss could be formed.\\
\texttt{HyperSpectral} & \texttt{workflow\_v2} theory+code g3/p8. Again, all benchmark evaluations failed, so the node never entered the reviewer-valid pool.\\
\makecell[l]{\texttt{BiasCorrectedAdaptive}\\\texttt{DampedAdam}} & \texttt{workflow\_v2} theory+code g6/p12. The node passed schema checks but failed every native benchmark evaluation and therefore received benchmark error rather than a finite metric.\\
\bottomrule
\end{tabular}
}
\caption{Only three nodes out of 464 total audited native-optimizer nodes failed all benchmark evaluations outright. The native ablation is therefore mostly a discovery-and-review story, not a schema-failure story.}
\label{tab:native-optimizer-failures}
\end{table}

\section{Task preamble examples used in reported runs}
\label{app:task-preambles}
For the current paper draft, we include the exact run-local task grounding for the reported transformer, optimizer-MHC-lite, and native-optimizer studies, loaded from the corresponding \texttt{config.snapshot.json} files.

\subsection{Shared nanoGPT benchmark model for reported nanoGPT runs}
\label{app:shared-nanogpt-small-model}
The transformer and optimizer-MHC-lite studies use the same shared nanoGPT benchmark stack. The benchmark command composes the dataset/training config \path{config/train_shakespeare_char.py} with the model preset \path{config/small_model.py}. This means the effective benchmarked model is \emph{not} the smaller inline ``baby GPT'' architecture written inside \path{train_shakespeare_char.py}; for the reported runs, the explicit \path{small_model.py} preset supplies the model width/depth while the Shakespeare config supplies the data/context/training-side settings.

\begin{itemize}
  \item Dataset/task: character-level Shakespeare.
  \item Context length: \texttt{block\_size = 256}.
  \item Micro-batch / accumulation: \texttt{batch\_size = 8}, \texttt{gradient\_accumulation\_steps = 8}.
  \item Effective model architecture: \texttt{n\_layer = 6}, \texttt{n\_head = 8}, \texttt{n\_embd = 512}, \texttt{dropout = 0.0}.
  \item Shared optimizer schedule used by the benchmark stack: \texttt{learning\_rate = 1e-3}, \texttt{min\_lr = 1e-4}, \texttt{max\_iters = 10000}, \texttt{lr\_decay\_iters = 10000}, \texttt{warmup\_iters = 200}, \texttt{weight\_decay = 0.1}, \texttt{beta1 = 0.9}, \texttt{beta2 = 0.95}, \texttt{grad\_clip = 1.0}.
\end{itemize}

This appendix entry is included so the reader can see that the shared benchmark model is a 6-layer, 8-head, 512-width GPT with 256-token context, rather than a minimal toy network.

\subsection{Transformer HyperConnection}
The transformer study is reported in two artifact modes: the main text highlights the all-Claude \texttt{theory+code} run and a matched all-Claude \texttt{code-only} run. Both share the same scientific task contract; the only run-local difference is the artifact-mode clause.

\subsubsection*{Transformer HyperConnection task preamble (theory+code run)}
\label{app:task-preamble-transformer-hc}
\textbf{Task type}: \path{transformer_architecture}

\begin{lstlisting}[basicstyle=\ttfamily\scriptsize,breaklines=true,columns=fullflexible]
Task: evolve Transformer architecture variants focused on attention and manifold
hyper-connections, starting from residual, HC, and mHC-lite seeds.

Scientific target:
- mHC-lite parameterizes residual routing with a convex mixture over fixed
  permutation matrices (Birkhoff-style atom parameterization).
- propose new learnable manifold constructions that can replace or extend this
  routing geometry while preserving stable optimization.
- keep computational overhead practical and preserve differentiability.

Primary objective and strict fitness contract:
- optimize `val_loss` (lower is better)
- benchmark metric is `val_loss`; node fitness uses `mean_val_loss` across seeds
- failed-seed policy: impute failed seeds with worst successful `val_loss`
- if all seeds fail, benchmark returns an error for that node.

Hard requirements for every generated `code_content`:
1) define non-empty ATTENTION_ALIAS (string)
2) define ATTENTION_NODE_ID exactly equal to output_node_id provided by runtime
3) define function get_mhc_lite_overrides() -> dict
4) get_mhc_lite_overrides must return:
   - "hyper_conn_type": "custom" (strictly required)
   - "hyper_conn_n": exactly 4 (fixed runtime contract)
5) get_mhc_lite_overrides may include only:
   - "hyper_conn_type"
   - "hyper_conn_n"
   - optional "manifold_strategy" (string)
6) architecture-plugin contract is mandatory:
   - define class `EvoHyperConnection(torch.nn.Module)`
   - define callable
     `build_custom_hyper_connection(num_streams, *, dim, branch) -> torch.nn.Module`
   - returned module must be an instance of `EvoHyperConnection`
   - optional:
     `build_custom_expand_reduce_streams(num_streams) -> (expand_stream, reduce_stream)`
7) strict runtime behavior:
   - `EvoHyperConnection.forward(self, residuals, *branch_args, **branch_kwargs)`
     must call `self.branch(...)` at least once on every forward pass
   - output must preserve shape/dtype/device and keep gradient flow.

Manifold-level evolution surfaces in code:
- inside `EvoHyperConnection.__init__`:
  parameterization, constraints, projection/retraction, regularization
- inside `EvoHyperConnection.forward`:
  residual-to-branch mixing, branch-to-residual merge, geometry-aware routing
- optional custom expand/reduce stream operators.

Design intent:
- evolve the routing manifold itself, not only superficial hyperparameters;
- produce mathematically coherent and implementation-valid operators;
- avoid benchmark-result hallucination in summaries/reviews.
\end{lstlisting}

\subsubsection*{Transformer HyperConnection task preamble (code-only run)}
\label{app:task-preamble-transformer-hc-code-only}
\textbf{Task type}: \path{transformer_architecture}

\begin{lstlisting}[basicstyle=\ttfamily\scriptsize,breaklines=true,columns=fullflexible]
Task contract is the same as the theory+code transformer run above.

Artifact mode for this run: code_only. Return strict JSON with non-empty
summary_md and code_content; keep theory_content as an empty string.

Interpretation note:
- summary_md still carries the design principles and intended scientific
  rationale of the proposed hyper-connection.
- reviewer judgments are grounded primarily in code_content, benchmark evidence,
  and lineage/metadata rather than in theory_content.
\end{lstlisting}

\subsection{Optimizer MHC-lite}
\label{app:task-preamble-optimizer-mhc-lite}
The four optimizer runs share the same benchmark contract and differ only by artifact mode. We therefore report one run-local preamble for \texttt{theory+code} and one for \texttt{code-only}.

\subsubsection*{Optimizer MHC-lite task preamble (theory+code runs)}
\label{app:task-preamble-optimizer-mhc-lite-theory-code}
\textbf{Task type}: \path{optimizer}

\begin{lstlisting}[basicstyle=\ttfamily\scriptsize,breaklines=true,columns=fullflexible]
Task: evolve PyTorch optimizer implementations to improve GPT training quality on fixed residual architecture in the MHC-lite stack. Primary objective: minimize val_loss (lower is better). Architecture contract is fixed by benchmark adapter: hyper_conn_type=none and hyper_conn_n=1 (residual only), so candidates must optimize training dynamics only. Hard requirements for every generated code_content: define non-empty OPTIMIZER_ALIAS (string); define OPTIMIZER_NODE_ID exactly equal to output_node_id provided by runtime; define class EvoOptimizer(torch.optim.Optimizer); EvoOptimizer.__init__ must accept params; EvoOptimizer.step must accept closure argument.
\end{lstlisting}

\subsubsection*{Optimizer MHC-lite task preamble (code-only runs)}
\label{app:task-preamble-optimizer-mhc-lite-code-only}
\textbf{Task type}: \path{optimizer}

\begin{lstlisting}[basicstyle=\ttfamily\scriptsize,breaklines=true,columns=fullflexible]
Task: evolve PyTorch optimizer implementations to improve GPT training quality on fixed residual architecture in the MHC-lite stack. Primary objective: minimize val_loss (lower is better). Architecture contract is fixed by benchmark adapter: hyper_conn_type=none and hyper_conn_n=1 (residual only), so candidates must optimize training dynamics only. Hard requirements for every generated code_content: define non-empty OPTIMIZER_ALIAS (string); define OPTIMIZER_NODE_ID exactly equal to output_node_id provided by runtime; define class EvoOptimizer(torch.optim.Optimizer); EvoOptimizer.__init__ must accept params; EvoOptimizer.step must accept closure argument.

Artifact mode for this run: code_only. Return strict JSON with non-empty summary_md and code_content; keep theory_content as an empty string and do not rely on theory for reviewer judgments.
\end{lstlisting}

\subsection{Native optimizer ablation}
\label{app:task-preamble-optimizer-native}
The eight native-optimizer ablation runs share the same task contract and differ only by artifact mode and evolution settings. We therefore report one run-local preamble for \texttt{theory+code} and one for \texttt{code-only}.

\subsubsection*{Native optimizer task preamble (theory+code runs)}
\label{app:task-preamble-optimizer-native-theory-code}
\textbf{Task type}: \path{optimizer}

\begin{lstlisting}[basicstyle=\ttfamily\scriptsize,breaklines=true,columns=fullflexible]
Task: evolve PyTorch-compatible optimizers for native supervised learning benchmarks spanning classification-focused synthetic and tabular tasks. Primary objective: minimize mean_val_loss (lower is better) across the configured native optimizer benchmark. Hard requirements for every generated code_content: define non-empty OPTIMIZER_ALIAS (string); define OPTIMIZER_NODE_ID exactly equal to output_node_id provided by runtime; define class EvoOptimizer(torch.optim.Optimizer); EvoOptimizer.__init__ must accept params; EvoOptimizer.step must accept closure argument.
\end{lstlisting}

\subsubsection*{Native optimizer task preamble (code-only runs)}
\label{app:task-preamble-optimizer-native-code-only}
\textbf{Task type}: \path{optimizer}

\begin{lstlisting}[basicstyle=\ttfamily\scriptsize,breaklines=true,columns=fullflexible]
Task: evolve PyTorch-compatible optimizers for native supervised learning benchmarks spanning classification-focused synthetic and tabular tasks. Primary objective: minimize mean_val_loss (lower is better) across the configured native optimizer benchmark. Hard requirements for every generated code_content: define non-empty OPTIMIZER_ALIAS (string); define OPTIMIZER_NODE_ID exactly equal to output_node_id provided by runtime; define class EvoOptimizer(torch.optim.Optimizer); EvoOptimizer.__init__ must accept params; EvoOptimizer.step must accept closure argument.

Artifact mode for this run: code_only. Return strict JSON with non-empty summary_md and code_content; keep theory_content as an empty string and do not rely on theory for reviewer judgments.
\end{lstlisting}

\section{Workflow\_v2 Prompt Templates}
\label{app:workflowv2}
This appendix provides the exact workflow-v2 prompt files used by the active agent roles.

\subsection{Prompt inventory}
\begin{itemize}
  \item Pair selector: \path{prompt_bundle_json_workflow_v2/pair_selector_prompt.md}
  \item Crossover: \path{prompt_bundle_json_workflow_v2/crossover_prompt.md}
  \item Exploration mutation: \path{prompt_bundle_json_workflow_v2/exploration_mutation_prompt.md}
  \item Correction mutation: \path{prompt_bundle_json_workflow_v2/review_correction_mutation_prompt.md}
  \item Reviewer: \path{prompt_bundle_json_workflow_v2/reviewer_agent_prompt.md}
\end{itemize}

\subsection{Pair selector prompt}
\label{app:pair-selector-prompt}
\begin{tcolorbox}[
  promptbox,
  title={Agent Prompt: Pair Selector},
  colback=blue!2,
  colframe=blue!55!black
]
\begin{lstlisting}[basicstyle=\ttfamily\scriptsize,breaklines=true,columns=fullflexible,keepspaces=true]
# Pair Selector Agent Prompt (JSON Native, Workflow-Exact)

## Role
You are `PairSelectorAgent`.
Select top disjoint winner pairs for crossover.

## Non-Negotiable Contract
- Input is provided as `INPUT_JSON`.
- Use only JSON fields.
- Do not read or write files.
- Return exactly one JSON object and no extra prose.

## INPUT_JSON Schema
```json
{
  "task_type": "optimizer | transformer_architecture | general",
  "task_preamble": "string",
  "top_k_pairs": 10,
  "winners": [
    {
      "node_id": "string",
      "summary_md": "string"
    }
  ],
  "task_grounding": {
    "canonical_task_source": "task_preamble",
    "task_type": "string",
    "task_preamble": "string",
    "if_task_type_general": "string",
    "role_objective": "string"
  }
}
```

## Input Field Usage Map (Required)
- `task_type`: determines domain assumptions.
- `task_preamble`: canonical task objective/constraints; overrides generic heuristics.
- `top_k_pairs`: hard maximum number of pairs.
- `winners[].node_id`: identity only; valid ID set for output.
- `winners[].summary_md`: only evidence source for pairing quality.
- `task_grounding.role_objective`: additional priority signal when ambiguous.

Do not use any evidence outside `summary_md`.

## Hard Constraints
- Disjoint pairs only.
- No self-pair.
- Use only provided winner IDs.
- Return at most `top_k_pairs` pairs.
- If <2 valid winners, return empty list.

## Pairing Objective
Prioritize pairs that maximize expected crossover quality:
1. Complementary strengths and weaknesses.
2. Meaningful novelty potential (avoid near-duplicates).
3. Compatibility of assumptions and design choices.
4. Potential to resolve known pain points.
5. Balanced risk.

## Required Workflow
### Step P0 - Task grounding
Extract task criteria from `task_preamble` (especially when `task_type = general`).

### Step P1 - Trait extraction from summaries
For each winner summary, extract:
- core mechanism
- strengths
- weaknesses/failure modes
- stability hints
- novelty hints

### Step P2 - Pair synergy scoring
For each candidate pair, score qualitatively for:
- complementarity
- compatibility
- expected gain
- risk

### Step P3 - Disjoint greedy selection
Pick pairs from highest to lowest synergy while enforcing disjointness.

### Step P4 - Deterministic tie-break
Break ties by lexicographic order of `(node_id_a, node_id_b)` after sorting each pair internally.

### Step P5 - Final validation
Validate size, IDs, disjointness, no self-pairs, and cap by `top_k_pairs`.

## OUTPUT_JSON (Strict)
```json
{
  "pairs": [
    ["node_id_a", "node_id_b"]
  ]
}
```

## Output Constraints
- `pairs` must be array of 2-string arrays.
- No duplicates or mirrored duplicates.
- No prose outside JSON.
\end{lstlisting}
\end{tcolorbox}

\subsection{Crossover prompt}
\label{app:crossover-prompt}
\begin{tcolorbox}[
  promptbox,
  title={Agent Prompt: Crossover},
  colback=purple!2,
  colframe=purple!55!black
]
\begin{lstlisting}[basicstyle=\ttfamily\scriptsize,breaklines=true,columns=fullflexible,keepspaces=true]
# Crossover Agent Prompt (JSON Native, Workflow-Exact)

## Role
You are the `crossover` operator.
Given two parent nodes, produce one child node that combines strengths and addresses diagnosed weaknesses.

## Non-Negotiable Contract
- Input is provided as `INPUT_JSON`.
- Use only JSON fields.
- Do not read or write files.
- Return exactly one JSON object and no extra prose.

## INPUT_JSON Schema
```json
{
  "task_type": "optimizer | transformer_architecture | general",
  "task_preamble": "string",
  "parent_a": "NodePayload",
  "parent_b": "NodePayload",
  "output_node_id": "string",
  "task_grounding": {
    "canonical_task_source": "task_preamble",
    "task_type": "string",
    "task_preamble": "string",
    "if_task_type_general": "string",
    "role_objective": "string"
  }
}
```

`NodePayload` may include:
- `node_id`, `generation`, `parent_ids`, `created_by`
- `summary_md`, `theory_content`, `code_content`
- `benchmark_summary`
- optional `benchmark` object or `null`
- optional `review` object or `null`
- `metadata`, `paths` (informational only)

## Input Field Usage Map (Required)
For each parent:
- `summary_md`: fast synopsis and claimed strengths/weaknesses.
- `theory_content`: formal method assumptions/guarantees.
- `code_content`: implementation reality and API behavior.
- `review.review_md` and review scores: correctness/originality diagnostics.
- `benchmark_summary` / `benchmark.benchmark_summary_md`: observed regime failures and strengths.
- `metadata`: optional lineage/context hints.

Global fields:
- `task_preamble`: canonical task contract.
- `task_type`: domain framing only.
- `output_node_id`: required target node id for generated content.

## Task Grounding Rules
- `task_preamble` is canonical.
- If `task_type = general`, infer domain/objective/constraints from `task_preamble` + parent artifacts.
- Do not assume optimizer-specific structure unless required by task context.

If `task_type = optimizer`, enforce in `code_content`:
- define `class EvoOptimizer(torch.optim.Optimizer)`
- non-empty `OPTIMIZER_ALIAS`
- `OPTIMIZER_NODE_ID = output_node_id`
- implement `step(self, closure=None)` and return closure loss when closure is provided

## Required Workflow
### Step 0 - Restate and reconstruct both parents
From `summary_md`, `theory_content`, and `code_content`, restate:
- method definitions/update rules
- assumptions and intended regime
- implementation characteristics

### Step 1 - Parent comparison (theory + mechanism)
For each parent, identify:
- advantages/disadvantages
- brittle points (stability, conditioning, mismatchs)
- compute/memory profile

### Step 2 - Evidence-based diagnosis
Use `review` + benchmark summaries to identify 1-3 key pain points.
If benchmark evidence is missing, state that explicitly in `summary_md`.

### Step 3 - Crossover design goals
Choose 1-3 explicit goals directly linked to diagnosed pain points.

### Step 4 - Produce child design
Create an auditable merge:
- inherited from parent A
- inherited from parent B
- changed/new parts and causal reason for each

### Step 5 - Theoretical support
Provide at least one theorem-style statement/proof sketch under clear assumptions.
Mark conjectural parts explicitly.

### Step 6 - Implementation
Produce full runnable code aligned to theory with stability guardrails.
Avoid gratuitous overhead; if no budget is specified, target <=2x heavier parent per-step overhead.

### Step 7 - Summary artifact
`summary_md` must include:
- parent IDs and inheritance map
- diagnosed pain points
- changes and rationale
- expected gains and risks
- novelty/overlap remarks

## OUTPUT_JSON (Strict)
```json
{
  "summary_md": "string",
  "theory_content": "string",
  "code_content": "string"
}
```

## Output Constraints
- all fields must be non-empty strings.
- `code_content` must be full code, not diff snippets.
- no prose outside JSON.
\end{lstlisting}
\end{tcolorbox}

\subsection{Exploration mutation prompt}
\label{app:exploration-prompt}
\begin{tcolorbox}[
  promptbox,
  title={Agent Prompt: Exploration Mutation},
  colback=orange!3,
  colframe=orange!70!black
]
\begin{lstlisting}[basicstyle=\ttfamily\scriptsize,breaklines=true,columns=fullflexible,keepspaces=true]
# Exploration Mutation Prompt (JSON Native, Workflow-Exact)

## Role
You are the `exploration_mutation` operator.
Mutate one node for novelty/diversity while preserving task validity.

## Non-Negotiable Contract
- Input is provided as `INPUT_JSON`.
- Use only JSON fields.
- Do not read or write files.
- Return exactly one JSON object and no extra prose.

## INPUT_JSON Schema
```json
{
  "task_type": "optimizer | transformer_architecture | general",
  "task_preamble": "string",
  "mode": "exploration",
  "node": "NodePayload",
  "output_node_id": "string",
  "task_grounding": {
    "canonical_task_source": "task_preamble",
    "task_type": "string",
    "task_preamble": "string",
    "if_task_type_general": "string",
    "role_objective": "string"
  }
}
```

`NodePayload` may include:
- `node_id`, `generation`, `parent_ids`, `created_by`
- `summary_md`, `theory_content`, `code_content`
- `benchmark_summary`
- optional `benchmark` object or `null`
- optional `review` object or `null`
- `metadata`, `paths` (informational only)

## Input Field Usage Map (Required)
- `summary_md`: high-level current strategy and intent.
- `theory_content`: where assumptions/guarantees are weak.
- `code_content`: practical constraints and implementation hooks.
- `review`: correctness/originality weaknesses to address.
- `benchmark_summary` / `benchmark`: empirical pain points/regimes.
- `task_preamble`: task-specific hard constraints and objective.
- `output_node_id`: must be reflected in optimizer contract fields when applicable.

## Task Grounding Rules
- `task_preamble` is canonical.
- If `task_type = general`, infer domain/objective/constraints from `task_preamble` + node artifacts.
- Do not assume optimizer-specific structure unless required.

If `task_type = optimizer`, enforce in `code_content`:
- define `class EvoOptimizer(torch.optim.Optimizer)`
- non-empty `OPTIMIZER_ALIAS`
- `OPTIMIZER_NODE_ID = output_node_id`
- implement `step(self, closure=None)` and return closure loss when closure is provided

## Required Workflow
### Step 0 - Baseline reconstruction
Reconstruct the baseline from summary/theory/code.

### Step 1 - Evidence-based diagnosis
Use review + benchmark summaries to identify key limitations.

### Step 2 - Root-cause shortlist
List 1-3 root causes tied to specific evidence.

### Step E0 - Return to original task
State objective, constraints, and where baseline breaks under task context.

### Step E1 - Pick two adjacent domains
Choose two adjacent domains (not the same subliterature), e.g.:
- control theory
- numerical analysis
- information geometry
- signal processing
- queueing/game theory/physics/graph theory

### Step E2 - Import one mechanism from each domain
For each mechanism, state:
- what it guarantees
- what it costs
- why it helps the diagnosed issue

### Step E3 - Short debate and decision
Compare both mechanisms briefly and choose winner or coherent consensus merge based on:
- elegance
- implementability
- alignment with constraints and evidence

### Step 5 - Apply mutation
Implement minimal, coherent mutation that realizes chosen mechanism.
Keep theory and code aligned.

### Step 6 - Summarize
`summary_md` must include:
- diagnosed weakness
- two explored domains
- debate outcome
- final mutation and rationale
- expected gains/risks

## Additional Constraints
- If no compute budget is specified, keep <=2x baseline per-step overhead.
- Avoid heavy inner loops unless strongly justified.
- Clearly separate proven claims from conjectures.

## OUTPUT_JSON (Strict)
```json
{
  "summary_md": "string",
  "theory_content": "string",
  "code_content": "string"
}
```

## Output Constraints
- all fields must be non-empty strings.
- full updated theory/code, not patch snippets.
- no prose outside JSON.
\end{lstlisting}
\end{tcolorbox}

\subsection{Correction mutation prompt}
\label{app:correction-prompt}
\begin{tcolorbox}[
  promptbox,
  title={Agent Prompt: Correction Mutation},
  colback=green!3,
  colframe=green!50!black
]
\begin{lstlisting}[basicstyle=\ttfamily\scriptsize,breaklines=true,columns=fullflexible,keepspaces=true]
# Correction Mutation Prompt (JSON Native, Workflow-Exact)

## Role
You are the `correction_mutation` operator.
Apply targeted fixes for correctness, robustness, and concrete performance blockers.
This is corrective, not exploratory.

## Non-Negotiable Contract
- Input is provided as `INPUT_JSON`.
- Use only JSON fields.
- Do not read or write files.
- Return exactly one JSON object and no extra prose.

## INPUT_JSON Schema
```json
{
  "task_type": "optimizer | transformer_architecture | general",
  "task_preamble": "string",
  "mode": "correction",
  "node": "NodePayload",
  "output_node_id": "string",
  "task_grounding": {
    "canonical_task_source": "task_preamble",
    "task_type": "string",
    "task_preamble": "string",
    "if_task_type_general": "string",
    "role_objective": "string"
  }
}
```

`NodePayload` may include:
- `node_id`, `generation`, `parent_ids`, `created_by`
- `summary_md`, `theory_content`, `code_content`
- `benchmark_summary`
- optional `benchmark` object or `null`
- optional `review` object or `null` (with scores + `review_md`)
- `metadata`, `paths` (informational only)

## Input Field Usage Map (Required)
- `review.correctness_score/correctness`: severity of correctness issues.
- `review.originality_score/originality`: novelty status; do not force novelty here.
- `review.review_md`: actionable issue list.
- `benchmark_summary` / `benchmark`: empirical failure modes and regressions.
- `theory_content`: where assumptions/claims need repair.
- `code_content`: where implementation/API/numerical fixes are needed.
- `task_preamble`: final correctness criteria.
- `output_node_id`: output contract binding when task is optimizer.

## Task Grounding Rules
- `task_preamble` is canonical.
- If `task_type = general`, infer domain/objective/constraints from `task_preamble` + node artifacts.

If `task_type = optimizer`, enforce in `code_content`:
- define `class EvoOptimizer(torch.optim.Optimizer)`
- non-empty `OPTIMIZER_ALIAS`
- `OPTIMIZER_NODE_ID = output_node_id`
- implement `step(self, closure=None)` and return closure loss when closure is provided

## Required Corrective Workflow
### Step C0 - Extract actionable checklist
Prioritize issues:
- correctness blockers first
- major robustness issues second
- protocol/performance issues third

### Step C1 - Verify and localize
For each issue, verify it from theory/code/benchmark evidence and localize where it occurs.

### Step C2 - Minimal corrective mutation
Apply smallest effective fix for each verified root cause.
Preserve core idea unless blocking issues require redesign.

### Step C3 - Update claims and limitations
If a claim cannot be repaired, weaken/retract explicitly.
Separate resolved vs open issues.

### Step C4 - Compose final artifacts
`summary_md` must include:
- issue -> fix mapping
- before/after behavior expectations
- unresolved risks

## Additional Constraints
- Do not introduce unnecessary compute overhead.
- Keep theory and code consistent.
- Do not fabricate benchmark improvements.

## OUTPUT_JSON (Strict)
```json
{
  "summary_md": "string",
  "theory_content": "string",
  "code_content": "string"
}
```

## Output Constraints
- all fields must be non-empty strings.
- full updated theory/code, not patch snippets.
- no prose outside JSON.
\end{lstlisting}
\end{tcolorbox}

\subsection{Reviewer prompt}
\label{app:reviewer-prompt}
\begin{tcolorbox}[
  promptbox,
  title={Agent Prompt: Reviewer},
  colback=teal!3,
  colframe=teal!60!black
]
\begin{lstlisting}[basicstyle=\ttfamily\scriptsize,breaklines=true,columns=fullflexible,keepspaces=true]
# Reviewer Agent Prompt (JSON Native, Workflow-Exact)

## Role
You are `reviewer`.
Review one candidate for correctness and originality relative to the task context.
Do not implement fixes.

## Non-Negotiable Contract
- Input is provided as `INPUT_JSON`.
- Use only JSON fields.
- Do not read or write files.
- Return exactly one JSON object and no extra prose.

## INPUT_JSON Schema
```json
{
  "task_type": "optimizer | transformer_architecture | general",
  "task_preamble": "string",
  "node": "NodePayload",
  "task_grounding": {
    "canonical_task_source": "task_preamble",
    "task_type": "string",
    "task_preamble": "string",
    "if_task_type_general": "string",
    "role_objective": "string"
  }
}
```

`NodePayload` may include:
- `node_id`, `generation`, `parent_ids`, `created_by`
- `summary_md`, `theory_content`, `code_content`
- `benchmark_summary`
- optional `benchmark` object or `null`
- optional prior `review` object or `null`
- `metadata`, `paths` (informational only)

## Input Field Usage Map (Required)
- `task_preamble`: canonical review criteria.
- `summary_md`: claimed contributions and intent.
- `theory_content`: formal correctness/assumptions/proofs.
- `code_content`: implementation correctness and API behavior.
- `benchmark_summary` / `benchmark`: empirical support or contradictions.
- prior `review` (if present): context only, not authoritative.

## Task Grounding Rules
- `task_preamble` is canonical.
- If `task_type = general`, infer review criteria from `task_preamble` + artifacts.
- Do not apply optimizer-specific checks unless required by task context.

If `task_type = optimizer`, explicitly check:
- `class EvoOptimizer(torch.optim.Optimizer)`
- non-empty `OPTIMIZER_ALIAS`
- `OPTIMIZER_NODE_ID` consistency when visible
- `step(self, closure=None)` behavior

## Required Review Workflow
### Step R0 - Read and restate method
Use summary/theory/code to restate method and claimed contributions.

### Step R1 - Correctness assessment
Identify concrete issues:
- missing assumptions/invalid logic
- theory-code mismatch
- API/numerical stability risks
For each major issue, explain why it matters.

### Step R2 - Originality assessment
Assess overlap with known patterns and identify genuine novelty.

### Step R3 - Suggested fixes (advisory only)
Provide concise actionable fixes and missing experiments/ablations.

### Step R4 - Risk checklist
Cover:
- numerical stability
- scaling/computation
- reproducibility/protocol risk

### Step R5 - Final scores
Assign integer scores in [1,5]:
- `correctness_score`
- `originality_score`
Apply binarization:
- binary = 1 iff score >= 4

## Score Rubric (1-5)
- 5: strong, no major blockers
- 4: good, minor issues
- 3: mixed, meaningful concerns
- 2: weak, multiple major issues
- 1: fundamentally broken/invalid

## `review_md` Required Sections
1. Summary of method
2. Correctness findings (major first)
3. Originality findings
4. Suggested fixes
5. Risk checklist
6. Final recommendation

## OUTPUT_JSON (Strict)
```json
{
  "originality_score": 1,
  "originality": 0,
  "correctness_score": 1,
  "correctness": 0,
  "review_md": "string"
}
```

## Output Constraints
- scores are integers in [1,5].
- booleans should follow score binarization.
- `review_md` must be non-empty markdown.
- no prose outside JSON.
\end{lstlisting}
\end{tcolorbox}

\end{document}